\documentclass[sigconf]{acmart}

\AtBeginDocument{%
  \providecommand\BibTeX{{%
    \normalfont B\kern-0.5em{\scshape i\kern-0.25em b}\kern-0.8em\TeX}}}

\copyrightyear{2024}
\acmYear{2024}
\setcopyright{rightsretained}
\acmConference[IUI '24]{29th International Conference on Intelligent User Interfaces}{March 18--21, 2024}{Greenville, SC, USA}
\acmBooktitle{29th International Conference on Intelligent User Interfaces (IUI '24), March 18--21, 2024, Greenville, SC, USA}
\acmDOI{10.1145/3640543.3645210}
\acmISBN{979-8-4007-0508-3/24/03}
 
\usepackage{soul}
\usepackage{tikz, makecell, colortbl}
\usetikzlibrary{arrows,decorations.markings,shapes,matrix,positioning, calc, fit,decorations.pathreplacing,angles,quotes,backgrounds}
\usepackage{siunitx}
\usepackage{multirow}
\usepackage{subcaption}
\usepackage{enumitem}
\usepackage{hyperref}
\usepackage{framed}

\definecolor{customrowcolor}{gray}{0.9}
\definecolor{shadecolor}{gray}{0.9}

\makeatletter
\renewcommand{\sectionautorefname}{\S\@gobble}
\renewcommand{\subsectionautorefname}{\S\@gobble}
\renewcommand{\subsubsectionautorefname}{\S\@gobble}
\makeatother

\newcommand{\ex}[1]{\mathbb{E}[#1]}

\usepackage{acmart-taps}

\begin{document}

\title[The Impact of Explanations on Fairness in Human-AI Decision-Making]{The Impact of Explanations on Fairness in Human-AI Decision-Making: Protected vs Proxy Features}

\author{Navita Goyal}
\authornote{Both co-first authors contributed equally to this work, and each has the right to list their name first on their CV.}
\email{navita@cs.umd.edu}
\affiliation{%
  \institution{University of Maryland}
  \country{}
}

\author{Connor Baumler}
\authornotemark[1]
\email{baumler@cs.umd.edu}
\affiliation{%
  \institution{University of Maryland}
  \country{}
}

\author{Tin Nguyen}
\email{tintn@cs.umd.edu}
\affiliation{%
  \institution{University of Maryland}
  \country{}
}

\author{Hal Daum\'e III}
\email{me@hal3.name}
\affiliation{%
  \institution{University of Maryland \& Microsoft
Research}
    \country{}
}


\begin{abstract}
AI systems have been known to amplify biases in real-world data. Explanations may help human-AI teams address these biases for fairer decision-making. Typically, explanations focus on salient input features. If a model is biased against some protected group, explanations may include features that demonstrate this bias, but when biases are realized through proxy features, the relationship between this proxy feature and the protected one may be less clear to a human. In this work, we study the effect of the presence of protected and proxy features on participants' perception of model fairness and their ability to improve demographic parity over an AI alone. Further, we examine how different treatments---explanations, model bias disclosure and proxy correlation disclosure---affect fairness perception and parity. We find that explanations help people detect direct but not indirect biases. Additionally, regardless of bias type, explanations tend to increase agreement with model biases. Disclosures can help mitigate this effect for indirect biases, improving both unfairness recognition and decision-making fairness. We hope that our findings can help guide further research into advancing explanations in support of fair human-AI decision-making.
\end{abstract}



\maketitle

\section{Introduction}

Improving the fairness and trustworthiness of AI systems is often cited as a goal of explainable AI (XAI)~\citep[e.g.,][]{velez2017towards, lipton2018mythos, BARREDOARRIETA202082, das2020opportunities, langer2021what, du2021ffairness, Warner2021-WARMAI-5}. 
Research in XAI aims to improve fairness in human-AI decision-making by providing insights into model predictions, and thereby allowing humans to understand and correct for model biases. 
On the other hand, in the context of human-AI decision-making, previous work has noted that humans often over-rely on AI predictions, and explanations can exacerbate this concern~\cite{bucinca2021to}.
This is especially troubling if the underlying model contains systematic biases, which may go unnoticed even when teamed with a human. 
For the human-AI team to succeed, the human needs to be able to determine when to rely on or override potentially biased AI predictions. 
Previous work has shown that explanations can help human-AI teams alleviate model biases when those biases depend directly on protected attributes~\cite{dodge2019explaning,schoeffer2022explanations}, but little is known in the very common case that  protected attributes are not explicitly included, and rather the features used for prediction contain proxies thereof (e.g., zip code for race, length of credit for age, and university attended for gender). 
In particular, it may be difficult for humans to identify and resolve biased model predictions based on the proxy features present in real-world data, even when explanations are provided. 

In this work, we study whether explanations can help people to identify model biases and to calibrate their reliance on a biased AI model.
We extend work in this space by moving beyond direct biases that are revealed through the use of protected (i.e., sensitive) to indirect biases that are revealed through proxy features that may be less obvious to a human.
Further, we examine whether explicitly disclosing model biases and correlations between the proxy and protected features can help humans calibrate their trust in a biased model. 
Our study aims to evaluate whether explanations can directly help notice model biases, even when the biases are obfuscated by the presence of proxy features and whether explanations can help users correct model biases when they are known to be present, through the use of bias disclosure and correlation disclosure.
We study the effect of these treatments (explanations, model bias disclosure, and proxy correlation disclosure) on the 
fairness, including fairness perception and fairness in decision-making (measured by group-wise parity), as well as the accuracy of the decisions made by human-AI teams. 

We conduct our study in the context of micro-lending outcome prediction---a setting that entails judging whether a loan applicant will fulfill their loan request based on profile information of the applicant (e.g., size of the loan, borrower occupation, etc). %
For our experiments, we use semi-synthetic data where the majority of the features in an applicant profile as well as the final loan repayment status comes from the website Prosper.\footnotemark{} 
\footnotetext{\label{footnote:prosper}\href{https://www.kaggle.com/datasets/yousuf28/prosper-loan}{https://www.kaggle.com/datasets/yousuf28/prosper-loan}}
To incorporate fairness considerations, we add to the applicant profiles (binary) gender, which is a protected feature, and university which, when considering women's vs co-ed colleges, can be a proxy for gender. %
Because we seek to test whether people can correct for model bias, we intentionally train a biased predictor with outcomes skewed against applicants with gender assigned as female or university assigned as a women's college.%

We find that explanations alone can help people notice unfairness in the case of direct bias (through protected features, e.g., gender), but not in the case of indirect bias (through proxy features, e.g., university).
Surprisingly, regardless of whether people notice the unfairness in the AI decisions, explanations lead people to accept model's biased decisions leading to less fair decisions. 
In the case of direct bias, as participants often recognize clear-cut gender bias before an explicit disclosure, disclosing model biases does not further affect participants' fairness perception. However, in the case of indirect bias, disclosing both model bias and the correlation between protected and proxy features or disclosing partial information with the addition of explanations significantly improves participants' awareness of the unfairness.
However, contrary to explanations alone, this change is \textit{not} paired with a worsening of decision-making fairness. 
Instead, with these disclosures, people increase their rate of positive predictions for the disadvantaged group, improving decision-making fairness. Our work aims to highlight methods to assist users in effectively leveraging explanations, especially in scenarios where bias may be indirect and not apparent through explanations alone.

\section{Background and Related Work}

\paragraph{Biases in Models and Humans}
Both models and humans can be biased. Humans are known to exhibit many implicit and unconscious biases~\cite{greenwald1995implicit}. For instance, ~\citet{bertrand2004emily-and-greg} find that an applicant with a ``White-sounding name'' on a resume that is otherwise identical to a resume with an ``African-American-sounding name'' is more likely to receive an interview callback. 

Models, in turn, can inherit human-like biases (e.g., through biased data~\cite[i.a.]{barocas2016big}), even if this is not intended by the developers. For instance, \citet{angwin2016machine} show that training on data collected from a racist justice system can lead to a model that predicts that white defendants are less likely to recidivate than their black peers.

This paper explores how humans interact with predictions from a biased model, wherein AI systems may be able to uncover helpful patterns in existing data and humans may be able to apply their contextual understanding and societal awareness to contribute to correcting these biases within the model.

\paragraph{XAI and Decision-Making}
The potentially complementary strengths and weaknesses of humans and machines raises a question of whether human-AI teams can overcome the biases that exist in each individually (e.g., in the case of recidivism prediction~\cite[e.g.,][]{duan2022crowdsourced, wang2022comparisons, chiang2023twoheads}).
Existing work in explainable AI (XAI) has focused on providing explanations of the model decisions to help improve the outcomes of human-AI decision-making~\cite{lai2019deception, feng2019what, yang2020visual, lai2020chicago, bucinca2020proxytasks, Carton_Mei_Resnick_2020, zhang2020calibration, bansal2021whole, poursabzi2021manipulating, liu2021interactive, gonzalez-etal-2021-explanations, bucinca2021to, wang2021helpful, wang2022comparisons, cau2023highuncertainty, chen2023understanding, sivaraman2023ClinicianAcceptance, kim2023assisting, vasconcelos2023explanations}. 
However, these studies find varying utility of explanations. Much work has found that explanations can help humans collaborate more effectively with AI~\cite{feng2019what, lai2019deception, lai2020chicago, yang2020visual, gonzalez-etal-2021-explanations, vasconcelos2023explanations, goyal2023what, cau2023highuncertainty}, for instance helping them answer trivia questions more accurately~\cite{feng2019what} or understanding how the AI system works~\cite{cheng2019ui}. Other work has found that explanations can worsen human-AI performance~\cite{bucinca2020proxytasks, Carton_Mei_Resnick_2020, bucinca2021to, bansal2021whole, poursabzi2021manipulating, wang2021helpful, wang2022comparisons, chen2023understanding, sivaraman2023ClinicianAcceptance, kim2023assisting} even below the performance of the human or AI alone. 
Further, the utility of explanation can also vary based on the participant's level of expertise in the task~\cite[e.g.,][]{wang2022comparisons}, the participant's math and logic skills~\cite{suresh2020misplaced}, how easy the explanations are to understand~\cite{yang2020visual}, etc. 

Beyond explanations, other work has considered how further transparency can or cannot be beneficial to a human-AI team such as tutorials~\cite{lai2019deception}, disclosing model confidence~\cite{rechkemmer2022dating}, disclosing model accuracy~\cite{dzindolet2003reliance}, and disclosing whether test examples fall into the scope of model training data~\cite{chiang2021youdbetter}. 

Building upon previous research, this paper investigates the impact of explanations on the behavior of a human-AI team, especially their influence on the fairness of human-AI decisions in cases where the underlying model exhibits bias. In addition to explanations, we draw inspiration from work considering other methods of improving transparency in human-AI decision-making~\cite[e.g.,][]{rechkemmer2022dating, dzindolet2003reliance, chiang2021youdbetter}, exploring the implications of disclosing model bias and the correlation between protected and proxy features on the overall fairness of a human-AI team.

\paragraph{XAI and Fairness}

Improving model fairness is often cited as a potential benefit of XAI systems~\cite{velez2017towards, lipton2018mythos, BARREDOARRIETA202082, das2020opportunities, langer2021what, du2021ffairness, Warner2021-WARMAI-5}. XAI is hoped to help ``diagnose the reasons that lead to algorithmic discrimination''~\cite{du2021ffairness}, to ``highlight an incompleteness'' in problem formalization that leads to unfairness~\cite{velez2017towards}, or to show compliance with fairness requirements~\cite{Warner2021-WARMAI-5}.

Previous work has examined how explanations affect humans' perceptions of AI systems' fairness~\cite{rader2018explanations, lee2019procedural, binns2018reducing, dodge2019explaning, schoeffer2022explanations,yurrita2023disentangling}. 
\citet{rader2018explanations} find that participants that are told that an AI system is being used in decision-making rate the system as significantly less fair even without any specific system information. 
\citet{lee2019procedural} find that explanations of an AI system's general decision-making process do not increase perceived fairness while input-output level explanations of individual outcomes have mixed effects on fairness perceptions. 
\citet{binns2018reducing} consider how four different styles of explanations affect justice perception, finding no clear winner between the approaches. \citet{dodge2019explaning} further study the explanations styles in~\cite{binns2018reducing} and find that local explanations (such as presenting outcomes for similar examples) help surface fairness discrepancies between different cases while global explanations (such as describing how each feature influenced the decision for a given example) increase user confidence in their understanding of the model and enhance users' fairness perceptions. 

As self-reported perceptions do not always align with observed behaviors in human-AI decision-making~\cite{bucinca2020proxytasks, papenmeier2022complicated}, recent work has begun to expand out of fairness perceptions and into observed fairness in decision-making~\cite{schoeffer2022explanations, wang2023effects}. 
\citet{schoeffer2022explanations} study how explanations can help users appropriately rely on potentially unfair AI predictions. %
They find that explanations that highlight protected features negatively affect fairness perceptions and that decreases in fairness perception are associated with an increase in overrides of AI predictions, even on examples where this override is detrimental to the fairness of the human-AI team. 
\citet{wang2023effects} study the effects of the level of model bias and the presence of explanation on the fairness of human decisions. They find that explanations lead participants to make more unfair decisions, even when participants were no longer given access to model predictions or explanations. %

Existing work has primarily studied fairness when the model decision is directly based on a protected feature, like gender or race. However, models can produce biased outcomes, even without access to protected features, by relying on proxy features~\cite{pedreshi2008discrimination, kilbertus2017avoiding}. For instance, a model that has direct access to a ``race'' feature and one with access to features like zip code, name, or language spoken at home could produce similarly biased predictions. In contrast to existing work considering the relationship between explanations and fairness perceptions or decision-making fairness, we consider not only direct bias through a protected feature but also indirect bias through a proxy feature. 

\section{Research Questions}\label{sec:rqs}

We study the effect of explanations and disclosures in improving the fairness perception and fairness of decisions made by human-AI teams.
In our study, model biases can be direct: stemming from the protected feature (gender), or indirect: stemming from a proxy feature (university) that is correlated with the protected feature. For explanation, we consider an input-influence explanation of how each feature contributed to the AI's prediction. For disclosures, participants may be told about the demographic parity (described in \aptLtoX[graphic=no,type=html]{\S\ref{sec:metrics}}{\autoref{sec:metrics}}) of the system (model bias disclosure) and the strength of correlation between the proxy and protected features (proxy correlation disclosure; see \aptLtoX[graphic=no,type=html]{\S\ref{subsec:procedure}}{\autoref{subsec:procedure}}). 
Our study addresses the following research questions:
\begin{description}
    \item{}
\textbf{RQ1a}: Are explanations beneficial to the fairness of a human-AI team?
    \item{}
\textbf{RQ2a}: Without explanations, does disclosing only model bias or disclosing model bias and proxy correlation benefit human-AI fairness? 
    \item{}
\textbf{RQ3a}: With explanations, does disclosing only model bias or disclosing model bias and proxy correlation benefit human-AI fairness? 
    \item{}
\textbf{RQ4a}: Does the joint intervention of adding explanations and disclosures benefit human-AI fairness?
     \item{}
\textbf{RQ1-4b}: Do the answers to RQ1-4a change when models exhibit direct (e.g., gender) vs indirect (e.g., university) bias?%
\end{description}

We consider the utility of explanations and disclosures under three lenses: the accuracy of \textbf{human's perception of fairness} and the improvement in \textbf{demographic parity in human-AI decision-making} over AI-only parity, and the \textbf{decision-making quality} (namely, accuracy, false negative rate (FNR), and false positive rate (FPR)) \textbf{of human-AI decisions} compared to the AI alone.

Beyond these primary research questions, we also consider:
\begin{description}
    \item{}
\textbf{RQ5}: Does dispositional trust (``an individual’s
enduring tendency to trust automation'' \cite{hoff2015trust, marsh2003trust}) affect decision-making and fairness perceptions when working with models exhibiting direct or indirect bias? 
    \item{}
\textbf{RQ6}: Do explanations and disclosures affect self-reported learned trust (based on ``past experience or the current interaction''~\cite{hoff2015trust, marsh2003trust}) in models exhibiting direct or indirect bias? 
\end{description}

In our study, we vary conditions based on the directness of bias, whether explanations are shown, and the kind of disclosure the participant receives. We consider six conditions. In the first three conditions, we do not show participants the explanations. Here, we consider one \textbf{Protected} condition with bias disclosure, and two Proxy conditions: one \textbf{Proxy with correlation disclosure} and one \textbf{Proxy without correlation disclosure}. We similarly consider three conditions with explanations allocating the biased feature and disclosure types in the same fashion.

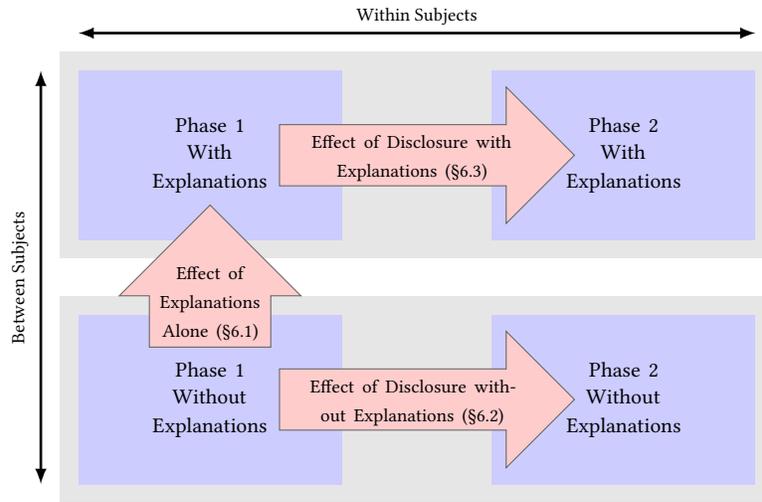
\begin{figure*}[t]
    \centering
    \tikzset{MyArrow/.style={single arrow, fill=red!20, align=center, draw=black!60, minimum width=5mm, minimum height=5mm,  inner sep=0mm, single arrow head extend=5mm}}
    \begin{tikzpicture}[]
        \path[fill=customrowcolor] (-4.75cm,.25cm) rectangle (4.75cm,3.00cm);
        \path[fill=customrowcolor] (-4.75cm,-.25cm) rectangle (4.75cm,-3.00cm); %
        \draw[latex-latex,  line width=1pt] (-4.5cm,3.25cm) -- (4.5cm,3.25cm);
        \node[above] at (0,3.25cm) {\footnotesize Within Subjects};
        \draw[latex-latex,  line width=1pt] (-5.0cm,2.75cm) -- (-5.0cm,-2.75cm);
        \node[anchor=north,rotate=90] at (-5.5cm,0) {\footnotesize Between Subjects};
        \matrix (m) [
            matrix of nodes, 
            column sep = 20mm,
            row sep = 1cm,
            nodes={ fill=blue!20,  text=black, align=center, minimum width=3.5cm, minimum height=2.25cm, text width = 3cm, line width=0mm}
        ] {
             \node(p1e)[rectangle]{\small Phase 1\\With\\ Explanations}; &
             \node(p2e){\small Phase 2\\With \\Explanations}; \\
             \node(p1ne){\small Phase 1\\Without \\Explanations}; &
             \node(p2ne){\small Phase 2\\Without \\Explanations}; \\
        };
        \path (p1ne.north) -- (p1e.south) node[midway, MyArrow,shape border rotate=90,text width=1.6cm,  text depth = .9cm, yshift=-11, single arrow head extend=4mm, minimum width=2.0cm] {\footnotesize Effect of Explanations Alone (\S \ref{sec:expl})}; 
        \path (p1ne.east) -- (p2ne.west) node[midway, MyArrow, text width=3.5cm, text height=.3cm, text depth = .5cm, xshift=-2] {\footnotesize Effect of Disclosure without Explanations (\S \ref{sec:disc})}; 
        \path (p1e.east) -- (p2e.west) node[midway, MyArrow, text width=3.5cm, text height=.3cm, text depth = .5cm, xshift=-2] {\footnotesize Effect of Disclosure with Explanations (\S \ref{sec:disc_expl})}; 
    \end{tikzpicture}
    \caption{Summary of primary effects considered in our study. Participants are assigned to either with or without explanations conditions and then complete the study moving horizontally from phase 1 to phase 2. We then compare the results of different combinations of phases and explanation conditions to investigate the effects of explanations alone, disclosures without explanations, and disclosures with explanations.}
\label{fig:expl_disclosure}
\end{figure*}

We assess the effect of explanations (RQ1a) by comparing conditions with and without explanations (before any disclosures) in a between-subjects analysis. We assess the effect of disclosures without explanations (RQ2a) and the effect of disclosures with explanations (RQ3a) in a within-subject analysis comparing fairness perceptions and human-AI decision pre- and post-disclosures. This allows us to study how disclosures may help participants identify model biases over what is apparent before any disclosures. These first three effects are summarized in \autoref{fig:expl_disclosure}. Lastly, we assess the effect of explanations and disclosures jointly (RQ4a) by comparing conditions in which participants are \textit{not} shown explanations pre-disclosures with conditions in which participants are shown explanations post-disclosures. These experiments are repeated for protected and proxy conditions to assess the differences in interventions therein (RQ1-4b).

\section{Study Design}
To answer the research questions posed in \aptLtoX[graphic=no,type=html]{\S\ref{sec:rqs}}{\autoref{sec:rqs}}, we study decisions made by human-AI teams.\footnote{This study design is IRB approved. %
} In this study, the AI teammate is a classification model trained on partially-synthetic data in the context of loan prediction. We choose the task of loan prediction from a micro-lending platform as it is a decision-making task performed by laypeople which means that crowd-workers are more likely to have intuitions about the task and the features used in predictions. In our study, participants are shown either the protected feature of binary gender\footnote{We only consider binary gender in this study. Since each participant sees only a handful of examples per task phase, it would be difficult to both show the participants a statistically realistic number of non-binary applicants and get a good sense of how participants handle anti-trans model bias. We leave this for future work.} or the proxy feature of university.

\subsection{Procedure}
\label{subsec:procedure}
\begin{figure*}
    \footnotesize
    \centering
    \begin{tikzpicture}
        \matrix (m) [
                matrix of nodes, 
                column sep = 1mm,
                row sep = 1mm,
                nodes={ fill=white,  text=black, align=left, minimum height=2.5cm, text width = 1.5cm, line width=0mm, anchor=center}
            ] {
                 \node(1)[minimum height = 1mm,align=center]{S0};&
                 \node[minimum height = 1mm,align=center, text width=.05mm]{};&
                 \node(2)[minimum height = 1mm,align=center]{P0};&
                 \node[minimum height = 1mm,align=center, text width=.05mm]{};&
                 \node(3)[minimum height = 1mm,align=center]{P1};&
                 \node[minimum height = 1mm,align=center, text width=.05mm]{};&
                 \node(4)[minimum height = 1mm,align=center]{S1};&
                 \node[minimum height = 1mm,align=center, text width=.05mm]{};&
                 \node(5)[minimum height = 1mm,align=center]{D};&
                 \node[minimum height = 1mm,align=center, text width=.05mm]{};&
                 \node(6)[minimum height = 1mm,align=center]{P2};&
                 \node[minimum height = 1mm,align=center, text width=.05mm]{};&
                 \node(7)[minimum height = 1mm,align=center]{S2};\\
                 \node(8)[]{Initial survey:\\ \vspace{2pt} Dispositional trust}; &
                 \node[regular polygon, regular polygon sides=3,text width=.05mm , minimum height = .05mm, rotate=-90]{};&
                 \node(9)[]{Instructions and warm-up}; &
                 \node[regular polygon, regular polygon sides=3,text width=.05mm , minimum height = .05mm, rotate=-90]{};&
                 \node(10)[align=center]{\includegraphics[width=.9\textwidth, trim={110 0 116 0},clip]{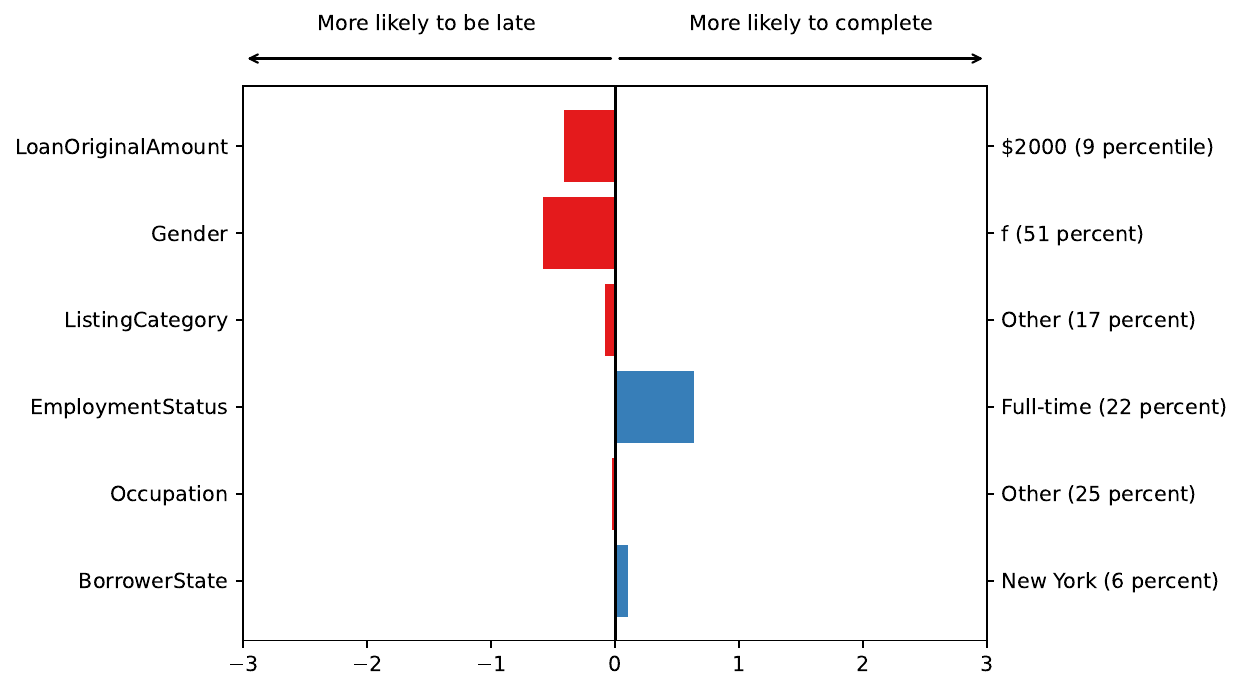}
                 ($\times10$)}; &
                 \node[regular polygon, regular polygon sides=3,text width=.05mm , minimum height = .05mm, rotate=-90]{};&
                 \node(11)[]{Survey:\\ \vspace{2pt} Learned trust and fairness perceptions}; &
                 \node[regular polygon, regular polygon sides=3,text width=.05mm , minimum height = .05mm, rotate=-90]{};&
                 \node(12)[]{Disclosures:\\ \vspace{2pt} \textbf{model bias} and \\(according to condition) \textbf{feature correlation}}; &
                 \node[regular polygon, regular polygon sides=3,text width=.05mm , minimum height = .05mm, rotate=-90]{};&
                 \node(13)[align=center]{\includegraphics[width=.9\textwidth, trim={110 0 116 0},clip]{figures/f_Completed_e.pdf}
                 ($\times10$)}; &
                 \node[regular polygon, regular polygon sides=3,text width=.05mm , minimum height = .05mm, rotate=-90]{};&
                 \node(14)[]{Survey:\\ \vspace{2pt} Learned trust and fairness perceptions}; \\
            };
            
            \begin{scope}[on background layer]
            \node(i) [fill=black!20, fit={(8)(14)}]{};
            \end{scope}
    \end{tikzpicture}
    \caption{Order of study phases.}
    \label{tab:order}
\end{figure*}

Our study procedure consists of three surveys (S0, S1, S2), one tutorial and warm-up phase (P0), two task phases (P1,~P2), and a disclosure interlude (D) ordered as shown in \autoref{tab:order}.

\begin{figure*} \centering
\includegraphics[width=.79\textwidth, trim={48 8 48 0},clip]{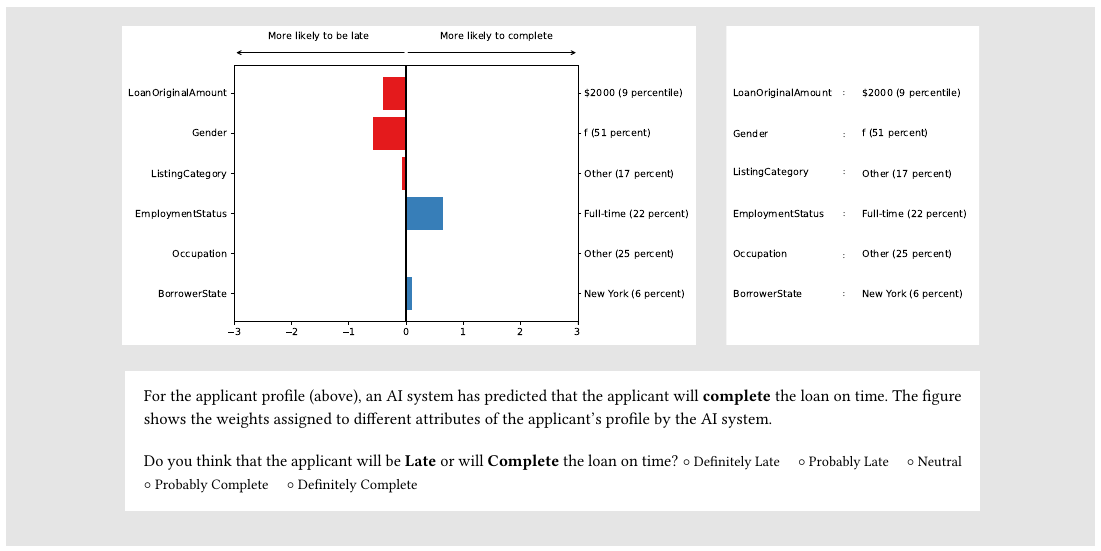}
    \caption{Example profile with explanation from the ``protected'' model (left) without an explanation (right) and question to the user (below). The predicted outcome is completing the loan on time. The labels on the left show the name of each feature. The labels on the right show the value of each feature for the current applicant and the percent/percentile of this value in the training data. For the explanation, on the x-axis positive blue values correspond to ``Complete'' predictions and negative red to ``Late''. See Figure~\ref{fig:task} in the Appendix for an example profile as shown in the study interface.}%
    \label{fig:example}
\end{figure*}

\paragraph{Task Phases}
In each task phase (P1, P2), the participant is shown 10 profiles of loan applicants:  their features and the overall AI prediction. Depending on the condition, they may or may not be shown an explanation of the AI prediction (\autoref{fig:example} left and right, respectively). 
This profile will, according to the condition, include either a ``gender'' or a ``university'' feature but not both. Participants are asked to mark on a five-point-scale whether they think the applicant will complete their loan on time or be late in repaying their loan (\autoref{fig:example}, below). Their response to this question serves as the decision made by the human-AI team.

In each phase, we control the distribution of gender and AI predictions. The participant sees applications from 2 women who are predicted as ``Complete'' and 3 women who are predicted as ``Late'' and vice versa for men. (This is true in the underlying data even if the participant and the model do not directly see each applicant's gender.) We hold this ratio constant to avoid any effect due to the gender distribution or the rejection rate observed by different participants.

To discourage participants from making decisions without any consideration of the prediction and (when applicable) explanation, we ask participants for a free-text justification of why they agreed or disagreed with the model prediction (or was neutral) after they have chosen their prediction on selected profiles. 
We randomly select one application in each gender + prediction combination for collecting these free-text justifications. These justifications also help us qualitatively assess the reasoning behind participants' decisions. 
Further, to help filter out low-quality responses, participants are shown an attention check question, asking them to recall the previous AI prediction (Figure~\ref{fig:attention_check} in the Appendix) after seeing the first applicant in P1.

\paragraph{Disclosures}

\begin{figure*}
\centering
    \begin{subfigure}{.32\textwidth}
    \framebox{\parbox[T]{\textwidth}{ 
    {\footnotesize 
    For decision-making tasks, such as microlending outcome prediction, AI systems can be biased against different demographic groups, such as gender, race, etc. These systems may be used to recommend acceptance for microlending applications (that is, to accept loan request if the applicant will likely complete the loan on time and reject it if the applicant will likely be late on the loan). Unfairness in the AI systems can potentially limit the access to loans for certain demographic groups.\\

    To avoid discrimination, decision makers should follow the 80\% rule: the acceptance rate for the disadvantaged group should be within $\mathbf{80\%}$ of the acceptance rate for the advantaged group.\\

    For the 10 applicants in phase 1, the model predicted 60\% of the men would \textit{complete} the loan on time and 40\% of the women would \textit{complete} the loan on time. This leads to the acceptance rate for the women to be about $\mathbf{65\%}$ of that of the men.
    
    }}}

    \caption{}
    \label{fig:bias_disclosure_in_body}
    \end{subfigure}\hspace{.05\textwidth}    
    \begin{subfigure}{.59\textwidth}
    \framebox{\parbox[T]{\textwidth}{ 
    {\footnotesize 
    
    One thing to note is that AI systems can be discrimininatory even based on features that you may not expect. For example, even if a system does not explicitly know applicants' gender, it can still discriminate against applicants who went to women's colleges. \\

    In the figure below, you can see the associations between different colleges and binary gender. (This is based on the historical data used to train our AI system.)

    \begin{center}
        \includegraphics[width=.8\linewidth]{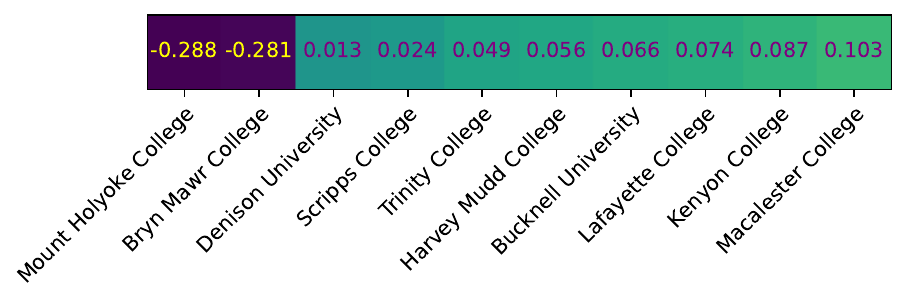}
    \end{center}

    The colleges towards the left (in purple) are more associated with women. On the other hand, the colleges towards the right (in green) are more associated with men. The values on the figure indicate the strength of association (the closer to zero, the weaker the association).
    
    }}}
    \caption{}
    \label{fig:proxy_disclosure_in_body}
    \end{subfigure}
    \caption{a) Bias disclosure. b) Full correlation disclosure. Proxy ``no correlation disclosure'' conditions include the top paragraph but with the example of a hiring system relying on the relationship between zip code and race. See Figure~\ref{fig:bias_disclosure} and Figure~\ref{fig:corr_disclosure} in the Appendix for how these disclosures are shown in the study interface.}\label{fig:disclosures}
\end{figure*}

Before proceeding to P2, participants may be shown general explanatory materials or specific disclosures on model bias and feature correlations. 
In the model bias disclosure (\aptLtoX[graphic=no,type=html]{Figure~\ref{fig:bias_disclosure_in_body}}{\autoref{fig:bias_disclosure_in_body}}), participants are told that the model they saw in P1 had a low demographic parity (below $80\%$) (see \aptLtoX[graphic=no,type=html]{\S\ref{sec:metrics}}{\autoref{sec:metrics}} for details about demographic parity). 
In the correlation disclosure \aptLtoX[graphic=no,type=html]{Figure~\ref{fig:proxy_disclosure_in_body}}{\autoref{fig:proxy_disclosure_in_body}}, participants are told the correlation between each university and gender in the model's training data. The bias disclosure is shown across conditions, whereas correlation disclosure is only shown in the proxy conditions \textit{with correlation disclosure}. In the proxy conditions \textit{without correlation disclosure}, participants are only told that models can rely on proxy features to make biased predictions, without specifying the correlation between gender and university. This is done to make participants aware of potential biases without explicitly disclosing the correlations. 

Based on the disclosures seen, participants are asked up to two comprehension questions (\aptLtoX[graphic=no,type=html]{Figure~\ref{fig:corr_check}}{\autoref{fig:corr_check}} in the Appendix). %
All are asked whether the model's demographic parity was above $80\%$. Those who received correlation disclosure are asked to select one university that is highly associated with women. %

Note that participants are not encouraged or primed to consider fairness explicitly before or during the first phase of the task. We only refer to fairness directly after phase 1. This allows us to measure how well participants can notice, or account for, unfairness when they aren't explicitly told to look out for it in phase 1. Subsequently, in phase 2, we can measure how participants perceive and account for unfairness when they know it is a salient concern. 

\paragraph{Surveys}\label{sec:surveys}
The three surveys (S0, S1, S2) aim to capture participants' trust and fairness perceptions.
All surveys include questions asking participants to rate their level of agreement with statements relating to trust (on a scale of $1$-$5$) ~\cite{hoffman2018metrics}. 
In S0, participants are asked about their trust in AI systems generally, assessing their dispositional trust (Figure~\ref{fig:pre_study_survey} in the Appendix). In S1 and S2, participants are asked about their trust in the system presented in the task phases, assessing their learned trust in the AI system that they interact with in the study (Figure~\ref{fig:survey_p2} in the Appendix). 

In the post-task surveys (S1 and S2), alongside trust-related questions, participants are also asked about their perception of the fairness of the system they have been interacting with (whether ``\textit{the AI system was fair across different genders}''). Additionally, participants are asked the reason(s) that led to their disagreements with AI such as the explanations including irrelevant features or the decisions being unfair towards applicants of different genders.

\paragraph{Tutorial and Warm-up}
In P0, participants are acclimatized to the task with a full tutorial example.
They are shown one tutorial example with a walk-through of the task, the AI decisions and explanations (when applicable). 
Then, they are shown warm-up examples. 
In the conditions without explanation, they are shown two examples with no AI prediction or explanation. In the conditions with explanations, they are first shown a version of this example with no AI prediction or explanation. This is designed to encourage participants to properly engage with the features present. Second, they are shown the same example with the AI feature explanation (still without any prediction) as this setting has been shown to benefit decision quality and support learning by encouraging participants to cognitively engage with explanations~\cite{gajos2022do}.

\subsection{Participants}
We recruit $369$ participants for our study through the crowdsourcing platform Prolific.\footnote{\href{https://www.prolific.com/}{https://www.prolific.com/}} Each participant is restricted to taking the study only once. Participation is restricted to US participants, fluent in English. We compensate all participants at an average rate of US\$$15$ per hour.
We discard responses that fail more than one attention check, leaving a total of $350$ participants, with $51, 48, 45$ participants in the \textit{protected condition}, the \textit{proxy condition with correlation disclosure}, and the \textit{proxy condition without correlation disclosure} without model explanation and $68, 69, 69$ participants in the three respective conditions with model explanations. $42\%$ of participants self-identified as women, $52\%$ as men, $3\%$ as non-binary/non-conforming, $3\%$ as transgender, and $1\%$ as a different gender identity, with $1\%$ of participants opting not to respond.\footnote{These do not add up to $100$ as participants may have selected multiple options.} $19\%$ of participants were between the ages of $18$-$25$, $46\%$ between $25$-$40$, $27\%$ between $40$-$60$, and $6\%$ over the age of $60$, with $2\%$ of participants opting not to respond.

\section{System Overview}

We conduct our study using model predictions and explanations from logistic regression models trained on partially synthetic micro-lending data.\footnote{\href{https://github.com/ctbaumler/protected-vs-proxy}{https://github.com/ctbaumler/protected-vs-proxy}}
Since the participant's perceptions of how the model is interacting with the profile features is key to answering our research questions, we want to avoid any potential confounding effects from using artificial or Wizard-of-Oz model explanations, or entirely synthetic data. 

The scenario of predicting whether an applicant will complete microloan repayment on time or will be late is one that our participants will likely be sufficiently familiar with to have reasonable prior intuitions about what features are relevant. A challenge is that under US law, protected features like gender cannot be considered when making loan allocation decisions~\cite{FRB2006} and, therefore, is not in the dataset that we consider. For this reason, we augment our data with a synthetic ``gender'' feature, which we correlate with outcome to induce model bias. We also generate a proxy feature, university, which allows us to finely control the level of correlation between the proxy and gender.

\paragraph{Data}\label{sec:data}
Our loan prediction data comes from a modified set of microloans from the website Prosper.\footref{footnote:prosper} %
The original dataset contains $79$ features of microloans including their status (completed, past due, etc). 
We group the loan statuses into ``Complete'' (including ``Final Payment in Progress''), ``Late'' (including ``Defaulted'' and ``Charged-Off''), or ``Other'' (including ``Current'' or ``Canceld''). We keep the $\sim$$14000$ profiles with ``Complete'' or ``Late'' statuses (with a $7$:$3$ train-test split). 
This grouped loan status is the feature that the participants and the model will predict. As showing all $79$ features to the participant may be overwhelming,~\cite{poursabzi2021manipulating} we select $5$ features (the original amount of the loan, the category of the listing, the applicant's occupation and employment status, and their state of residence) that are both important to loan prediction and are likely interpretable by a layperson. 

As described above, we synthetically generate values for our protected characteristic (binary gender). The existing applicants are assigned a gender in such a way that the ratio for ``Complete'' vs ``Late'' outcomes is $2$:$3$ for female applicants and vice versa for male applicants. This simulates historically biased data, which will cause the model to associate femaleness with being late on loans and maleness with completing them.

Using the generated ``gender'' feature, we further generate the proxy feature (university). %
We include co-ed and women's colleges, setting the joint distribution of gender and university such that most co-ed universities have relatively balanced gender ratios (See \autoref{fig:proxy_disclosure_in_body}). For women's colleges, %
the distributions reflect real-life statistics. We choose exclusively liberal arts colleges with similar US News rankings\footnote{\href{https://www.usnews.com/best-colleges/rankings/national-liberal-arts-colleges}{https://www.usnews.com/best-colleges/rankings/national-liberal-arts-colleges}} to avoid confounding due to the effect of perceptions of liberal arts vs non-liberal arts schools and perceptions of school rankings.

Since, in our biased dataset, gender is correlated with outcome and, of course, the existing features are correlated with outcome, all features may be weakly correlated with gender. To confirm that university is the only strong proxy in our data, we compare the correlation of each categorical and continuous feature with gender. For continuous features (and one-hot features of each university), we use Pearson’s r coefficient. We find that the women's colleges have at least an absolute correlation of $0.273$ across Proxy conditions, whereas the maximum absolute correlation for other continuous features is $0.014$, which is much lower. Similarly, for categorical features, we use Cramer's V, finding that the university feature has at least an absolute correlation of $0.417$ while the maximum absolute correlation for the remaining categorical features is $0.082$, which is also lower. Overall, we see that university (especially women's colleges) has a much stronger correlation with gender than any other feature shown to the participants.

\paragraph{Models}
For our AI predictions, we use logistic regression models as explanations on simple models may be more useful to humans~\cite{lai2020chicago}. We train the models on $14$ pre-selected features from the Prosper dataset (of which participants will only see 5) and, when applicable, the gender or university feature. These models have an average accuracy of about $65\%$ when compared to the original ground-truth values before adding synthetic features. %
Since we are using logistic regression, we can create a simple input-influence explanation of the AIs' predictions using feature weights. For continuous features like $\texttt{LoanOriginalAmount}$, we multiply the normalized feature value by the corresponding feature weight. For categorical features like $\texttt{EmploymentStatus}$, we take only the feature weight corresponding to the feature value (e.g., the weight of the $\texttt{EmploymentStatus} = \texttt{Full-Time}$ feature). These values are graphed as in~\autoref{fig:example} (left).

\subsection{Metrics}\label{sec:metrics}
We evaluate study outcomes based on participants' perceptions and decisions. We consider two fairness perception metrics based on questions in the post-phase surveys, and we consider the decisions made by the human-AI teams based on one fairness measure---demographic parity---and three decision quality metrics: accuracy, false negative rate, and false positive rate. 

We measure all metrics in the two task phases across conditions.
We count both ``Likely Complete'' and ``Definitely Complete'' as ``Complete'' and similarly for ``Late''. 
We count ``Neutral'' as agreement with the system prediction.

\subsubsection{Decision-Making Fairness Measure}\label{sec:dm fairness metric}

We employ demographic parity~\citep[i.a.]{feldman2015certifying} as a measure of fairness in decision-making, which captures the independence between protected characteristics and prediction. There are other measures of fairness~\cite{narayanan2018translation}, however, not all definitions can be simultaneously satisfied~\cite{chouldechova2017fair,kleinberginherent2017}. Demographic parity has been found to be more understandable to laypeople and better capture their perception of fairness than competing metrics~\cite{saha2020measuring, srivastava2019mathematical}.
We can calculate the demographic parity for human-AI teams in task phases 1 and 2 across conditions as follows.
$$\text{Parity} = \frac{\mathbb{E}[\hat{Y}_{i,j}=1 \mid \text{Gender}=\texttt{female}]}{\mathbb{E}[\hat{Y}_{i,j}=1 \mid\text{Gender}=\texttt{male}]},$$
where $\hat{Y}_{i,j}$ is the predicted decision for the applicant by the participant $j$.
We obtain one demographic parity score in this way for each participant's decisions in each phase. 

A parity close to $1$ means an equal acceptance rate. As the acceptance rate for the advantaged group increases over the disadvantaged group, parity becomes closer to $0$. If the acceptance rate of the disadvantaged group increases above the advantaged group, then the parity can increase above $1$. A parity of less than $\frac{4}{5}$ is considered ``evidence of adverse impact'' under US Anti-Discrimination law~\cite{four-fifths}.
In our model bias disclosure, we tell the participants about this $80\%$ rule and that the model failed this test in phase 1, that is, the demographic parity of the model is below $80\%$ (\autoref{fig:bias_disclosure_in_body}). 

\subsubsection{Fairness Perception Measures}\label{subsec:fairness perception}
Based on our post-task surveys (described in \aptLtoX[graphic=no,type=html]{\S\ref{sec:surveys}}{\autoref{sec:surveys}}), we calculate two measures of how participants perceive the degree of model unfairness. First, we consider how much participants agree with the statement ``The AI model was fair across different genders''. Here, the participant's \textit{fairness rating} is higher when they believe the model is more fair. We also consider whether participants mark ``unfairness'' as a reason that they disagreed with model decisions. Here, the participant's \textit{fairness saliency} is higher when they have a greater belief that they disagreed with the model due to unfairness.

\subsubsection{Decision Quality Measures}\label{sec:DQM}

Measures such as accuracy require $Y_i$'s: a ground-truth ``Complete'' or ``Late'' value to compute. We have access to the ground-truth loan completion status for the original applicants. However, as we discuss in \aptLtoX[graphic=no,type=html]{\S\ref{sec:data}}{\autoref{sec:data}}, our study uses an edited set of applicants with synthetic gender and university features which are made to be correlated with the outcome. 
We estimate the loan completion status for the edited profile from the ground-truth completion status of the original applicants and our defined sampling rates of the synthetic features using Bayes' rule. In turn, we compute an expected accuracy, expected FPR, and expected FNR using the estimated loan completion status as our decision-quality measures. 
See \aptLtoX[graphic=no,type=html]{Appendix~\ref{sec:deriving_ground_truth}}{\autoref{sec:deriving_ground_truth}} for more details. %

\begin{figure*}[t]
    \centering
    \includegraphics[height=12pt]{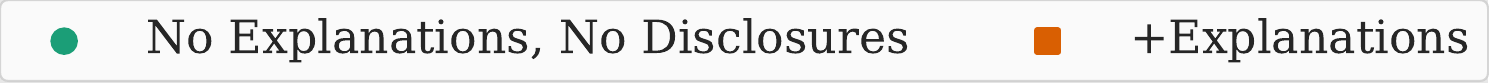}
    \begin{tikzpicture}[]
        \matrix (m) [
            matrix of nodes, 
            column sep = 5mm,
            row sep = 0mm,
            nodes={  minimum width=.3\textwidth, text width=.29\textwidth, line width=0mm, name=table}
        ] {
             \node(1){\includegraphics[width=\textwidth]{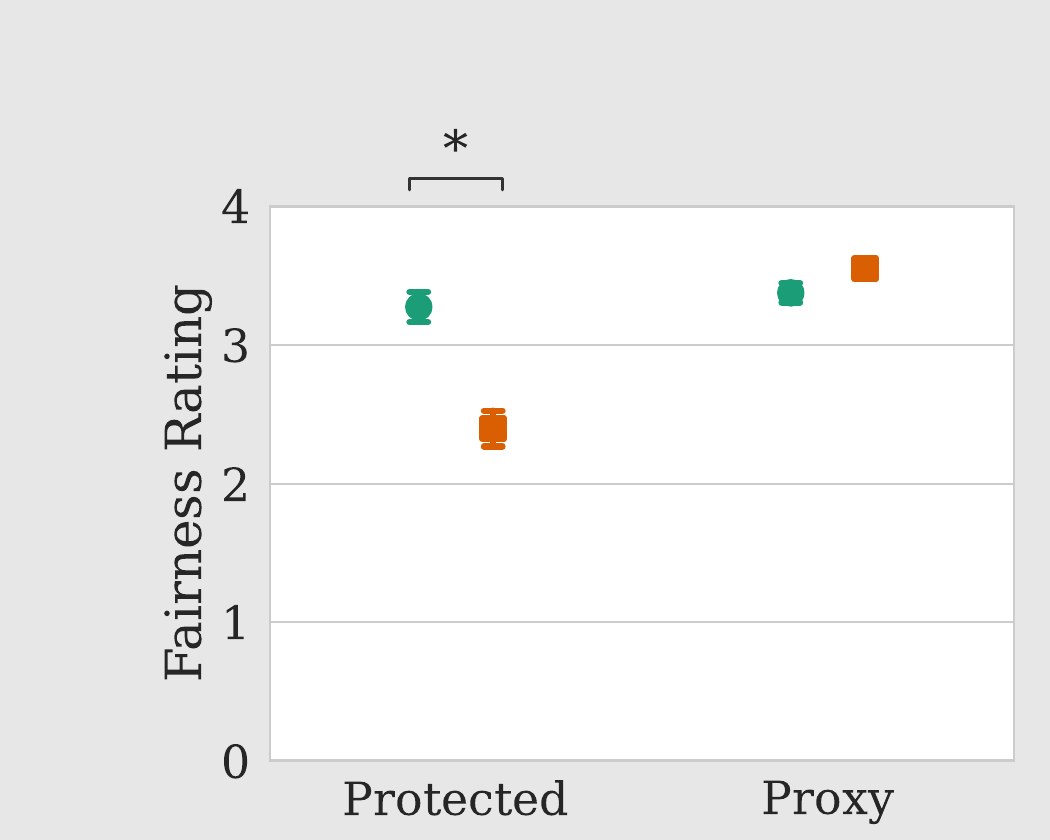}}; &
             \node(2){\includegraphics[width=\textwidth]{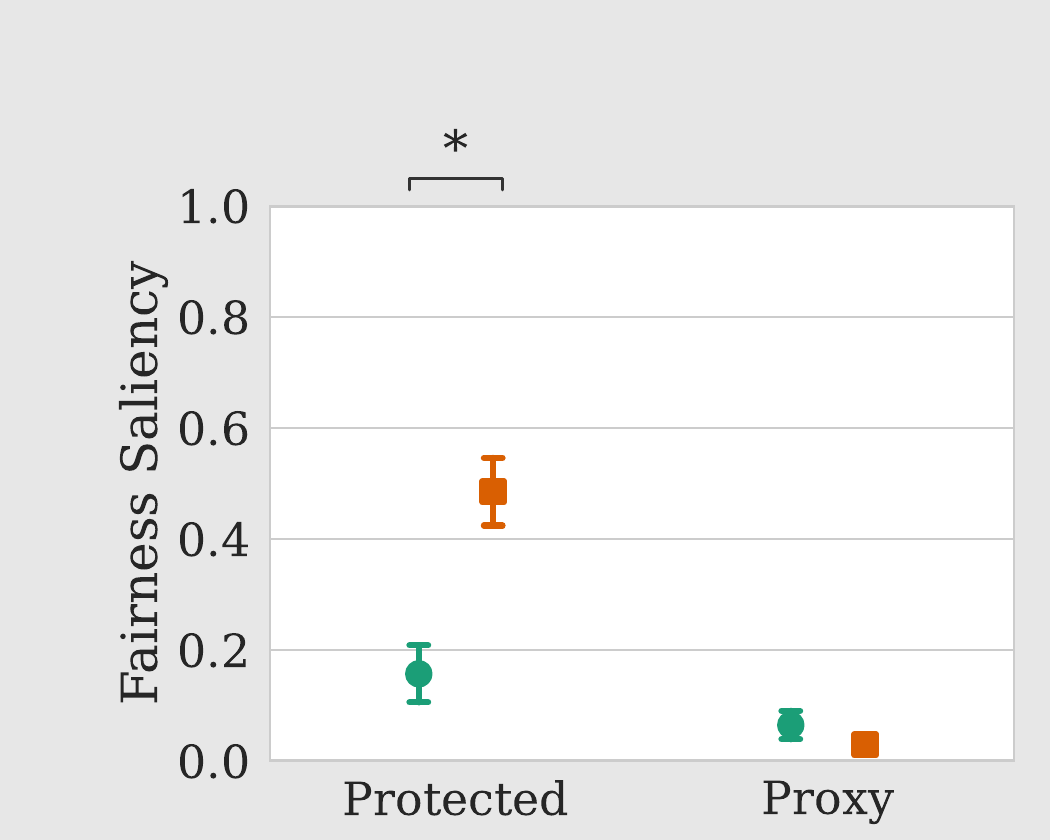}};&
             \node(3){\includegraphics[width=\textwidth]{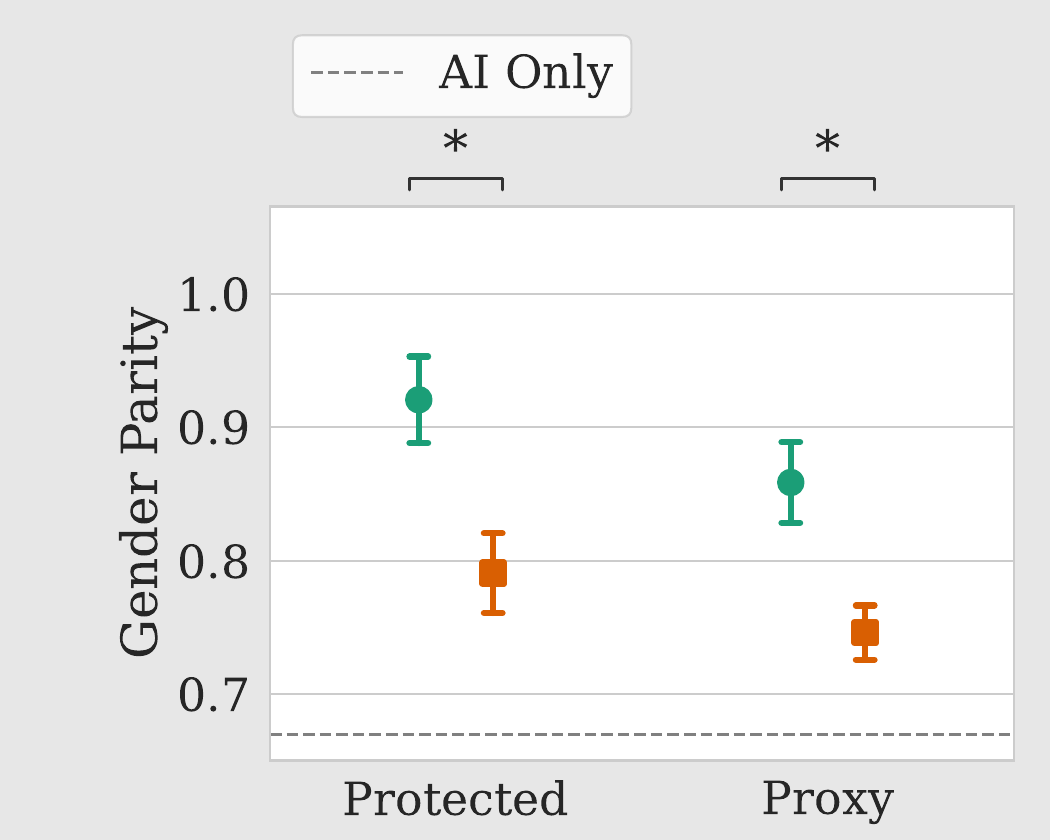}};\\
             \node(x){};&
             \node(y){};&
             \node(z){};\\ [5mm]
             \node(4){\includegraphics[width=\textwidth]{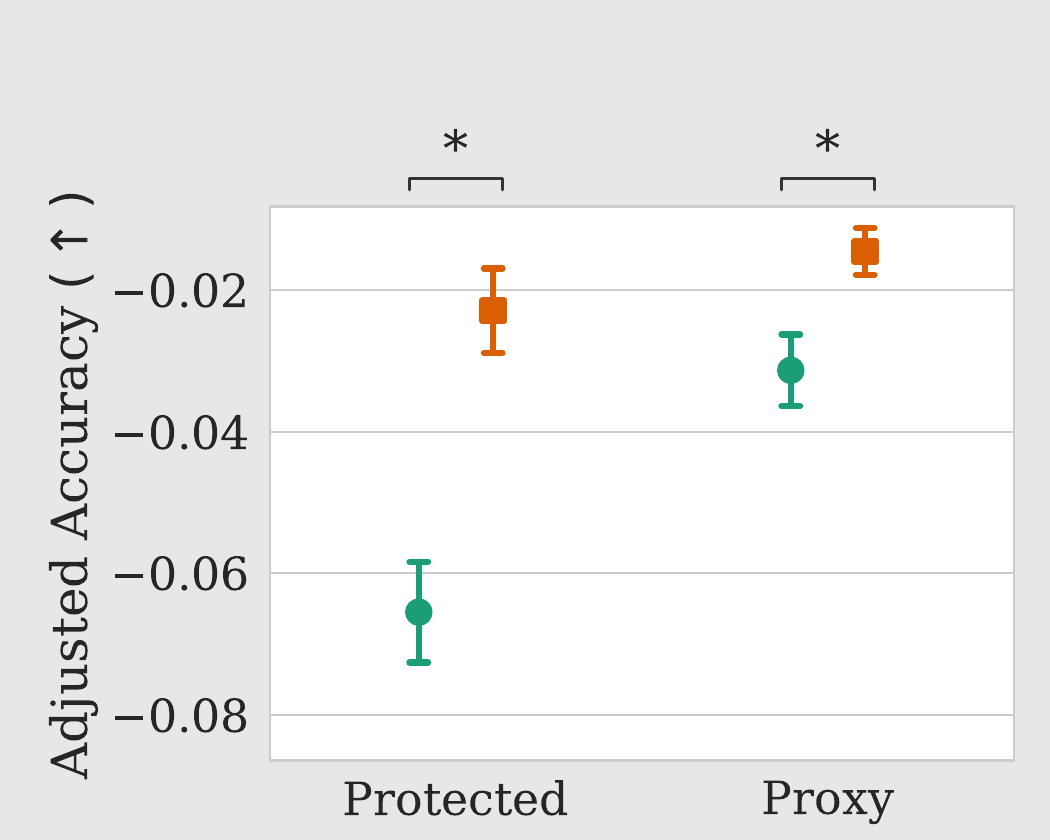}}; &
             \node(5){\includegraphics[width=\textwidth]{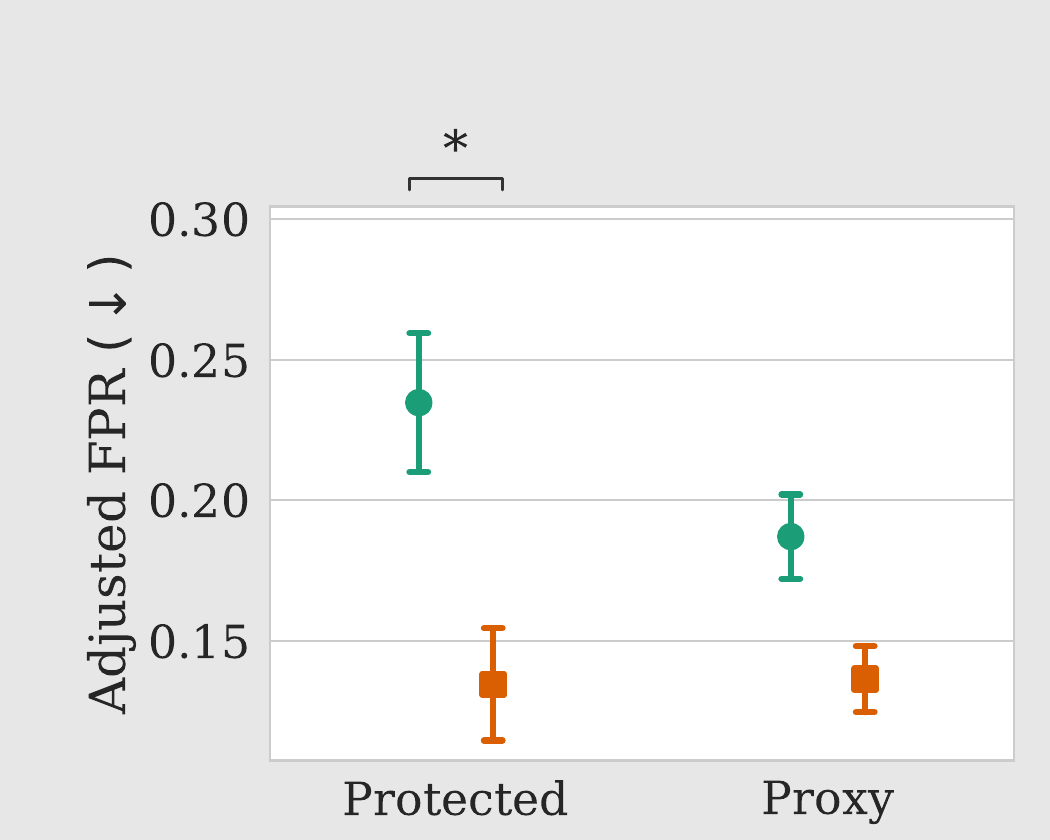}};&
             \node(6){\includegraphics[width=\textwidth]{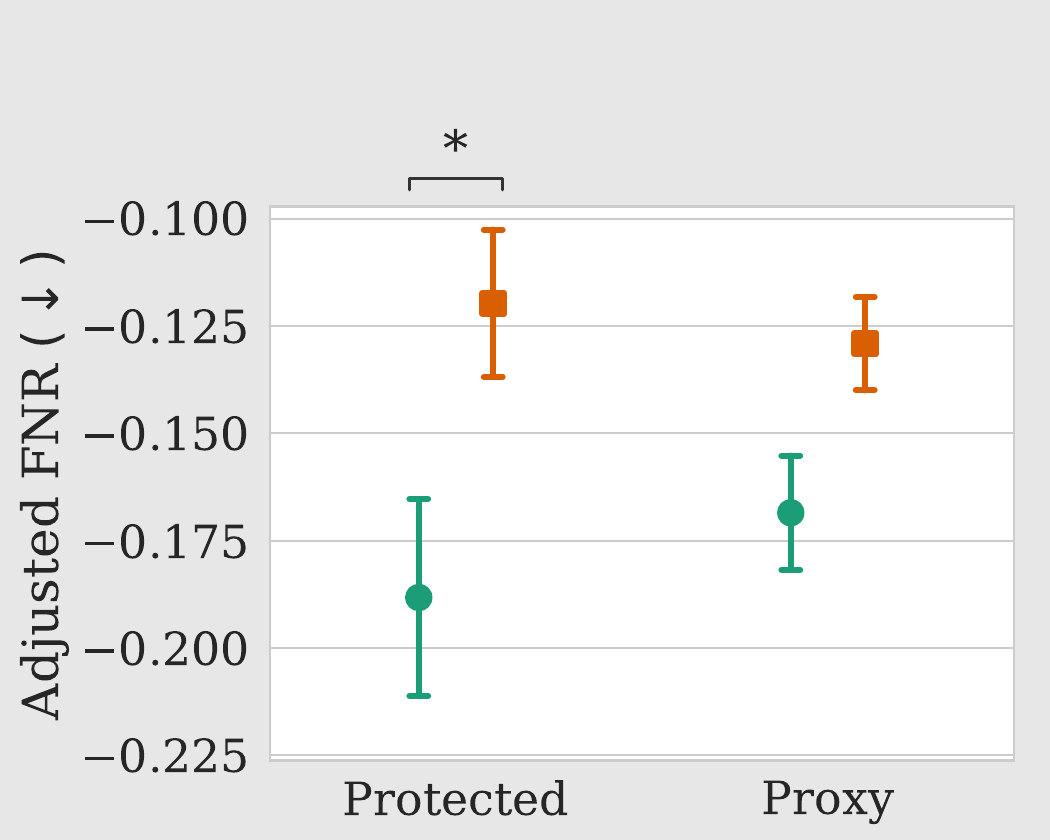}};\\
             \node(xx){};&
             \node(yy){};&
             \node(zz){};\\
        };

        \node(a)[fit={(x)(y)}]{\subcaption{Fairness Perceptions (\S \ref{subsec:fairness perception})\label{subfig:expl_fp}}};
        \node(b)[fit={(z)}]{\subcaption{Decision-Making Fairness (\S \ref{sec:dm fairness metric})\label{subfig:expl_dmf}}};
        \node(c)[fit={(xx)(yy)(zz)}]{\subcaption{Decision-Making Quality (\S \ref{sec:DQM})\label{subfig:expl_dmq}}};;

        \begin{scope}[on background layer]
        \node(i) [fill=customrowcolor, fit={(1)(2)(a)(x)(y)}]{};
        \node(j) [fill=customrowcolor, fit={(3)(b)(z)}]{};
        \node(k) [fill=customrowcolor, fit={(4)(5)(6)(c)(xx)(yy)(zz)}]{};
        \end{scope}
    \end{tikzpicture}
    \caption{Effect of explanations alone on various metrics when bias stems from usage of a protected vs proxy feature. The marks show the average and standard error of the given metric across participants in the given condition.}
    \label{fig:expl}
\end{figure*}

\subsection{Statistical Analyses}\label{sec:stats}

To answer our research questions (\aptLtoX[graphic=no,type=html]{\S\ref{sec:rqs}}{\autoref{sec:rqs}}), we perform separate multi-way ANOVA tests for different treatments (explanations, disclosures without explanations, and disclosures with explanations) for both protected and proxy conditions. 
For each statistical test, we construct a linear model with a fixed effect term for each independent treatment variable and one fixed effect term representing the participant's dispositional trust, which is calculated by averaging the scores from the pre-study trust survey. 
Additionally, in the within study comparisons (that is, moving from phase 1 to phase 2), we include the participant ID as a random effect. 

The independent treatment variables are determined by the factors that vary between the effect of interest. For instance, to estimate the effect of explanation (that is, the vertical arrow in \autoref{fig:expl_disclosure}), the treatment variable is the \textit{presence of explanations}. For the effect of disclosure (that is, the horizontal arrows in \autoref{fig:expl_disclosure}), the treatment variables are: (1) whether only \textit{bias disclosure} has been shown (i.e., is this a phase 2 measurement with no correlation disclosure), and (2) whether full \textit{bias and correlation disclosure} has been shown. For the effect of adding both explanations and disclosures (that is, the diagonal in \autoref{fig:expl_disclosure} going from without explanation and disclosures in phase 1 to with explanation and disclosures in phase 2), we include all three treatment variables.

In each ANOVA test, we consider the data from the relevant sections. For instance, to estimate the effect of explanations alone in the case of direct bias through a protected feature, we only consider the data in phase 1 of the ``protected'' conditions (left vertical section in \autoref{fig:expl_disclosure}), and similarly for ``proxy'' conditions.

We fit a separate model for each fairness perception and decision-making metric as the dependent variable for each of the above effects. Although the ratio of ``Complete'' and ``Late'' model decisions shown to each participant is kept the same in each phase across conditions (leading to a constant AI-only parity of $\sim$$0.67$), the model accuracy, FPR, and FNR vary across conditions. To account for this variation, we subtract the model score from the score of the human-AI team. Similarly, in assessing learned trust measures, we adjust for dispositional trust in AI by subtracting the participant's response to the corresponding question in the pre-study survey.

In addition to our key metrics detailed in \aptLtoX[graphic=no,type=html]{\S\ref{sec:metrics}}{\autoref{sec:metrics}}, we also study the effect of the treatments on learned trust. We follow the same procedure as before but with the learned trust measures as the dependent variable in the ANOVA test. In this case, however, we adjust participants' post-phase 1 or post-phase 2 survey responses based on their baseline responses, and we do not include the overall dispositional trust term. Lastly, we perform additional ANOVA tests to analyze the difference between dispositional and learned trust. For this, we use the participants' trust ratings in the pre-study survey and surveys after phase 1 or phase 2 as the dependent variable. We fit two linear models, one for each phase, testing whether the phase has a fixed effect on the participants' trust ratings in different conditions (with the participant added as a random effect). 

We perform Benjamini-Hochberg correction to avoid multiple testing effect with a false discovery threshold of $0.05$~\cite{1995benjaminicontrolling}. This leads to a significance threshold of $0.0175$ for the reported results. 

\section{Quantitative Results}\label{sec:results}
In this section, we report our findings on the effects of different interventions on the decision-making and fairness perception metrics. We first discuss the primary effects detailed in \autoref{fig:expl_disclosure}: the effect of explanations (\aptLtoX[graphic=no,type=html]{\S\ref{sec:expl}}{\autoref{sec:expl}}), the effect of disclosures without explanations (\aptLtoX[graphic=no,type=html]{\S\ref{sec:disc}}{\autoref{sec:disc}}) and the effect of disclosures with explanations (\aptLtoX[graphic=no,type=html]{\S\ref{sec:disc_expl}}{\autoref{sec:disc_expl}}). Next, we examine the effect of the joint intervention of adding both explanations and disclosures (\aptLtoX[graphic=no,type=html]{\S\ref{sec:kitchen sink}}{\autoref{sec:kitchen sink}}). Lastly, we discuss the effect of dispositional trust on the decision-making and fairness perception, the effect of different interventions on participants' trust, and the differences in participants' dispositional trust vs learned trust in \aptLtoX[graphic=no,type=html]{\S\ref{subsec:additional_results}}{\autoref{subsec:additional_results}} (full results in \aptLtoX[graphic=no,type=html]{Appendix~\ref{sec:result_overflow}}{\autoref{sec:result_overflow}}). 


\begin{figure*}[t]
    \centering
    \includegraphics[height=12pt]{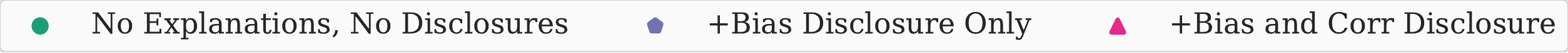}
    \begin{tikzpicture}[]
        \matrix (m) [
            matrix of nodes, 
            column sep = 5mm,
            row sep = 0mm,
            nodes={  minimum width=.3\textwidth, text width=.29\textwidth, line width=0mm, name=table}
        ] {
             \node(1){\includegraphics[width=\textwidth]{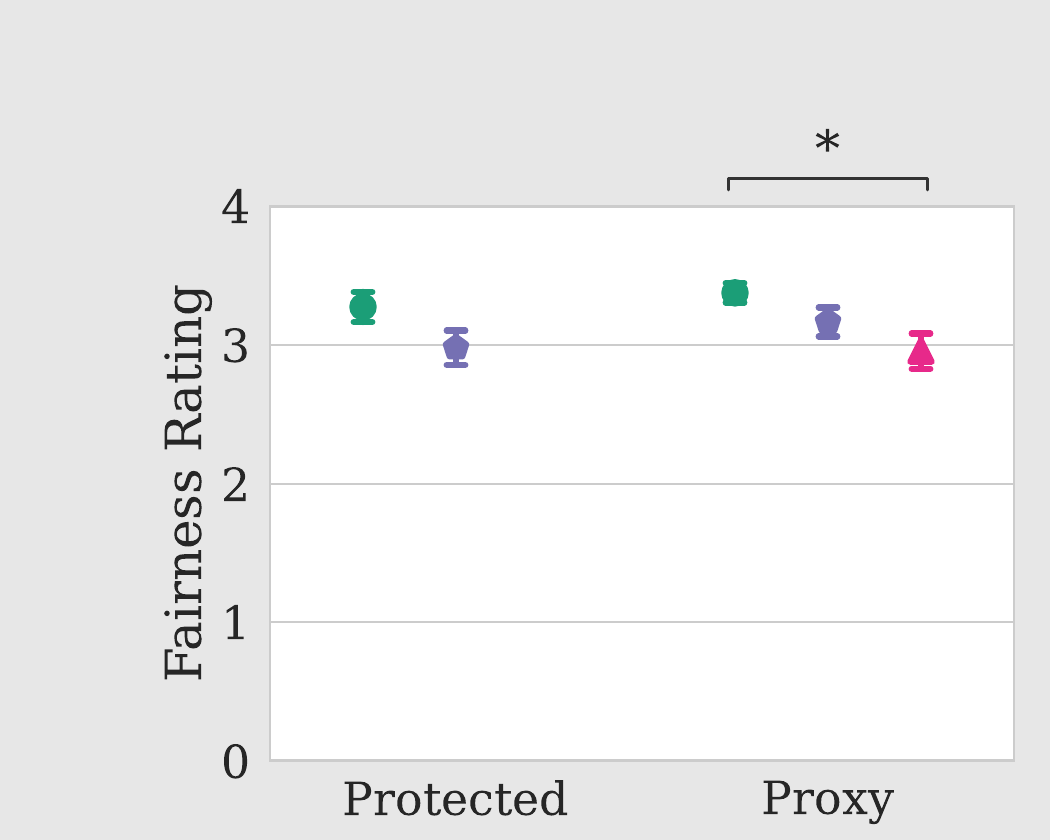}}; &
             \node(2){\includegraphics[width=\textwidth]{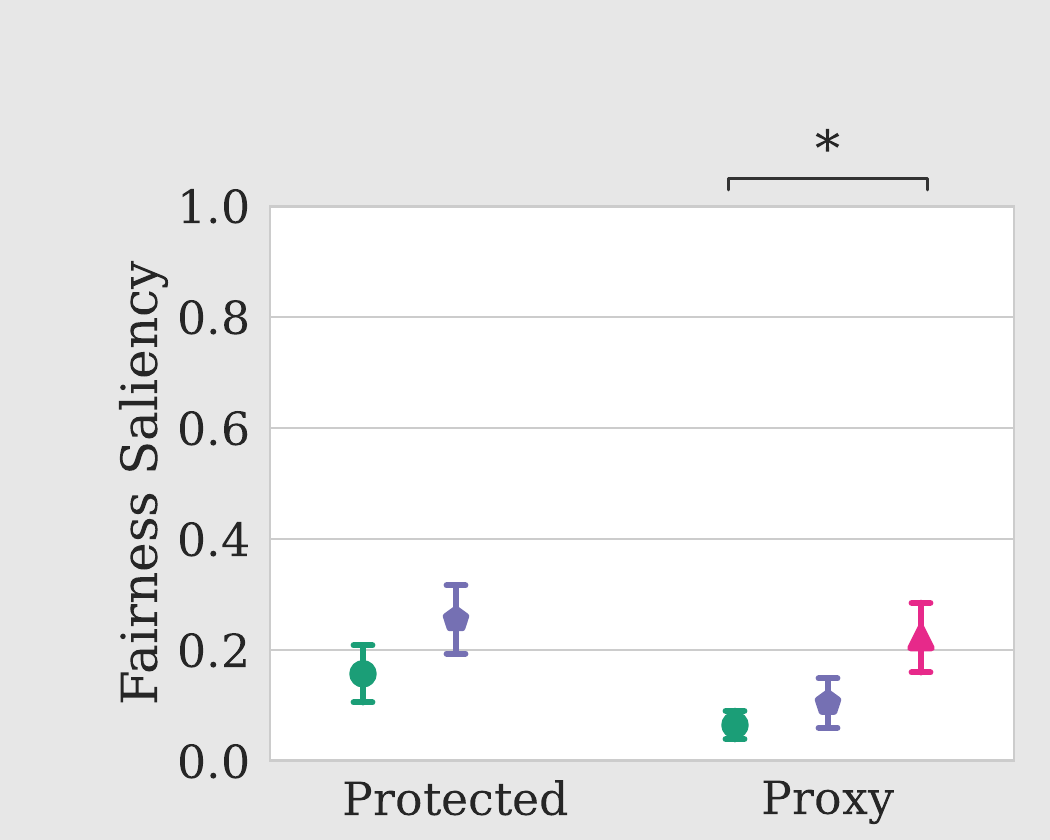}};&
             \node(3){\includegraphics[width=\textwidth]{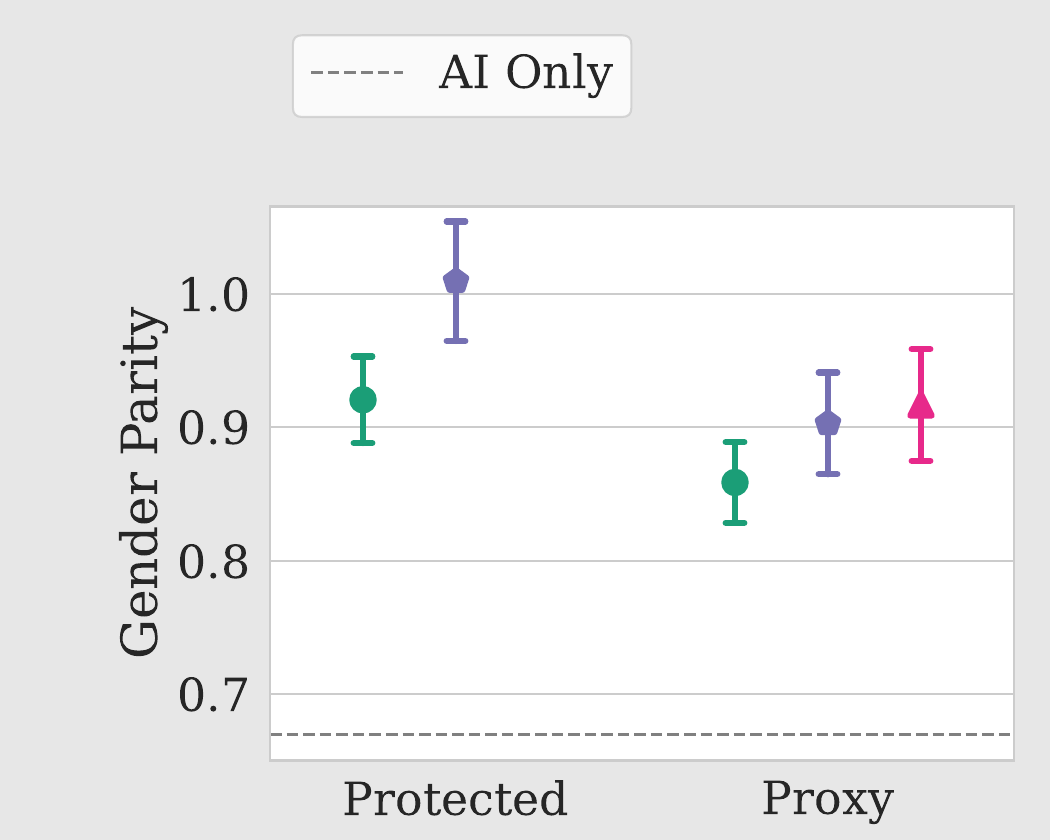}};\\
             \node(x){};&
             \node(y){};&
             \node(z){};\\ [5mm]
             \node(4){\includegraphics[width=\textwidth]{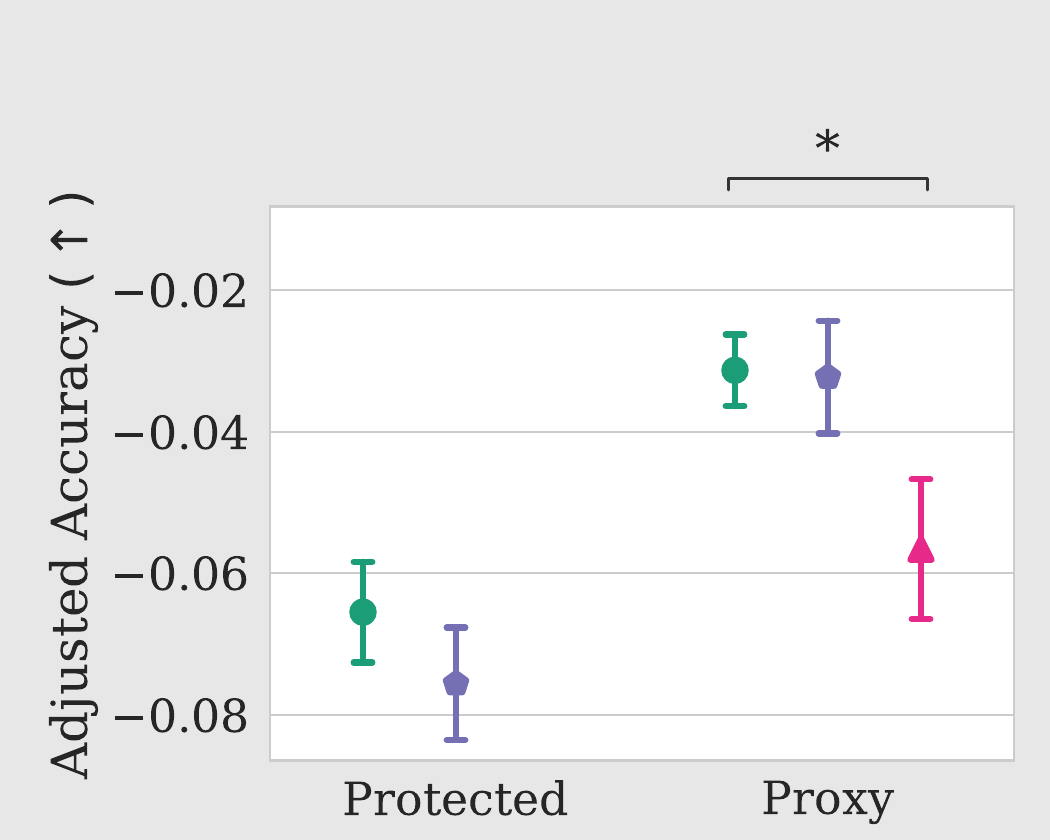}}; &
             \node(5){\includegraphics[width=\textwidth]{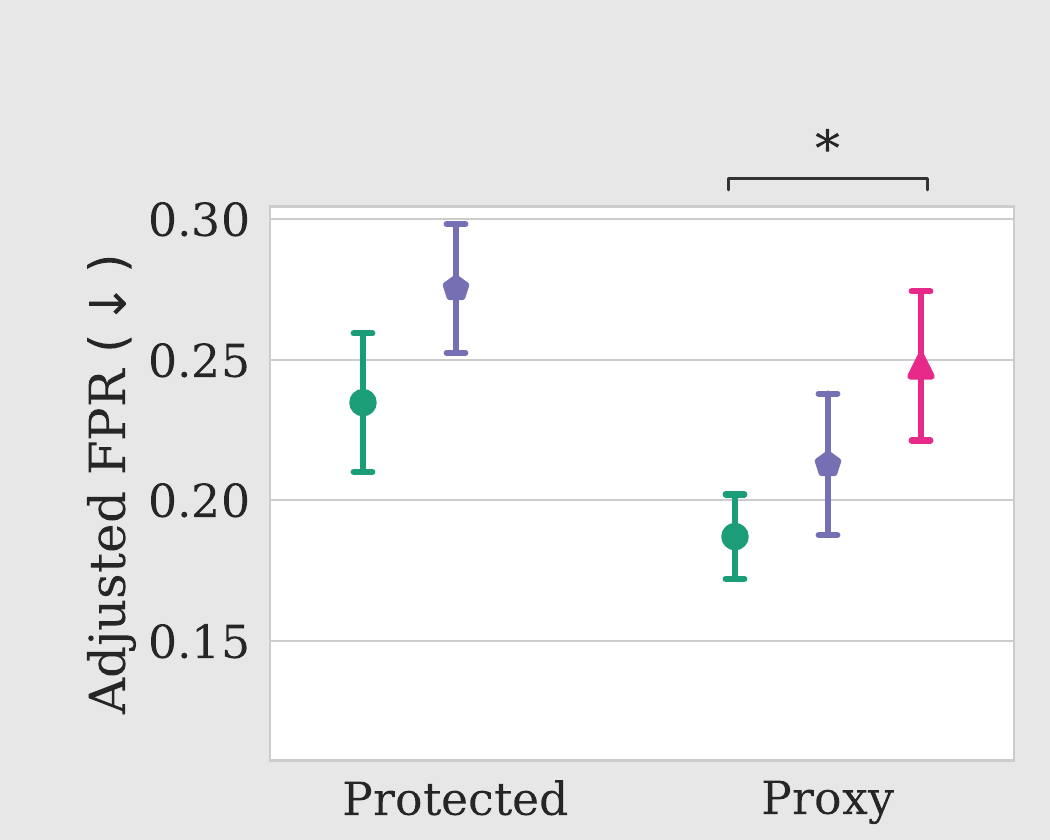}};&
             \node(6){\includegraphics[width=\textwidth]{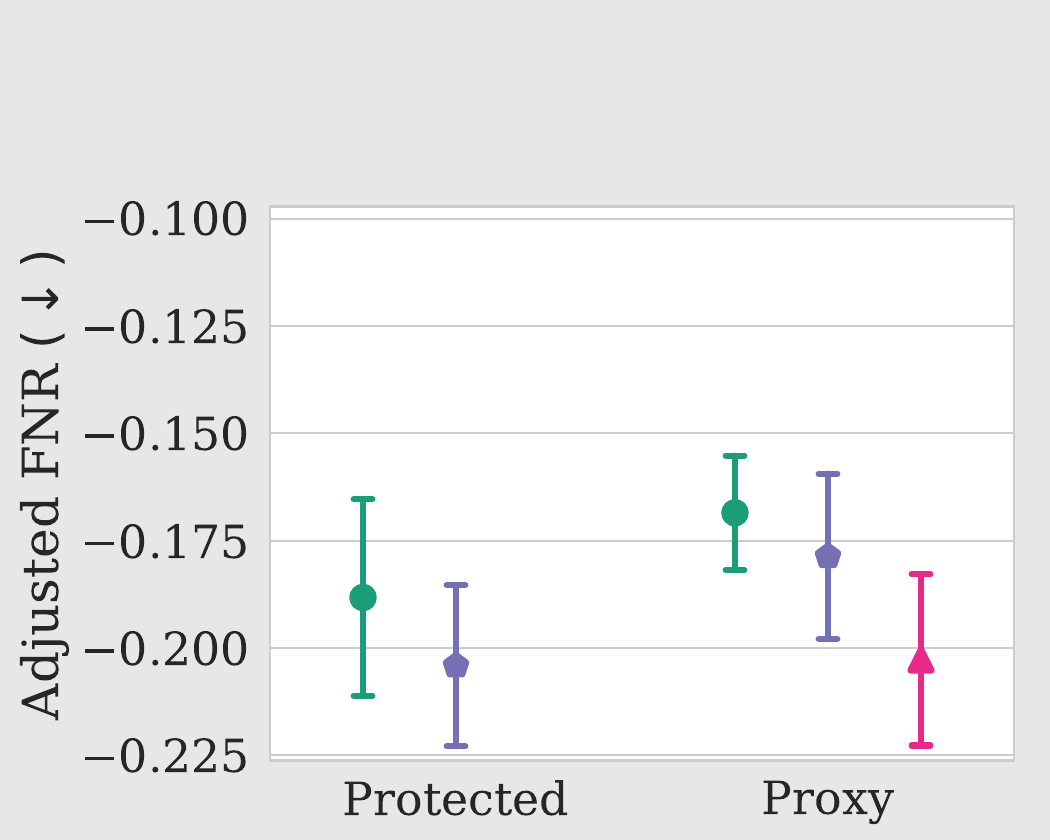}};\\
             \node(xx){};&
             \node(yy){};&
             \node(zz){};\\
        };

        \node(a)[fit={(x)(y)}]{\subcaption{Fairness Perceptions (\S \ref{subsec:fairness perception})\label{subfig:disc_fp}}};
        \node(b)[fit={(z)}]{\subcaption{Decision-Making Fairness (\S \ref{sec:dm fairness metric})\label{subfig:disc_dmf}}};
        \node(c)[fit={(xx)(yy)(zz)}]{\subcaption{Decision-Making Quality (\S \ref{sec:DQM})\label{subfig:disc_dmq}}};;

        \begin{scope}[on background layer]
        \node(i) [fill=customrowcolor, fit={(1)(2)(a)(x)(y)}]{};
        \node(j) [fill=customrowcolor, fit={(3)(b)(z)}]{};
        \node(k) [fill=customrowcolor, fit={(4)(5)(6)(c)(xx)(yy)(zz)}]{};
        \end{scope}
    \end{tikzpicture}
    \caption{Effect of disclosures \textit{without} explanations on various metrics when bias stems from usage of a protected vs proxy feature. The marks show the average and standard error of the given metric across participants in the given condition.}
    \label{fig:disc}
\end{figure*}

\subsection{Effect of Explanations Alone}\label{sec:expl}
First, we consider the effects of explanations alone by comparing the first phase of with and without explanations conditions with either type of bias. 

In the case of direct bias through a protected feature, we find that explanations alone have a significant effect on all metrics; however, the direction of the effect is not consistent. Explanations alone significantly improve participants' ability to recognize unfairness (\aptLtoX[graphic=no,type=html]{Figure~\ref{subfig:expl_fp}}{\autoref{subfig:expl_fp}}). Surprisingly, despite participants being more able to recognize that the model is unfair when shown explanations, when considering decisions instead of perceptions, we see that explanations significantly decrease gender parity (\aptLtoX[graphic=no,type=html]{Figure~\ref{subfig:expl_dmf}}{\autoref{subfig:expl_dmf}}).
Looking closer, we find that explanations lead to a significantly lower acceptance rate for female applicants, whereas the acceptance rate for male applicants does not change significantly (\aptLtoX[graphic=no,type=html]{Table~\ref{tab:flip_rate}}{\autoref{tab:flip_rate}} in the Appendix). 
This decrease in acceptance rates also leads to a significant increase in the FNR and a significant decrease in FPR, with an overall higher accuracy (\aptLtoX[graphic=no,type=html]{Figure~\ref{subfig:expl_dmq}}{\autoref{subfig:expl_dmq}}). 

In the case of indirect bias through a proxy feature, we find that explanations alone significantly reduce gender parity, similar to the case of direct bias (\aptLtoX[graphic=no,type=html]{Figure~\ref{subfig:expl_dmf}}{\autoref{subfig:expl_dmf}}). Analogous to direct bias, this occurs due to a significant decrease in acceptance of female applicants (\aptLtoX[graphic=no,type=html]{Table~\ref{tab:flip_rate}}{\autoref{tab:flip_rate}} in the Appendix). Further, explanations also lead to a significant increase in accuracy in the case of indirect bias. However, unlike with direct bias, explanations have no significant effect on fairness perceptions in the case of indirect bias (\aptLtoX[graphic=no,type=html]{Figure~\ref{subfig:expl_fp}}{\autoref{subfig:expl_fp}}). 

Overall, we find that explanations can help people recognize unfairness in the case of direct bias but not indirect. This is in line with our intuition that indirect biases are harder for participants to notice. 
However, regardless of fairness perceptions, in line \citet{wang2023effects}, we find that explanations lead people to accept model biases leading to less fair decisions. This could be attributed to the presence of explanations assisting humans in rationalizing AI's unfair predictions rather than challenging them.

\subsection{Effect of Disclosures without Explanations}\label{sec:disc}

We consider the effects of disclosures without explanations by comparing between phase 1 and phase 2 in without explanations conditions with either type of bias. 

For both direct and indirect bias, we find that disclosing model bias alone does not have a significant effect on any of the outcome metrics (gender parity, fairness perception, accuracy, FPR, and FNR). 

However, in the case of indirect bias, when we disclose both the model bias and the relationship between the protected and proxy feature (i.e., that some universities in the study are women's colleges), participants were significantly more likely to report that the model is unfair or that this unfairness caused them to disagree with the model's decisions (\aptLtoX[graphic=no,type=html]{Figure~\ref{subfig:disc_fp}}{\autoref{subfig:disc_fp})}. Interestingly, this still does not translate to fairer decisions---as seen in \aptLtoX[graphic=no,type=html]{Figure~\ref{subfig:disc_dmf}}{\autoref{subfig:disc_dmf}}, the gender parity does not change significantly on disclosing both model bias and correlations in the case of indirect bias.

In sum, we find that, interestingly, being explicitly told that the model is biased does not affect participants' fairness perception of the model decisions (in both direct and indirect bias conditions). 
In the direct bias condition, this could be because the model is perceived as unfair even pre-disclosures. In the indirect bias condition, this might be because disclosures alone, without explanations, are insufficient for participants to fully acknowledge the bias in the model's predictions. 
But, disclosing both the model bias and the correlation between protected and proxy features does lead to participants perceiving the model as less fair in the case of indirect bias. However, this is not sufficient to improve decision-making fairness. This may be because, although disclosures assist participants in recognizing the unfairness of model predictions, they still lack sufficient information to overturn individual predictions without additional guidance on how the model utilizes the correlation between protected and proxy features.

\begin{figure*}[t]
    \centering
    \includegraphics[height=12pt]{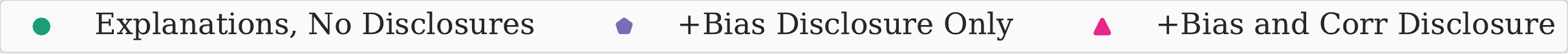}
    \begin{tikzpicture}[]
        \matrix (m) [
            matrix of nodes, 
            column sep = 5mm,
            row sep = 0mm,
            nodes={  minimum width=.3\textwidth, text width=.29\textwidth, line width=0mm, name=table}
        ] {
             \node(1){\includegraphics[width=\textwidth]{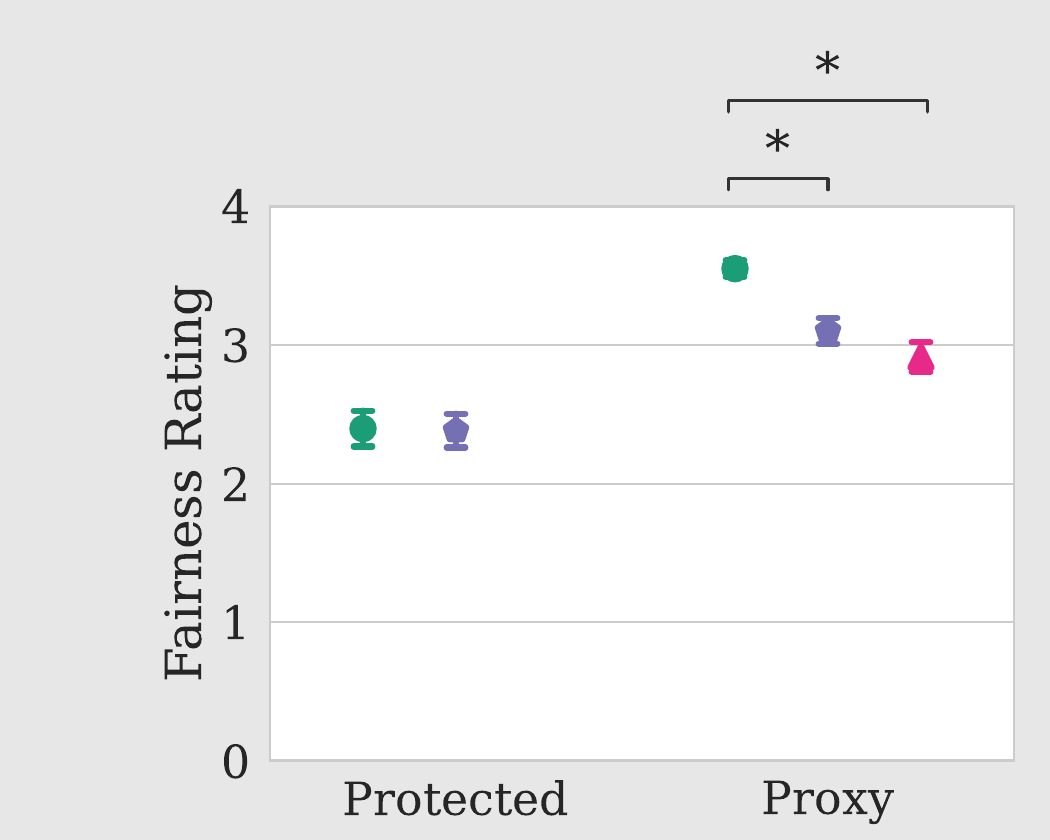}}; &
             \node(2){\includegraphics[width=\textwidth]{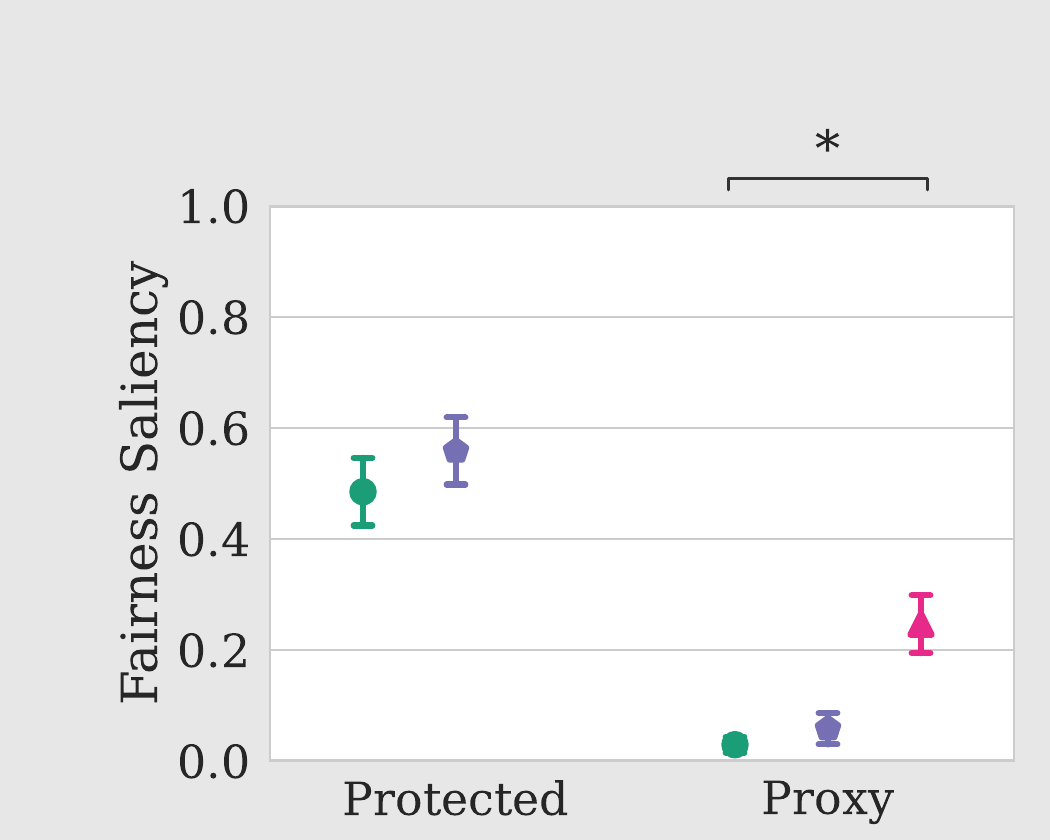}};&
             \node(3){\includegraphics[width=\textwidth]{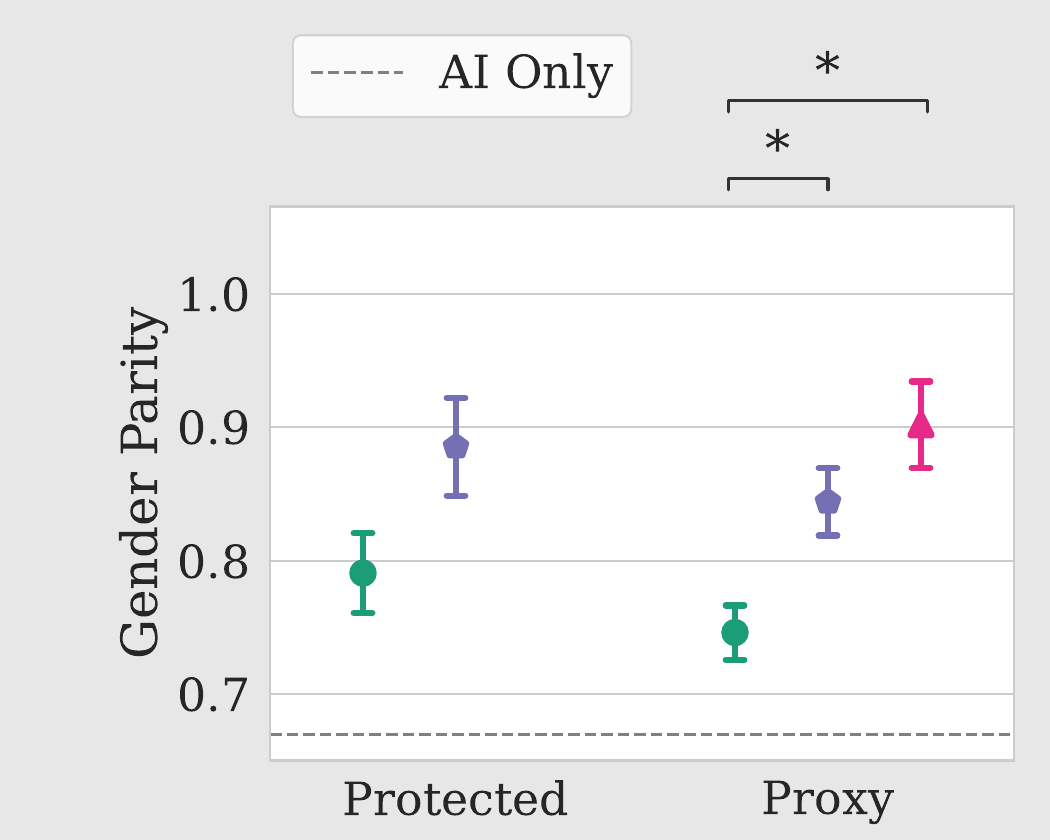}};\\
             \node(x){};&
             \node(y){};&
             \node(z){};\\ [5mm]
             \node(4){\includegraphics[width=\textwidth]{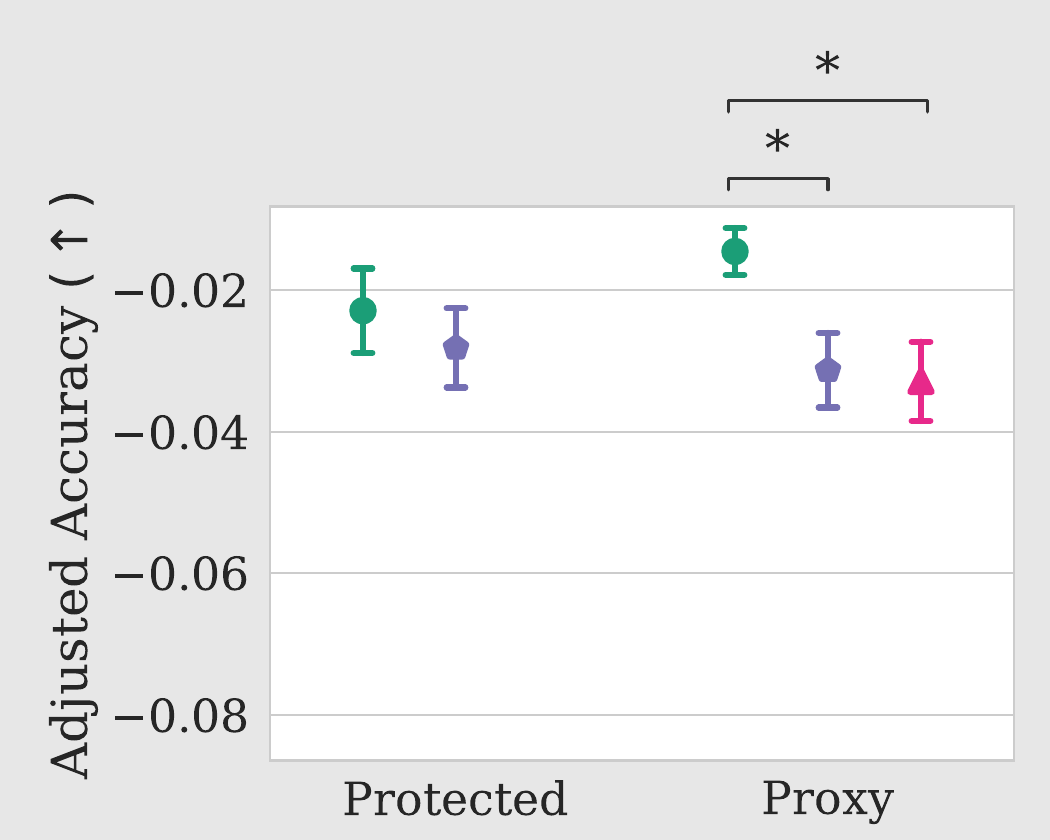}}; &
             \node(5){\includegraphics[width=\textwidth]{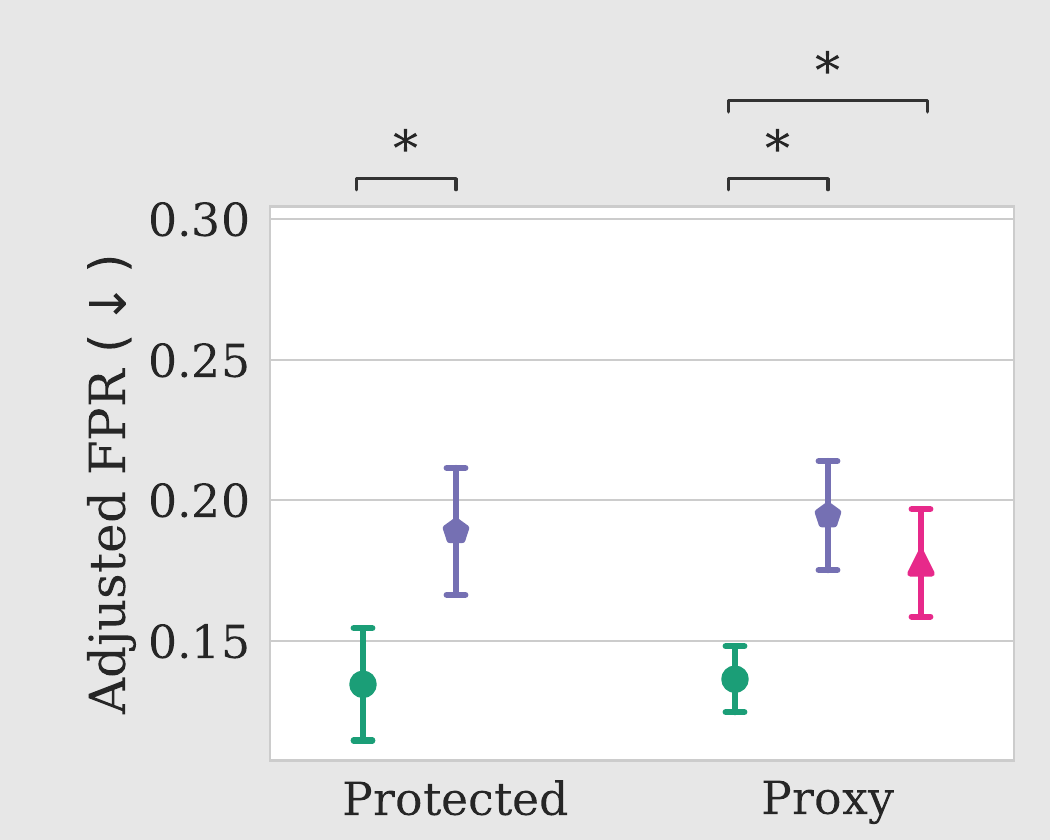}};&
             \node(6){\includegraphics[width=\textwidth]{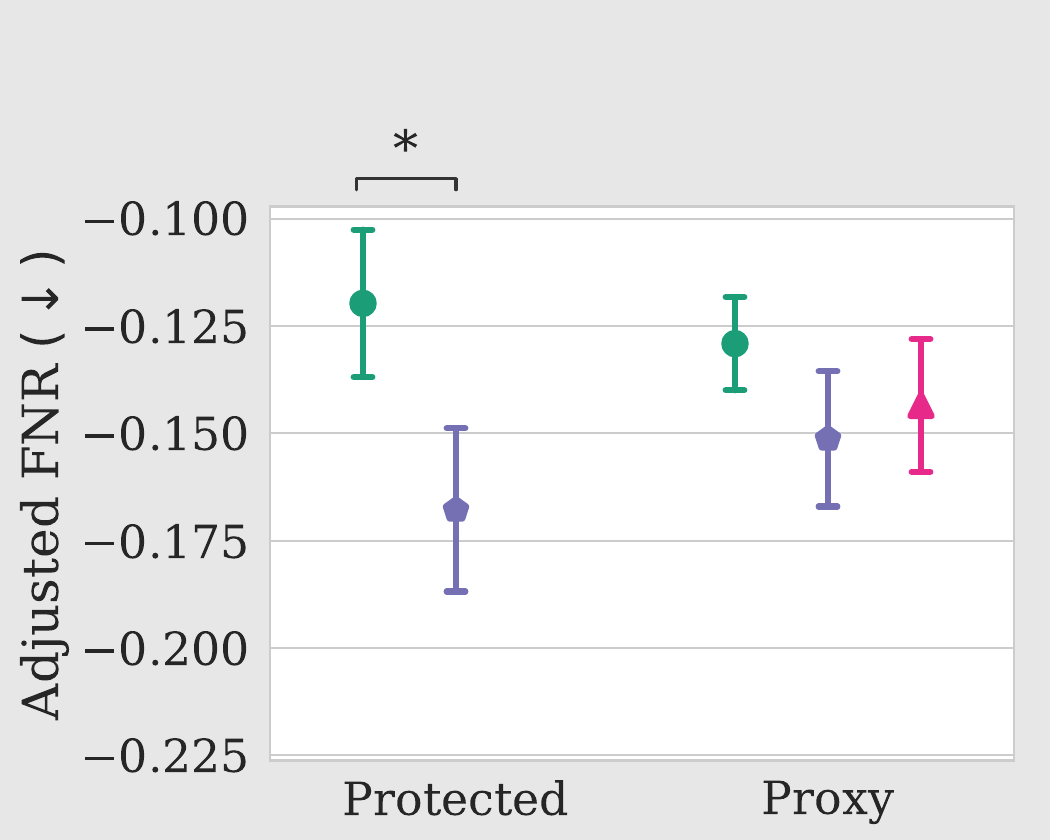}};\\
             \node(xx){};&
             \node(yy){};&
             \node(zz){};\\
        };

        \node(a)[fit={(x)(y)}]{\subcaption{Fairness Perceptions (\S \ref{subsec:fairness perception})\label{subfig:disc_expl_fp}}};
        \node(b)[fit={(z)}]{\subcaption{Decision-Making Fairness (\S \ref{sec:dm fairness metric})\label{subfig:disc_expl_dmf}}};
        \node(c)[fit={(xx)(yy)(zz)}]{\subcaption{Decision-Making Quality (\S \ref{sec:DQM})\label{subfig:disc_expl_dmq}}};;

        \begin{scope}[on background layer]
        \node(i) [fill=customrowcolor, fit={(1)(2)(a)(x)(y)}]{};
        \node(j) [fill=customrowcolor, fit={(3)(b)(z)}]{};
        \node(k) [fill=customrowcolor, fit={(4)(5)(6)(c)(xx)(yy)(zz)}]{};
        \end{scope}
    \end{tikzpicture}
    \caption{Effect of disclosures \textit{with} explanations on various metrics when bias stems from usage of a protected vs proxy feature. The marks show the average and standard error of the given metric across participants in the given condition.}
    \label{fig:disc_expl}
\end{figure*}

\subsection{Effect of Disclosures with Explanations}\label{sec:disc_expl}

We consider the effects of disclosures with explanations by comparing between phase 1 and phase 2 in explanations conditions with either type of bias.

In the case of direct bias through a protected feature, we find that bias disclosure with explanations has no significant effect on fairness perceptions (\aptLtoX[graphic=no,type=html]{Figure~\ref{subfig:disc_expl_fp}}{\autoref{subfig:disc_expl_fp}}).
Unlike with bias disclosure without explanations (\aptLtoX[graphic=no,type=html]{\S\ref{sec:disc}}{\autoref{sec:disc}}), we find that bias disclosure with explanations significantly increases the acceptance rate for female applicants
(participants flip models' ``Late'' predictions for female applicants at a much higher rate), with the acceptance rate for the male applicants unchanged (\aptLtoX[graphic=no,type=html]{Table~\ref{tab:flip_rate}}{\autoref{tab:flip_rate}} in the Appendix). 
Bias disclosure with explanations also results in a significant increase in FPR and a significant decrease in FNR, leading to an overall insignificant change in accuracy (\aptLtoX[graphic=no,type=html]{Figure~\ref{subfig:disc_expl_dmq}}{\autoref{subfig:disc_expl_dmq}}). Even despite a higher acceptance rate for female applicants, the increase in gender parity is not significant (\aptLtoX[graphic=no,type=html]{Figure~\ref{subfig:disc_expl_dmf}}{\autoref{subfig:disc_expl_dmf}}), likely due to the normalizing effect of the acceptance rate of male applicants, which also increases insignificantly.

In the case of indirect bias through a proxy feature, we find that disclosures with explanations have a positive impact on fairness both with respect to perceptions and decision-making. Similar to the without explanations case (\aptLtoX[graphic=no,type=html]{\S\ref{sec:disc}}{\autoref{sec:disc}}), disclosing both model bias and the association between gender and university while including explanations significantly decreases perceived model fairness (\aptLtoX[graphic=no,type=html]{Figure~\ref{subfig:disc_expl_fp}}{\autoref{subfig:disc_expl_fp}}). For fairness rating, this effect is significant even without the correlation disclosure. 
Further, disclosing model bias alone, as well as disclosing model bias along with the correlation between protected and proxy feature with explanations leads to a significant increase in gender parity (\aptLtoX[graphic=no,type=html]{Figure~\ref{subfig:disc_expl_dmf}}{\autoref{subfig:disc_expl_dmf}}). This stems from a significantly higher acceptance rate for female applicants., while the acceptance rate for male applicants remains unchanged (\aptLtoX[graphic=no,type=html]{Table~\ref{tab:flip_rate}}{\autoref{tab:flip_rate}} in the Appendix).  
This also results in a higher FPR (significant) and a lower FNR (not significant), with an overall drop in accuracy (significant). 

In sum, in the case of direct bias, even though bias disclosure with explanations does not improve recognition of model unfairness significantly (possibly because fairness ratings are low even pre-disclosures), it does reduce agreement with the model's biased decisions, leading to a significantly higher acceptance rate for female applicants (however, decision-making fairness does not improve, possibly because of the normalizing effect of acceptance rate of male applicants, which also increases, albeit insignificantly). This is in stark contrast with the effect of explanations alone (\aptLtoX[graphic=no,type=html]{\S\ref{sec:expl}}{\autoref{sec:expl}}), which improved recognition of unfairness but led to more biased decisions overall. 

Further, in the case of indirect bias, disclosing the model bias and correlations between protected and proxy feature with explanations significantly increases both recognition of unfairness and gender parity in decision-making. We also observe an approximately $1\%$ drop in accuracy in the case of indirect bias after disclosing model bias and correlations with explanations (which might be acceptable in certain cases). Overall, we conclude that neither explanations nor bias or correlation disclosures alone are sufficient. We observe better decision-making fairness outcomes when participants are not only shown the model explanations but also made aware of the biases underlying them.

\subsection{Effect of Joint Intervention}\label{sec:kitchen sink}
We have seen that explanations alone decrease decision-making fairness, while disclosures with explanations can, in the case of indirect bias, have the opposite effect. Here, we consider the effect of both adding explanations and giving disclosures over including neither (i.e., comparing phase 1 without explanations to phase 2 with explanations). 

We find that, for decision-making metrics (accuracy, FPR, and FNR), the effect of joint intervention is never significant (\aptLtoX[graphic=no,type=html]{Table~\ref{tab:full}}{\autoref{tab:full}} in the Appendix). For fairness perception metrics, since the effect of explanations alone and disclosures with explanations pointed in the same direction, as expected, adding both explanations and disclosures also significantly improves recognition of unfairness. In sum, the joint intervention of including both explanations and disclosures (over including neither) helps participants in recognizing model biases but not in correcting them. This is a curious finding as we would expect that giving participants full information about bias in the model and also an explanation of individual prediction would lead to more fair decisions. We believe that this indicates that even though disclosures help undo some of the over-reliance on the model's biased decisions stemming from the inclusion of explanations, users still tend to be more accepting of model decisions with explanations than without.

\subsection{Effects of Additional Variables}
\label{subsec:additional_results}

Beyond the primary effects considered in our study, we also investigate the effect of participants' dispositional trust levels on decisions and perceptions (RQ5) and the effect of our interventions on learned trust (RQ6). 
We include a detailed discussion of these results in \aptLtoX[graphic=no,type=html]{Appendix~\ref{sec:result_overflow}}{\autoref{sec:result_overflow}}, along with a discussion of the differences between dispositional and learned trust and the relationship between a participant's gender and their decisions and perceptions.

\paragraph{Does dispositional trust affect decision-making and fairness perception measures?} 
As discussed in \aptLtoX[graphic=no,type=html]{\S\ref{sec:stats}}{\autoref{sec:stats}}, we include a measurement of a participant's dispositional trust in AI as a fixed effect in our linear models. The effects and their significance were generally not consistent across models.
Overall, we find that dispositional trust does not affect fairness perception in the case of direct bias, but it indeed leads to a significantly higher perceived fairness in the case of indirect bias, that is, participants with higher levels of dispositional trust were also less able to recognize indirect bias. 
Additionally, we find higher dispositional trust in AI was associated with making less fair decisions by relying more on the biased model. This is in line with previous findings that a person's dispositional trust significantly affects their reliance on a machine~\cite{merritt2008dispositional}. 
However, we find that this effect is alleviated after including disclosures, both in the case of direct and indirect model bias.  

\paragraph{Do explanations and disclosures affect self-reported learned trust?}
In addition to the fairness perception, decision-making fairness and quality measures discussed above, we additionally consider the effect of the explanation and disclosure interventions on learned trust when compared to dispositional trust in AI generally. 
We find that our interventions generally have no effect on learned trust ratings in models exhibiting direct bias, except for explanations leading to significantly lowered feelings that the AI system works well.
When model biases are indirect, full disclosures with explanations (or sometimes full disclosures without explanations) lead to a drop in learned trust. 
Lastly, explanations alone and full disclosures alone also lead to an increase in the predictability of the underlying model in the case of indirect bias but not in the case of direct bias. 

\section{Qualitative Results}

As described in \aptLtoX[graphic=no,type=html]{\S\ref{subsec:procedure}}{\autoref{subsec:procedure}}, for a selected set of applicants in each phase, participants were asked to write a free text
justification for why they agreed or disagreed with AI (or marked it as ``\textit{neutral}'') after marking their agreement and confidence. 
In addition to encouraging careful thinking, this also helps us gauge the kinds of reasoning participants employ in their decision-making. 

As the main goal of our study is to understand how humans interact with AI decisions when the AI is biased, we primarily focus our qualitative analysis on rationales concerning biases. 
To analyze how participants perceive and use (or discard) the biased feature (gender or university), we consider justifications that directly reference the protected (gender) or proxy feature (university) by using a set of keywords for both. We started with an initial keyword set (e.g., ``gender'', ``female'', ``university'') and, based on reading a subset of the justifications, expanded to include spelling variations (e.g., ``skool'' and ``collage'') and other topically relevant words (e.g., abbreviated names of schools).
We discuss our qualitative findings on justifications involving the protected feature in \aptLtoX[graphic=no,type=html]{\S\ref{subsec:justification_protected} and justifications involving the proxy feature in \S\ref{subsec:justification_proxy}. Lastly, we discuss additional observations indicating over-reliance based on a random sample of justifications in \S\ref{subsec:additional_qualitative}.}{\autoref{subsec:justification_protected} and justifications involving the proxy feature in \autoref{subsec:justification_proxy}. Lastly, we discuss additional observations indicating over-reliance based on a random sample of justifications in \autoref{subsec:additional_qualitative}.}

\subsection{Justification Involving Protected Feature}
\label{subsec:justification_protected}

Here, we analyze justifications that explicitly mention 
gender in the direct bias conditions. In these justifications, we assess how participants incorporate gender into their judgment of AI predictions or when making their own predictions.
Pre-disclosure, when explanations are not provided, participants rarely discuss gender as a salient part of their justification. 
However, when explanations are provided, participants often mention trying to ``ignore gender'' when making their decision. Notably, participants who mentioned gender in predictions about female applicants tended to make ``Complete'' or ``Neutral'' predictions. Thus, even though explanations significantly decreased gender parity (\aptLtoX[graphic=no,type=html]{\S\ref{sec:expl}}{\autoref{sec:expl}}) overall, they appear to help participants correct model biases in some cases. 

Post-bias disclosure, justifications mentioning gender still predominantly appear in the condition with explanations. Nevertheless, some participants, even without explanations, mention gender bias and flip ``Late'' predictions for female applicants.
For instance, one participant explained overriding such a prediction based on observing a male applicant with the same occupation predicted as ``Complete''. In the condition with explanations, many participants asserted that ``\textit{gender should not be a deciding factor}'' and ignored gender when making their prediction. 
Some participants, when supporting a "Late" prediction for a female applicant, clarify that their decision was based on other features (``\textit{The large amount of negatives aside from gender still point towards being late.}''). 
However, even in the case of direct bias with both explanations and bias disclosure, some participants still align with model biases. For example, one participant agreed with a ``Complete'' prediction of a male applicant  ``\textit{[b]ecause according to AI, male gender is more likely to complete loan...}''

\subsection{Justifications Involving Proxy Feature}
\label{subsec:justification_proxy}

Here, we analyze justifications that explicitly mention universities in the indirect bias conditions and assess how participants incorporate universities in their judgments. Pre-disclosures, some participants mentioned that attending college generally increases the likelihood of repayment, regardless of the specific school (for example, mentioning that an applicant is ``college-educated'' and predicting ``Complete''). However, other participants factor in the specific college when deciding to accept or reject an applicant. In conditions without explanations, we see evidence of participants relying on their own judgment of school quality. 
For instance, participants mention that they have ``\textit{never heard of Kenyon College},'' or that a co-ed school in the application ``...\textit{is not a particularly prestigious university}'' (even though in the underlying model, attending a co-ed school is counted as a positive). In contrast, with explanations provided, we instead see examples of participants aligning their evaluation of a school with the model’s biases. 
For instance, on applicants from women's colleges, participants claimed ``\textit{The applicant didn't go to a good college},'' or ``...\textit{College history was a major contributing factor to being late on loan}.'' On applications from co-ed schools, participants claimed that the applicant ``...\textit{attended a good university},'' or ``\textit{the university is listed as a good one}.'' This supports our quantitative finding that explanations alone lead participants to align with model biases (\aptLtoX[graphic=no,type=html]{\S\ref{sec:expl}}{\autoref{sec:expl}}). 

After bias disclosure alone (and especially without explanations), mentions of universities were quite sparse. Some participants mentioned that the university feature is given excessive weight (``\textit{I just find it hilarious that the borrower's state and university is such a huge factor}.'') but may not recognize this as indirect bias. One participant, although aware that Bryn Mawr is a women’s college, expressed uncertainty about identifying biased predictions without direct access to protected features: ``\textit{After learning more about possible discriminatory predictions on the AI's part... I'm specifically concerned about gender and race... but don't quite know how to discern that from these charts... This applicant profile gave me pause because I *think* Bryn Mawr College is an all-women's college.}''

After full bias and correlation disclosure, the ``university'' feature appears frequently in justifications. Here, participants continue to highlight the excessive weight assigned to the university in explanations, noting particularly unwarranted negative weight on women's colleges (e.g., ``\textit{While the system said late, I thought this was unfair because it placed a strong negative value on the college, which might be a women's college}.''). 
Although some participants use this university bias as a justification for flipping model predictions, many acknowledge the bias and either make a neutral prediction or concur with predictions of women being late in repayment. Additionally, we observed that some participants struggled to recall which universities were co-ed, 
which may have limited their ability to intervene and correct model biases.

\subsection{Justifications Indicating Over-reliance}\label{subsec:additional_qualitative}

In addition to the positive examples of explanations and disclosures helping participants notice and correct model biases, we also observe instances where decisions were solely based on the AI prediction or the corresponding explanations, regardless of disclosures. For example, even after bias disclosure in a direct bias condition, a participant agreed with a prediction of a female applicant being ``Late'' saying that ``\textit{They seem to have more negatives than positives.}'' Similarly, after bias and correlation disclosures in an indirect bias condition, a participant changed the prediction for an applicant from a women's college from ``Complete'' to ``Late,'' providing a similar justification. This indicates some participants persist in using explanations containing known biases, since for these applicants, ignoring the biased features (gender and university, respectively) would have resulted in the positives outweighing the negatives.

We also find instances of over-trust in AI even after being told that AI is biased such as ``\textit{I have no reason to disagree with the AI, if the AI is discriminating it probably has a good reason to,}'' or ``\textit{An AI is usually better than a professional let alone an amateur like me.}'' This indicates that even despite explanations and disclosures, there is room for improvement in educating and training humans to avoid unwarranted trust in AI systems and promote fair decision-making.

\section{Discussion and Limitations}

\aptLtoX[graphic=no,type=html]{
\begin{table*}
    \small
    \centering
    \begin{tabular}{clccccccccccc}
        \toprule
        &&\multicolumn{2}{c}{\makecell{Explanations\\Only}}&\multicolumn{3}{c}{\makecell{Disclosures Without \\Explanations}}&\multicolumn{3}{c}{\makecell{Disclosures With \\Explanations}}&\multicolumn{3}{c}{\makecell{Joint\\Intervention}}\\
        \cmidrule(lr){3-4} \cmidrule(lr){5-7} \cmidrule(lr){8-10}  \cmidrule(lr){11-13}
        &&Prot&Prox&\makecell{Prot \\BD}&\makecell{Prox \\BD}&\makecell{Prox \\BD+CD}&\makecell{Prot \\BD}&\makecell{Prox \\BD}&\makecell{Prox \\BD+CD} & \makecell{Prot \\BD}&\makecell{Prox \\BD}&\makecell{Prox \\BD+CD}\\
        \midrule
        \rowcolor{customrowcolor} 
        Fairness Perception&Fairness Rating&$\downarrow$&$\cdot$&$\cdot$&$\cdot$&$\downarrow$&$\cdot$&$\downarrow$&$\downarrow$&$\downarrow$&$\downarrow$&$\downarrow$\\
        \rowcolor{customrowcolor}
        Fairness Perception&\makecell[l]{Fairness Saliency}&$\uparrow$&$\cdot$&$\cdot$&$\cdot$&$\uparrow$&$\cdot$&$\cdot$&$\uparrow$&$\uparrow$&$\cdot$&$\uparrow$\\
        \addlinespace
        DM Fairness&Gender Parity&$\downarrow$&$\downarrow$&$\cdot$&$\cdot$&$\cdot$&$\cdot$&$\uparrow$&$\uparrow$&$\cdot$&$\cdot$&$\cdot$\\
        \addlinespace
        \rowcolor{customrowcolor}
        DM Quality&Accuracy&$\uparrow$&$\uparrow$&$\cdot$&$\cdot$&$\downarrow$&$\cdot$&$\downarrow$&$\downarrow$&$\cdot$&$\cdot$&$\cdot$\\
        \rowcolor{customrowcolor}
        DM Quality&FNR&$\uparrow$&$\cdot$&$\cdot$&$\cdot$&$\cdot$&$\downarrow$&$\cdot$&$\cdot$&$\cdot$&$\cdot$&$\cdot$\\
        \rowcolor{customrowcolor}
        DM Quality&FPR&$\downarrow$&$\uparrow$&$\cdot$&$\cdot$&$\uparrow$&$\uparrow$&$\uparrow$&$\uparrow$&$\cdot$&$\cdot$&$\cdot$\\
        \bottomrule
    \end{tabular}
    \caption{Summary of our main results. Arrows represent significant effects and point in the direction of the change. ``BD'' and ``CD'' represent bias and correlation disclosures, respectively.}
    \label{tab:result_summary}
\end{table*}
}
{\begin{table*}
    \small
    \centering
    \begin{tabular}{clccccccccccc}
        \toprule
        &&\multicolumn{2}{c}{\makecell{Explanations\\Only}}&\multicolumn{3}{c}{\makecell{Disclosures Without \\Explanations}}&\multicolumn{3}{c}{\makecell{Disclosures With \\Explanations}}&\multicolumn{3}{c}{\makecell{Joint\\Intervention}}\\
        \cmidrule(lr){3-4} \cmidrule(lr){5-7} \cmidrule(lr){8-10}  \cmidrule(lr){11-13}
        &&Prot&Prox&\makecell{Prot \\BD}&\makecell{Prox \\BD}&\makecell{Prox \\BD+CD}&\makecell{Prot \\BD}&\makecell{Prox \\BD}&\makecell{Prox \\BD+CD} & \makecell{Prot \\BD}&\makecell{Prox \\BD}&\makecell{Prox \\BD+CD}\\
        \midrule
        \rowcolor{customrowcolor} 
        &Fairness Rating&$\downarrow$&$\cdot$&$\cdot$&$\cdot$&$\downarrow$&$\cdot$&$\downarrow$&$\downarrow$&$\downarrow$&$\downarrow$&$\downarrow$\\
        \rowcolor{customrowcolor}
        \multirow{-2}{*}{\parbox{1.3cm}{Fairness Perception}}&\makecell[l]{Fairness Saliency}&$\uparrow$&$\cdot$&$\cdot$&$\cdot$&$\uparrow$&$\cdot$&$\cdot$&$\uparrow$&$\uparrow$&$\cdot$&$\uparrow$\\
        \addlinespace
        DM Fairness&Gender Parity&$\downarrow$&$\downarrow$&$\cdot$&$\cdot$&$\cdot$&$\cdot$&$\uparrow$&$\uparrow$&$\cdot$&$\cdot$&$\cdot$\\
        \addlinespace
        \rowcolor{customrowcolor}
        &Accuracy&$\uparrow$&$\uparrow$&$\cdot$&$\cdot$&$\downarrow$&$\cdot$&$\downarrow$&$\downarrow$&$\cdot$&$\cdot$&$\cdot$\\
        \rowcolor{customrowcolor}
        &FNR&$\uparrow$&$\cdot$&$\cdot$&$\cdot$&$\cdot$&$\downarrow$&$\cdot$&$\cdot$&$\cdot$&$\cdot$&$\cdot$\\
        \rowcolor{customrowcolor}
        \multirow{-3}{*}{DM Quality}&FPR&$\downarrow$&$\uparrow$&$\cdot$&$\cdot$&$\uparrow$&$\uparrow$&$\uparrow$&$\uparrow$&$\cdot$&$\cdot$&$\cdot$\\
        \bottomrule
    \end{tabular}
    \caption{Summary of our main results. Arrows represent significant effects and point in the direction of the change. ``BD'' and ``CD'' represent bias and correlation disclosures, respectively.}
    \label{tab:result_summary}
\end{table*}}

In this work, we studied the effect of explanations and disclosures on fairness perceptions and decision-making when humans are provided predictions from models exhibiting direct or indirect bias. Our findings are summarized in \aptLtoX[graphic=no,type=html]{Table~\ref{tab:result_summary}}{\autoref{tab:result_summary}}. 
Regardless of intervention, we consistently observed that human-AI teams made fairer decisions than the AI alone. 
We found that explanations alone significantly improved participants' ability to notice unfairness in the case of direct bias only. However, explanations led participants to be more influenced by model biases, whether they noticed these biases or not. Disclosures were an effective tool for helping users recognize unfairness in the case of indirect bias, especially with the help of explanations. And we saw that this increased recognition of bias was paired with fairer human-AI decisions, showing that disclosures helped participants understand when and how to intervene on model decisions to produce fairer outcomes. 

However, we found that the joint intervention of including both explanations and disclosures (over including neither) was only effective in helping participants recognize model bias, not correct it. If the main objective is to help users notice direct model biases, we recommend including explanations, and if it is to help users notice indirect model bias, we recommend including explanations and disclosing both model bias and the correlations between protected and proxy features. 
But if the main objective is to help the human-AI team produce fairer outcomes, we did not find including explanations with disclosures to be an effective intervention. However, if explanations are to be used, then disclosures may help contextualize explanations and the potential biases, especially when these biases are indirect.
While in a perfect world, such known biases could be addressed in the model itself instead of relying on human intervention, this may not always be possible. In many cases, we may have limited access to the underlying model (e.g., only having API access) or may not be able to non-superficially ``debias'' it~\cite{gonen-goldberg-2019-lipstick}. Disclosures may help uncover these biases to humans, possibly leading to fairer human-AI decisions. 

A key limitation of work is that since we show our participants partially-synthetic loan data, we cannot directly rely on the existing ground truth. Instead, we calculate the expectation of ground-truth based metrics (accuracy, FNR, and FPR) which means that there are applicants for which neither choice is very likely to be ``correct'' (i.e., both $P(Y_i=1)$ and $P(Y_i=0)$ are close to $0.5$). We handle this in part by adjusting for the baseline AI-only scores; however, using a fully non-synthetic dataset and original ground truth values may lead to cleaner results. 
This lack of a true ground-truth, in part, led us to use demographic parity which has been argued to be insufficient as a notion of fairness~\cite{dwork2012fairness}. 

There is potential concern about the use of a loan prediction task since the participants are not financial experts. As we discuss in \aptLtoX[graphic=no,type=html]{\S\ref{sec:data}}{\autoref{sec:data}}, participants are shown a subset of the original Prosper features that we believe are relatively intuitive without more than a commonsense understanding of lending (e.g., size of the loan being requested and employment status). We also hope that a task mimicking the loan approval process is high-stakes enough to encourage more care from the crowd-workers in their decision-making. However, more work is needed to study how our findings generalize to settings with varied task stakes or domain expertise.

Another limitation is that our study design forces participants to make decisions one at a time, without seeing the entire pool of applicants. It is our hope that the percent/percentile information given for each feature gave participants a better sense of how each applicant's profile compared to the general pool, even without seeing many profiles. However, we recognize that it may be difficult for participants to conceptualize what a ``strong'' or ``weak'' candidate looks like under this design. This may make it more difficult for participants who, for example, wish to increase the acceptance rate of women in phase 2 to decide which female applicants are ``most deserving'' of having their prediction flipped to ``Complete''.

Despite these limitations, our work provides insights into the effect of explanations on fairness in human-AI decision-making, especially when the biases are indirect (through a proxy features). We conclude that neither explanations nor disclosures alone improve the fairness of decisions made by a human-AI team. Our findings serve to caution the wider community from treating explanations as a foolproof solution to human-AI collaborative decision-making: explanations may not always make model biases clear and may make people more prone to align with model biases, leading to less fair decisions.  
When people are repeatedly exposed to explanations that justify or rationalize biased predictions, they may begin to accept these biases as valid or even desirable, rather than critically questioning and challenging them.
We highlight that explanations and disclosures in conjunction may be helpful to some extent. However, more work is needed to further examine how best to aid humans not only in identifying indirect model biases, but also in systematically correcting these biases.

%

\begin{acks}
We sincerely thank the CHI TRAIT and IUI reviewers, Md Naimul Hoque, and the members of the UMD CLIP and HCIL labs for their valuable feedback. Additionally, we would like to thank Solon Barocas, Jennifer Wortman Vaughan, Forough Poursabzi-Sangdeh, and Mahsan Nourani for their work on the version of the Prosper Loan dataset used in this paper.
\end{acks}

\bibliographystyle{ACM-Reference-Format}
\bibliography{main}

\appendix

\section{Deriving Metrics Using Probabilistic Ground Truth}\label{sec:deriving_ground_truth}

Here, we consider in more detail how to calculate the probability of ground-truth completion and the expected value of the decision quality metrics. 

\subsection{Probability of Loan Completion}\label{sec:ground_truth}

In our study, each applicant $i$ is only evaluated by a single participant $j$ in a given condition, and each participant $j$ evaluates $20$ applicants across the two phases (\aptLtoX[graphic=no,type=html]{\S\ref{subsec:procedure}}{\autoref{subsec:procedure}}). We represent the set of observed decisions as $S=\{(i,j) \mid \text{participant $j$ sees applicant $i$}\}$. 
For the $i^\text{th}$ applicant, we want to know the probability that the true outcome should be complete, that is, $P(Y_i=1)$. Let $\boldsymbol{x}_i$ be the set of original features of the $i^{th}$ applicant's and $x_i^\ast$ be the assigned (synthetic) gender or university. Note, we drop the subscript $i$ when referring to a general applicant. We can write the probability of the true outcome for applicant with features $(\boldsymbol{x}, x^\ast)$ as
\begin{align*}
P(Y=1\mid \boldsymbol{x},x^\ast) = \frac{P(x^\ast\mid Y=1,\boldsymbol{x})P(Y=1\mid \boldsymbol{x})}{P(x^\ast\mid \boldsymbol{x})}
\end{align*}
Since we assign the values of the protected or proxy feature $x^\ast$ based solely on the ground-truth outcome, we can assume that $x^\ast$ and $\boldsymbol{x}$ are independent given $y$. Thus, $P(x^\ast\mid Y=1,\boldsymbol{x})=P(x^\ast\mid Y=1)$.
\begin{align*}
    P(Y=1\mid \boldsymbol{x},x^\ast) &= \frac{P(x^\ast\mid Y=1)P(Y=1\mid \boldsymbol{x})}{P(x^\ast\mid \boldsymbol{x})}\\
    &= \frac{P(Y=1\mid x^\ast)P(x^\ast)}{P(Y=1)}\frac{P(Y=1\mid \boldsymbol{x})}{P(x^\ast\mid \boldsymbol{x})}\\
\end{align*}
Then, removing any terms not containing $Y$, which can be normalized away, we are left with
\begin{align*}
    P(Y=1\mid \boldsymbol{x},x^\ast) = \frac{P(Y=1\mid x^\ast)P(Y=1\mid \boldsymbol{x})}{P(Y=1)}\\
\end{align*}
In the protected case, we know the probability of acceptance given the synthetic feature (that is, $P(Y=1\mid x^{\ast})$) based on our selected $60/40$ male/female acceptance ratio. In the proxy case, we estimate the probability of acceptance given the university using the joint distribution of gender and university and the probability of acceptance given gender. 
For each applicant, we estimate the probability of the applicant completing their loan $P(Y=1\mid \boldsymbol{x})$ based on only the non-synthetic features ($\boldsymbol{x}$) using a linear regression model that does not have access to the synthetic information. Finally, we can calculate $P(Y=1)$ based on the rate of ground-truth acceptances in the original data.

\subsection{Expected Accuracy, FNR, and FPR}\label{sec:metric_der}
Using our calculated probability of the ground-truth acceptance of a given applicant (with original features $\boldsymbol{x}$ and synthetic feature $x^\ast$), we can calculate an expected accuracy, FNR, and FPR for a given human-AI team.  
For example, for expected FNR, we consider the expected number of false negatives over the expected number of ground truth positives.

\begin{align*}
    \text{Expected FNR} &= \frac{\ex{\hbox{\# False Negatives}}}{\ex{\hbox{\# Positives}}}
\end{align*}
Let $\hat{y}_{i, j}$ be the human-AI decision for the $i^{th}$ applicant by the $j^{th}$ participant, such that $\hat{y}_{i, j}$ is $1$ if the human-AI decision is ``Complete'' and 0 if it is ``Late''. 
We can write the expected number of false negatives $\ex{\hbox{\# False Negatives}}$ as $\sum_{(i,j)\in S} (1-\hat{y}_{i, j})P(Y_{i}=1 | \boldsymbol{x}_i, x^\ast_i)$, 
that is, the probability of the ground truth label for the $i^{th}$ applicant being 1 but the human-AI decision for the same applicant being 0. Similarly, we can write the expected value of positives $\ex{\hbox{\# Positives}}$ as $\sum_i P(Y_i=1| \boldsymbol{x}_i, x^\ast_i)$. Thus, we can write
\begin{align*}
    \text{Expected FNR} = \frac{\sum_{(i,j)\in S}P(Y_i=1| \boldsymbol{x}_i, x^\ast_i) \times (1-\hat{y}_{i,j})}{\sum_iP(Y_i=1| \boldsymbol{x}_i, x^\ast_i)}.
\end{align*}
Similarly, we can calculate the expected FPR and Accuracy.
\begin{align*}
    \text{Expected FPR} &= \frac{\ex{\hbox{\# False Positives}}}{\ex{\hbox{\# Negatives}}}\\
    &= \frac{\sum_{(i,j)\in S}P(Y_i=0| \boldsymbol{x}_i, x^\ast_i) \times \hat{y}_{i,j}}{\sum_i P(Y_i=0| \boldsymbol{x}_i, x^\ast_i)}\\
    \text{Expected Accuracy} &= \frac{\ex{\hbox{\# True Positives + \# True Negatives}}}{\ex{\hbox{\# Samples}}}\\
    &= \frac{1}{\mid S\mid }\sum_{(i,j)\in S}\ex{TP_{i,j}+TN_{i,j}}\\
    &=  \frac{1}{\mid S\mid }\sum_{(i,j)\in S} \big(P(Y_i=1| \boldsymbol{x}_i, x^\ast_i) \times \hat{y}_{i,j}\big)\\ &\quad+ \big(P(Y_i=0| \boldsymbol{x}_i, x^\ast_i) \times (1-\hat{y}_{i,j})\big) \\
\end{align*}

\section{Extended Results}\label{sec:result_overflow}
In this section, we report additional results regarding dispositional trust, learned trust, and participant gender. We additionally include a full table detailing the primary effects considered in the study (\aptLtoX[graphic=no,type=html]{Table~\ref{tab:full}}{\autoref{tab:full}}) as well as the effects of our interventions on reliance split by applicant gender (\aptLtoX[graphic=no,type=html]{Table~\ref{tab:flip_rate}}{\autoref{tab:flip_rate}}).

\subsection{Does dispositional trust affect decision-making and fairness perception measures?}\label{sec:bl_trust}
As discussed in section \ref{sec:stats}, we include a measurement of a participant's dispositional trust in AI as a fixed effect in our linear models. The effects and their significance was generally not consistent across models (See \aptLtoX[graphic=no,type=html]{Table~\ref{tab:bl_trust}}{\autoref{tab:bl_trust}}).

\aptLtoX[graphic=no,type=html]{\begin{table*}
    \centering
    \small
    \begin{tabular}{lllrrlrc}
    \toprule
    Metric&Feature&Model&\multicolumn{1}{c}{$F$}&\multicolumn{1}{c}{$p$}&&Coef&Std Error\\
    \midrule
    \rowcolor{customrowcolor}\cellcolor{white}Fairness Rating&protected&Expl&$F(1,115)=4.07$&$0.0459$&$ $&$1.31$&$0.592$\\
    \rowcolor{customrowcolor}\cellcolor{white}Fairness Rating&protected&Disclosure&$F(1,49)=5.86$&$0.0192$&$ $&$1.70$&$0.700$\\
    \rowcolor{customrowcolor}\cellcolor{white}Fairness Rating&protected&Disclosure with Explanation&$F(1,66)=1.6$&$0.2104$&$ $&$0.89$&$0.702$\\
    Fairness Rating&proxy&Expl&$F(1,227)=32.7$&$0.0000$&$*$&$1.54$&$0.280$\\
    Fairness Rating&proxy&Disclosure&$F(1,91)=25.8$&$0.0000$&$*$&$1.69$&$0.332$\\
    Fairness Rating&proxy&Disclosure with Explanation&$F(1,136)=10.11$&$0.0018$&$*$&$1.13$&$0.355$\\
    \midrule
    \rowcolor{customrowcolor}\cellcolor{white}Fairness Saliency&protected&Expl&$F(1,115)=0.87$&$0.3540$&$ $&$-0.31$&$0.284$\\
    \rowcolor{customrowcolor}\cellcolor{white}Fairness Saliency&protected&Disclosure&$F(1,49)=0.18$&$0.6737$&$ $&$-0.15$&$0.354$\\
    \rowcolor{customrowcolor}\cellcolor{white}Fairness Saliency&protected&Disclosure with Explanation&$F(1,66)=0.85$&$0.3598$&$ $&$-0.30$&$0.331$\\
    Fairness Saliency&proxy&Expl&$F(1,227)=2.52$&$0.1135$&$ $&$-0.12$&$0.087$\\
    Fairness Saliency&proxy&Disclosure&$F(1,91)=9.11$&$0.0033$&$*$&$-0.45$&$0.150$\\
    Fairness Saliency&proxy&Disclosure with Explanation&$F(1,136)=0.01$&$0.9361$&$ $&$-0.01$&$0.125$\\
    \midrule
    \rowcolor{customrowcolor}\cellcolor{white}Parity&protected&Expl&$F(1,115)=6.25$&$0.0138$&$*$&$-0.35$&$0.149$\\
    \rowcolor{customrowcolor}\cellcolor{white}Parity&protected&Disclosure&$F(1,99)=0.53$&$0.4681$&$ $&$-0.15$&$0.208$\\
    \rowcolor{customrowcolor}\cellcolor{white}Parity&protected&Disclosure with Explanation&$F(1,133)=3.27$&$0.0730$&$ $&$-0.28$&$0.153$\\
    Parity&proxy&Expl&$F(1,227)=5.73$&$0.0175$&$*$&$-0.22$&$0.112$\\
    Parity&proxy&Disclosure&$F(1,182)=0.31$&$0.5778$&$ $&$0.07$&$0.128$\\
    Parity&proxy&Disclosure with Explanation&$F(1,135)=3.55$&$0.0618$&$ $&$-0.19$&$0.100$\\
    \midrule
    \rowcolor{customrowcolor}\cellcolor{white}Accuracy&protected&Expl&$F(1,115)=2.09$&$0.1511$&$ $&$0.04$&$0.031$\\
    \rowcolor{customrowcolor}\cellcolor{white}Accuracy&protected&Disclosure&$F(1,49)=4.71$&$0.0349$&$ $&$0.10$&$0.044$\\
    \rowcolor{customrowcolor}\cellcolor{white}Accuracy&protected&Disclosure with Explanation&$F(1,66)=0.23$&$0.6366$&$ $&$0.02$&$0.033$\\
    Accuracy&proxy&Expl&$F(1,227)=2.97$&$0.0863$&$ $&$0.02$&$0.018$\\
    Accuracy&proxy&Disclosure&$F(1,91)=1.73$&$0.1917$&$ $&$0.03$&$0.026$\\
    Accuracy&proxy&Disclosure with Explanation&$F(1,136)=2.22$&$0.1382$&$ $&$0.03$&$0.019$\\
    \midrule
    \rowcolor{customrowcolor}\cellcolor{white}FNR&protected&Expl&$F(1,115)=0.06$&$0.8126$&$ $&$0.01$&$0.096$\\
    \rowcolor{customrowcolor}\cellcolor{white}FNR&protected&Disclosure&$F(1,49)=0.01$&$0.9113$&$ $&$-0.01$&$0.125$\\
    \rowcolor{customrowcolor}\cellcolor{white}FNR&protected&Disclosure with Explanation&$F(1,66)=0.53$&$0.4674$&$ $&$0.08$&$0.104$\\
    FNR&proxy&Expl&$F(1,227)=8.8$&$0.0033$&$*$&$0.14$&$0.054$\\
    FNR&proxy&Disclosure&$F(1,91)=0.6$&$0.4400$&$ $&$0.06$&$0.072$\\
    FNR&proxy&Disclosure with Explanation&$F(1,136)=9.69$&$0.0023$&$*$&$0.20$&$0.063$\\
    \midrule
    \rowcolor{customrowcolor}\cellcolor{white}FPR&protected&Expl&$F(1,115)=0.7$&$0.4051$&$ $&$-0.08$&$0.107$\\
    \rowcolor{customrowcolor}\cellcolor{white}FPR&protected&Disclosure&$F(1,49)=0.73$&$0.3983$&$ $&$-0.12$&$0.144$\\
    \rowcolor{customrowcolor}\cellcolor{white}FPR&protected&Disclosure with Explanation&$F(1,66)=0.59$&$0.4458$&$ $&$-0.10$&$0.125$\\
    FPR&proxy&Expl&$F(1,227)=9.44$&$0.0024$&$*$&$-0.16$&$0.059$\\
    FPR&proxy&Disclosure&$F(1,91)=2.07$&$0.1533$&$ $&$-0.12$&$0.085$\\
    FPR&proxy&Disclosure with Explanation&$F(1,136)=8.79$&$0.0036$&$*$&$-0.22$&$0.073$\\
    \bottomrule
    \end{tabular}
    \caption{Effects of dispositional trust in AI on different outcome metrics (Perception, Parity, Accuracy, FPR, and FNR).}\label{tab:bl_trust}
\end{table*}}
{\begin{table*}
    \centering
    \small
    \begin{tabular}{lllrrlrc}
    \toprule
    Metric&Feature&Model&\multicolumn{1}{c}{$F$}&\multicolumn{1}{c}{$p$}&&Coef&Std Error\\
    \midrule
    \rowcolor{customrowcolor}\cellcolor{white}&&Expl&$F(1,115)=4.07$&$0.0459$&$ $&$1.31$&$0.592$\\
    \rowcolor{customrowcolor}\cellcolor{white}&&Disclosure&$F(1,49)=5.86$&$0.0192$&$ $&$1.70$&$0.700$\\
    \rowcolor{customrowcolor}\cellcolor{white}&\multirow{-3}{*}{protected}&Disclosure with Explanation&$F(1,66)=1.6$&$0.2104$&$ $&$0.89$&$0.702$\\
    &&Expl&$F(1,227)=32.7$&$0.0000$&$*$&$1.54$&$0.280$\\
    &&Disclosure&$F(1,91)=25.8$&$0.0000$&$*$&$1.69$&$0.332$\\
    \multirow{-6}{*}{\parbox{1.5cm}{Fairness\\Rating}}&\multirow{-3}{*}{proxy}&Disclosure with Explanation&$F(1,136)=10.11$&$0.0018$&$*$&$1.13$&$0.355$\\
    \midrule
    \rowcolor{customrowcolor}\cellcolor{white}&&Expl&$F(1,115)=0.87$&$0.3540$&$ $&$-0.31$&$0.284$\\
    \rowcolor{customrowcolor}\cellcolor{white}&&Disclosure&$F(1,49)=0.18$&$0.6737$&$ $&$-0.15$&$0.354$\\
    \rowcolor{customrowcolor}\cellcolor{white}&\multirow{-3}{*}{protected}&Disclosure with Explanation&$F(1,66)=0.85$&$0.3598$&$ $&$-0.30$&$0.331$\\
    &&Expl&$F(1,227)=2.52$&$0.1135$&$ $&$-0.12$&$0.087$\\
    &&Disclosure&$F(1,91)=9.11$&$0.0033$&$*$&$-0.45$&$0.150$\\
    \multirow{-6}{*}{\parbox{1.5cm}{Fairness\\Saliency}}&\multirow{-3}{*}{proxy}&Disclosure with Explanation&$F(1,136)=0.01$&$0.9361$&$ $&$-0.01$&$0.125$\\
    \midrule
    \rowcolor{customrowcolor}\cellcolor{white}&&Expl&$F(1,115)=6.25$&$0.0138$&$*$&$-0.35$&$0.149$\\
    \rowcolor{customrowcolor}\cellcolor{white}&&Disclosure&$F(1,99)=0.53$&$0.4681$&$ $&$-0.15$&$0.208$\\
    \rowcolor{customrowcolor}\cellcolor{white}&\multirow{-3}{*}{protected}&Disclosure with Explanation&$F(1,133)=3.27$&$0.0730$&$ $&$-0.28$&$0.153$\\
    &&Expl&$F(1,227)=5.73$&$0.0175$&$*$&$-0.22$&$0.112$\\
    &&Disclosure&$F(1,182)=0.31$&$0.5778$&$ $&$0.07$&$0.128$\\
    \multirow{-6}{*}{\parbox{1.5cm}{Parity}}&\multirow{-3}{*}{proxy}&Disclosure with Explanation&$F(1,135)=3.55$&$0.0618$&$ $&$-0.19$&$0.100$\\
    \midrule
    \rowcolor{customrowcolor}\cellcolor{white}&&Expl&$F(1,115)=2.09$&$0.1511$&$ $&$0.04$&$0.031$\\
    \rowcolor{customrowcolor}\cellcolor{white}&&Disclosure&$F(1,49)=4.71$&$0.0349$&$ $&$0.10$&$0.044$\\
    \rowcolor{customrowcolor}\cellcolor{white}&\multirow{-3}{*}{protected}&Disclosure with Explanation&$F(1,66)=0.23$&$0.6366$&$ $&$0.02$&$0.033$\\
    &&Expl&$F(1,227)=2.97$&$0.0863$&$ $&$0.02$&$0.018$\\
    &&Disclosure&$F(1,91)=1.73$&$0.1917$&$ $&$0.03$&$0.026$\\
    \multirow{-6}{*}{\parbox{1.5cm}{Accuracy}}&\multirow{-3}{*}{proxy}&Disclosure with Explanation&$F(1,136)=2.22$&$0.1382$&$ $&$0.03$&$0.019$\\
    \midrule
    \rowcolor{customrowcolor}\cellcolor{white}&&Expl&$F(1,115)=0.06$&$0.8126$&$ $&$0.01$&$0.096$\\
    \rowcolor{customrowcolor}\cellcolor{white}&&Disclosure&$F(1,49)=0.01$&$0.9113$&$ $&$-0.01$&$0.125$\\
    \rowcolor{customrowcolor}\cellcolor{white}&\multirow{-3}{*}{protected}&Disclosure with Explanation&$F(1,66)=0.53$&$0.4674$&$ $&$0.08$&$0.104$\\
    &&Expl&$F(1,227)=8.8$&$0.0033$&$*$&$0.14$&$0.054$\\
    &&Disclosure&$F(1,91)=0.6$&$0.4400$&$ $&$0.06$&$0.072$\\
    \multirow{-6}{*}{\parbox{1.5cm}{FNR}}&\multirow{-3}{*}{proxy}&Disclosure with Explanation&$F(1,136)=9.69$&$0.0023$&$*$&$0.20$&$0.063$\\
    \midrule
    \rowcolor{customrowcolor}\cellcolor{white}&&Expl&$F(1,115)=0.7$&$0.4051$&$ $&$-0.08$&$0.107$\\
    \rowcolor{customrowcolor}\cellcolor{white}&&Disclosure&$F(1,49)=0.73$&$0.3983$&$ $&$-0.12$&$0.144$\\
    \rowcolor{customrowcolor}\cellcolor{white}&\multirow{-3}{*}{protected}&Disclosure with Explanation&$F(1,66)=0.59$&$0.4458$&$ $&$-0.10$&$0.125$\\
    &&Expl&$F(1,227)=9.44$&$0.0024$&$*$&$-0.16$&$0.059$\\
    &&Disclosure&$F(1,91)=2.07$&$0.1533$&$ $&$-0.12$&$0.085$\\
    \multirow{-6}{*}{\parbox{1.5cm}{FPR}}&\multirow{-3}{*}{proxy}&Disclosure with Explanation&$F(1,136)=8.79$&$0.0036$&$*$&$-0.22$&$0.073$\\
    \bottomrule
    \end{tabular}
    \caption{Effects of dispositional trust in AI on different outcome metrics (Perception, Parity, Accuracy, FPR, and FNR).}\label{tab:bl_trust}
\end{table*}}

 We consistently find that dispositional trust has no significant impact on fairness ratings in the protected conditions, while it significantly increases fairness ratings in the proxy conditions. In other words, when biases are direct, people are equally able to notice model biases even when they tend to trust AI in general; however, when biases are indirect, people with higher dispositional trust in AI are less likely to believe that the model is unfair. For participants' fairness saliency, we see that there is never a significant effect in the case of direct bias, but higher dispositional trust significantly decreased the rate of disagreement only when considering disclosures without explanations. 

 We also find that, under models that consider the effect of explanations and disclosures with explanations, increases dispositional trust in AI significantly increased FPR and decreased FNR in the proxy conditions only. This is likely due to participants with higher trust in AI being more influenced by subtle indirect biases leading to lower acceptance rates for female applicants. This is also supported by the models measuring the effect of explanations on gender parity. Here, we see that an increased dispositional trust in AI significantly decreased parity under both types of bias.

 Overall, people with a greater dispositional trust in AI tended to make more unfair decisions (when working with a biased model) and were less likely to notice indirect bias.

\subsection{Do explanations and disclosures affect learned trust?}
\label{sec:effect_on_trust}

In this section, we discuss the effect of our interventions---explanations, disclosures without explanations, and disclosures with explanations (\autoref{fig:expl_disclosure})---on learned trust over dispositional trust in AI generally. We perform statistical tests similar to the ones described in \aptLtoX[graphic=no,type=html]{\S\ref{sec:stats}}{\autoref{sec:stats}}. We consider the different trust measures as the dependent variable and the treatment as the fixed effect term. We also control for the dispositional trust level as a fixed effect.  

As seen in \aptLtoX[graphic=no,type=html]{Table~\ref{tab:full}}{\autoref{tab:full}}, we find that our treatments generally have no effect on trust ratings in models exhibiting direct bias, except for explanations alone leading to significantly lowered feelings that the AI system works well. In the case of indirect bias, we often see that full disclosure with explanations (and sometimes also full disclosure without explanations or bias disclosure with explanations) has a significant effect on learned trust. These effects demonstrate lowered feelings that the AI system works well as well as decreased feelings that the AI system can perform as well as an untrained human, decreased confidence in the system, decreased feelings of safety when relying on the system, and increased wariness of the AI system. 

We also find that explanations alone significantly increase participant's perception of model predictability in the proxy conditions but not the protected conditions. Without explanations, full bias and correlation disclosure also significantly increased predictability. With explanations, however, disclosures do not increase predictability, likely due to high predictability ratings even with explanations alone. This is to say that our models are already seen as relatively predictable when biases are direct, but when biases are indirect, explanations or disclosure of model bias and the model's usage of the university feature help make the model more predictable.

\aptLtoX[graphic=no,type=html]{\begin{table*}
\begin{tabular}{cccccccccccccccccc}
\toprule
&&&&\multicolumn{2}{c}{better}&\multicolumn{2}{c}{confidence}&\multicolumn{2}{c}{predictable}&\multicolumn{2}{c}{safe}&\multicolumn{2}{c}{wary}&\multicolumn{2}{c}{works well}\\
\cline{5-6}\cline{7-8}\cline{9-10}\cline{11-12}\cline{13-14}\cline{15-16}
Feature&Phase&Expl?&Disclosure&sig&\text{coef}&sig&\text{coef}&sig&\text{coef}&sig&\text{coef}&sig&\text{coef}&sig&\text{coef}\\
\midrule
\rowcolor{customrowcolor}Protected&$1$&-&-&$*$&-0.61&$*$&-0.43&$*$&-0.51&$*$&-0.29& &0.04&$*$&-0.43\\
\rowcolor{customrowcolor}Protected&$1$&\checkmark&-&$*$&-0.93&$*$&-0.66&$*$&-0.37& &-0.26& &0.10&$*$&-0.81\\
\rowcolor{customrowcolor}Protected&$2$&-&Bias&$*$&-0.33&$*$&-0.23&$*$&-0.25&$*$&-0.20& &0.02&$*$&-0.28\\
\rowcolor{customrowcolor}Protected&$2$&\checkmark&Bias&$*$&-0.51&$*$&-0.38&$*$&-0.19&$*$&-0.22& &0.12&$*$&-0.48\\
Proxy&$1$&-&-&$*$&-0.55&$*$&-0.34&$*$&-0.68&$*$&-0.22& &-0.15&$*$&-0.32\\
Proxy&$1$&\checkmark&-&$*$&-0.54& &-0.19& &0.01& &0.03& &-0.14&$*$&-0.29\\
Proxy&$2$&-&Bias&$*$&-0.39&$*$&-0.21&$*$&-0.29& &-0.05& &0.02&$*$&-0.25\\
Proxy&$2$&-&Full&$*$&-0.31&$*$&-0.28&$*$&-0.23& &-0.13& &0.01&$*$&-0.37\\
Proxy&$2$&\checkmark&Bias&$*$&-0.30&$*$&-0.29& &0.02& &-0.09& &0.06&$*$&-0.30\\
Proxy&$2$&\checkmark&Full&$*$&-0.36&$*$&-0.20& &-0.04& &-0.08& &0.00&$*$&-0.26\\
\bottomrule
\end{tabular}
\caption{Comparison of dispositional trust vs learned trust in varied conditions and phases.}
\label{tab:trust_vs_bl}
\end{table*}}{\begin{table*}
\centering
\small
\begin{tabular}{cccccS[table-format=1.2]cS[table-format=1.2]cS[table-format=1.2]cS[table-format=1.2]cS[table-format=1.2]cS[table-format=1.2]}%
\toprule
&&&&\multicolumn{2}{c}{better}&\multicolumn{2}{c}{confidence}&\multicolumn{2}{c}{predictable}&\multicolumn{2}{c}{safe}&\multicolumn{2}{c}{wary}&\multicolumn{2}{c}{works well}\\
\cmidrule(lr){5-6}\cmidrule(lr){7-8}\cmidrule(lr){9-10}\cmidrule(lr){11-12}\cmidrule(lr){13-14}\cmidrule(lr){15-16}
Feature&Phase&Expl?&Disclosure&sig&\text{coef}&sig&\text{coef}&sig&\text{coef}&sig&\text{coef}&sig&\text{coef}&sig&\text{coef}\\
\midrule
\rowcolor{customrowcolor}&$1$&-&-&$*$&-0.61&$*$&-0.43&$*$&-0.51&$*$&-0.29& &0.04&$*$&-0.43\\
\rowcolor{customrowcolor}&$1$&\checkmark&-&$*$&-0.93&$*$&-0.66&$*$&-0.37& &-0.26& &0.10&$*$&-0.81\\
\rowcolor{customrowcolor}&$2$&-&Bias&$*$&-0.33&$*$&-0.23&$*$&-0.25&$*$&-0.20& &0.02&$*$&-0.28\\
\rowcolor{customrowcolor}\multirow{-4}{*}{\rotatebox[origin=c]{90}{Protected}}&$2$&\checkmark&Bias&$*$&-0.51&$*$&-0.38&$*$&-0.19&$*$&-0.22& &0.12&$*$&-0.48\\
&$1$&-&-&$*$&-0.55&$*$&-0.34&$*$&-0.68&$*$&-0.22& &-0.15&$*$&-0.32\\
&$1$&\checkmark&-&$*$&-0.54& &-0.19& &0.01& &0.03& &-0.14&$*$&-0.29\\
&$2$&-&Bias&$*$&-0.39&$*$&-0.21&$*$&-0.29& &-0.05& &0.02&$*$&-0.25\\
&$2$&-&Full&$*$&-0.31&$*$&-0.28&$*$&-0.23& &-0.13& &0.01&$*$&-0.37\\
&$2$&\checkmark&Bias&$*$&-0.30&$*$&-0.29& &0.02& &-0.09& &0.06&$*$&-0.30\\
\multirow{-6}{*}{\rotatebox[origin=c]{90}{Proxy}}&$2$&\checkmark&Full&$*$&-0.36&$*$&-0.20& &-0.04& &-0.08& &0.00&$*$&-0.26\\
\bottomrule
\end{tabular}
\caption{Comparison of dispositional trust vs learned trust in varied conditions and phases.}
\label{tab:trust_vs_bl}
\end{table*}}

\subsection{Does dispositional trust differ from learned trust?}\label{sec:baseline_vs_ourtrust}
In the previous section, we discussed how interventions affected learned trust when controlling for baseline dispositional trust. Here, we study whether there is a significant difference between dispositional trust and learned trust in the biased models across the questions described in \aptLtoX[graphic=no,type=html]{\S\ref{sec:surveys}}{\autoref{sec:surveys}}. These results are shown in \aptLtoX[graphic=no,type=html]{Table~\ref{tab:trust_vs_bl}}{\autoref{tab:trust_vs_bl}}. 

We find that participants usually thought our model worked worse than AI does generally, that it inspired less confidence, and was less predictable. 
Participants also regarded our AI system as less safe than AI in general, but this is primarily true only in the case of direct bias. Surprisingly, participants did not consider our AI systems to be less safe than general in phase 1 when they were given explanations (which would have directly indicated that the system used gender as a feature to determine loan outcomes). 
Participants' wariness of the biased models was not significantly different from their baseline wariness in AI.

\subsection{Does participant gender correlate with decision-making and fairness perception measures?}\label{sec:participant_gender}
Because our models exhibits gender bias, it stands to reason that participants of varied gender may react differently to the models. Namely, non-male participants may be more sensitive to bias against women. Using point-biserial correlation tests~\cite{PBC1949}, we consider whether gender\footnote{Here, we use a binary indicator variable of whether a participant's self-reported gender included the ``male'' checkbox. We did not have enough non-binary or gender non-conforming participants to analyze separately.} correlates with our fairness perception, decision-making fairness, and decision-making quality metrics as well as the rate of ``Complete'' predictions for female and male applicants directly.

We find no significant correlations between gender and behavior or perceptions in our task (See \aptLtoX[graphic=no,type=html]{Table~\ref{tab:gender_corr}}{\autoref{tab:gender_corr}}). However, we do find marginally significant correlations with acceptance rate for female applicants, FPR, and accuracy. This shows there may be a weak trend in male participants accepting fewer female candidates ($p=0.05$), leading to a lower FPR and higher accuracy. 

\aptLtoX[graphic=no,type=html]{\begin{table*}
    \centering
    \small
    \begin{tabular}{llcc}
        \toprule
        &Metric& {Coefficient} & {$p$}\\
        \midrule
        \rowcolor{customrowcolor}Fairness Perception&Fairness Rating&0.045 & 0.403\\
        \rowcolor{customrowcolor}Fairness Perception&Fairness Saliency&-0.026 & 0.624\\
        \addlinespace
        DM Fairness&Gender Parity&  -0.058 & 0.280\\
        \addlinespace
        \rowcolor{customrowcolor}DM Quality&Accuracy& 0.094 & 0.079\\
        \rowcolor{customrowcolor}DM Quality&FNR&0.083 & 0.120\\
        \rowcolor{customrowcolor}DM Quality&FPR&  -0.107 & 0.045\\
        Acceptance Rate&\parbox{3cm}{Female Applicants}&  -0.102 & 0.057\\
        Acceptance Rate&{Male Applicants}& -0.068 & 0.204\\
        \bottomrule
    \end{tabular}
    \caption{Correlation between participants self-describing as male and various performance metrics.}
    \label{tab:gender_corr}
\end{table*}}{\begin{table}
    \centering
    \small
    \begin{tabular}{llS@{}S@{}}
        \toprule
        &Metric& {Coefficient} & {$p$}\\
        \midrule
        \rowcolor{customrowcolor}
        &Fairness Rating&0.045 & 0.403\\
        \rowcolor{customrowcolor}
        \multirow{-2}{*}{\parbox{2.5cm}{Fairness Perception}}&Fairness Saliency&-0.026 & 0.624\\
        \addlinespace
        DM Fairness&Gender Parity&  -0.058 & 0.280\\
        \addlinespace
        \rowcolor{customrowcolor}
        &Accuracy& 0.094 & 0.079\\
        \rowcolor{customrowcolor}
        &FNR&0.083 & 0.120\\
        \rowcolor{customrowcolor}
        \multirow{-3}{*}{DM Quality}&FPR&  -0.107 & 0.045\\
        &\parbox{2.2cm}{Female Applicants}&  -0.102 & 0.057\\
        \multirow{-2}{*}{\parbox{2cm}{Acceptance Rate}}&\parbox{2cm}{Male Applicants}& -0.068 & 0.204\\
        \bottomrule
    \end{tabular}
    \caption{Correlation between participants self-describing as male and various performance metrics.}
    \label{tab:gender_corr}
\end{table}}

\aptLtoX[graphic=no,type=html]{\begin{table}
    \centering
    \small
    \begin{tabular}{clcccc}%
        \toprule
        &&\multicolumn{2}{c}{Female}&\multicolumn{2}{c}{Male}\\
        \cmidrule(lr){3-4}\cmidrule(lr){5-6}
        Feature&Intervention&sig&\text{coef}&sig&\text{coef}\\
        \midrule
        \rowcolor{customrowcolor}Protected&+Expl & * & -0.13  &  & -0.04 \\
        \rowcolor{customrowcolor}Protected&+Bias Disclosure &  & 0.06  &  & 0.00 \\
        \rowcolor{customrowcolor}Protected&+Bias Disclosure with Explanation & * & 0.08  &  & 0.02 \\
        Proxy&+Expl & * & -0.07  &  & -0.00 \\
        Proxy&+Bias Disclosure &  & 0.03  &  & 0.01 \\
        Proxy&+Bias and Corr Disclosure &  & 0.08  &  & 0.01 \\
        Proxy&+Bias Disclosure with Explanation & * & 0.07  &  & -0.00 \\
        Proxy&+Bias and Corr Disclosure with Explanation & * & 0.10  &  & -0.03 \\
        \bottomrule
    \end{tabular}
    \caption{Effect of interventions on acceptance rate for female and male applicants across conditions and phases.}
    \label{tab:flip_rate}
\end{table}}{\begin{table}
    \centering
    \small
    \begin{tabular}{clcS[table-format=1.2]cS[table-format=1.2]}%
        \toprule
        &&\multicolumn{2}{c}{Female}&\multicolumn{2}{c}{Male}\\
        \cmidrule(lr){3-4}\cmidrule(lr){5-6}
        Feature&Intervention&sig&\text{coef}&sig&\text{coef}\\
        \midrule
        \rowcolor{customrowcolor}&+Expl & * & -0.13  &  & -0.04 \\
        \rowcolor{customrowcolor}&+Bias Disclosure &  & 0.06  &  & 0.00 \\
        \rowcolor{customrowcolor}\multirow{-3}{*}{Protected}&+Bias Disclosure with Expl & * & 0.08  &  & 0.02 \\
        &+Expl & * & -0.07  &  & -0.00 \\
        &+Bias Disclosure &  & 0.03  &  & 0.01 \\
        &+Bias and Corr Disclosure &  & 0.08  &  & 0.01 \\
        &+Bias Disclosure with Expl & * & 0.07  &  & -0.00 \\
        &+Bias and Corr&&&&\\
        \multirow{-6}{*}{Proxy}&\phantom{+}Disclosure with Expl & * & 0.10  &  & -0.03 \\
        \bottomrule
    \end{tabular}
    \caption{Effect of interventions on acceptance rate for female and male applicants across conditions and phases.}
    \label{tab:flip_rate}
\end{table}}

\newpage
\onecolumn

%

\aptLtoX[graphic=no,type=html]{}{
\centering{\small \centering
    \begin{tabular}{lllrrlrc}
        \toprule
            Metric&Feature&Effect& \multicolumn{1}{c}{$F$}&\multicolumn{1}{c}{$p$}&&Coef&Std Error\\
        \midrule
        &\cellcolor{customrowcolor}\ &\cellcolor{customrowcolor}$+$Expl&\cellcolor{customrowcolor}$F(1,115)=26.33$&\cellcolor{customrowcolor}$0.0000$&\cellcolor{customrowcolor}$*$&\cellcolor{customrowcolor}$-0.89$&\cellcolor{customrowcolor}$0.174$\\
&\cellcolor{customrowcolor}\ &\cellcolor{customrowcolor}$+$Bias Disclosure&\cellcolor{customrowcolor}$F(1,50)=5.43$&\cellcolor{customrowcolor}$0.0238$&\cellcolor{customrowcolor}$ $&\cellcolor{customrowcolor}$-0.29$&\cellcolor{customrowcolor}$0.126$\\
&\cellcolor{customrowcolor}\ &\cellcolor{customrowcolor}$+$Bias Disclosure with Explanation&\cellcolor{customrowcolor}$F(1,67)=0.01$&\cellcolor{customrowcolor}$0.9091$&\cellcolor{customrowcolor}$ $&\cellcolor{customrowcolor}$-0.01$&\cellcolor{customrowcolor}$0.128$\\
        &\cellcolor{customrowcolor}\multirow{-4}{*}{protected}&\cellcolor{customrowcolor}Joint Intervention (BD)&\cellcolor{customrowcolor}$F(1,208)=73.06$&\cellcolor{customrowcolor}$0.0000$&\cellcolor{customrowcolor}$*$&\cellcolor{customrowcolor}$-1.01$&\cellcolor{customrowcolor}$0.118$\\
        &&$+$Expl&$F(1,227)=1.51$&$0.2207$&$ $&$0.11$&$0.088$\\
        &&$+$Bias Disclosure&$F(1,122)=3.54$&$0.0623$&$ $&$-0.21$&$0.114$\\
        &&$+$Bias and Corr Disclosure&$F(1,125)=12.84$&$0.0005$&$*$&$-0.42$&$0.116$\\
        &&$+$Bias Disclosure with Explanation&$F(1,178)=19.84$&$0.0000$&$*$&$-0.43$&$0.096$\\
        &&$+$Bias and Corr Disclosure with Explanation&$F(1,178)=46.57$&$0.0000$&$*$&$-0.66$&$0.096$\\
        &&Joint Intervention (BD)&$F(1,277)=6.14$&$0.0138$&$*$&$-0.27$&$0.109$\\
        \multirow{-11}{*}{\parbox{1.5cm}{Fairness\\Rating}}&\multirow{-7}{*}{proxy}&Joint Intervention (BD+CD)&$F(1,277)=13.19$&$0.0003$&$*$&$-0.46$&$0.109$\\
        \midrule
&\cellcolor{customrowcolor}\ &\cellcolor{customrowcolor}$+$Expl&\cellcolor{customrowcolor}$F(1,115)=15.83$&\cellcolor{customrowcolor}$0.0001$&\cellcolor{customrowcolor}$*$&\cellcolor{customrowcolor}$0.33$&\cellcolor{customrowcolor}$0.083$\\
&\cellcolor{customrowcolor}\ &\cellcolor{customrowcolor}$+$Bias Disclosure&\cellcolor{customrowcolor}$F(1,50)=2.33$&\cellcolor{customrowcolor}$0.1330$&\cellcolor{customrowcolor}$ $&\cellcolor{customrowcolor}$0.10$&\cellcolor{customrowcolor}$0.064$\\
&\cellcolor{customrowcolor}\ &\cellcolor{customrowcolor}$+$Bias Disclosure with Explanation&\cellcolor{customrowcolor}$F(1,67)=1.19$&\cellcolor{customrowcolor}$0.2785$&\cellcolor{customrowcolor}$ $&\cellcolor{customrowcolor}$0.07$&\cellcolor{customrowcolor}$0.067$\\
&\cellcolor{customrowcolor}\multirow{-4}{*}{protected}&\cellcolor{customrowcolor}Joint Intervention (BD)&\cellcolor{customrowcolor}$F(1,208)=72.7$&\cellcolor{customrowcolor}$0.0000$&\cellcolor{customrowcolor}$*$&\cellcolor{customrowcolor}$0.47$&\cellcolor{customrowcolor}$0.055$\\
        &&$+$Expl&$F(1,227)=1.21$&$0.2733$&$ $&$-0.03$&$0.028$\\
        &&$+$Bias Disclosure&$F(1,122)=0.67$&$0.4154$&$ $&$0.04$&$0.050$\\
        &&$+$Bias and Corr Disclosure&$F(1,125)=9.19$&$0.0030$&$*$&$0.16$&$0.052$\\
        &&$+$Bias Disclosure with Explanation&$F(1,182)=0.62$&$0.4311$&$ $&$0.03$&$0.037$\\
        &&$+$Bias and Corr Disclosure with Explanation&$F(1,182)=35.06$&$0.0000$&$*$&$0.22$&$0.037$\\
        &&Joint Intervention (BD)&$F(1,277)=0.58$&$0.4458$&$ $&$-0.04$&$0.047$\\
        \multirow{-11}{*}{\parbox{1.5cm}{Fairness\\Saliency}}&\multirow{-7}{*}{proxy}&Joint Intervention (BD+CD)&$F(1,277)=13.42$&$0.0003$&$*$&$0.15$&$0.048$\\
        \midrule
&\cellcolor{customrowcolor}&\cellcolor{customrowcolor}$+$Expl&\cellcolor{customrowcolor}$F(1,115)=8.32$&\cellcolor{customrowcolor}$0.0047$&\cellcolor{customrowcolor}$*$&\cellcolor{customrowcolor}$-0.13$&\cellcolor{customrowcolor}$0.044$\\
&\cellcolor{customrowcolor}&\cellcolor{customrowcolor}$+$Bias Disclosure&\cellcolor{customrowcolor}$F(1,99)=2.58$&\cellcolor{customrowcolor}$0.1117$&\cellcolor{customrowcolor}$ $&\cellcolor{customrowcolor}$0.09$&\cellcolor{customrowcolor}$0.055$\\
&\cellcolor{customrowcolor}&\cellcolor{customrowcolor}$+$Bias Disclosure with Explanation&\cellcolor{customrowcolor}$F(1,133)=4.08$&\cellcolor{customrowcolor}$0.0455$&\cellcolor{customrowcolor}$ $&\cellcolor{customrowcolor}$0.09$&\cellcolor{customrowcolor}$0.047$\\
&\cellcolor{customrowcolor}\multirow{-4}{*}{protected}&\cellcolor{customrowcolor}Joint Intervention (BD)&\cellcolor{customrowcolor}$F(1,208)=0.06$&\cellcolor{customrowcolor}$0.8096$&\cellcolor{customrowcolor}$ $&\cellcolor{customrowcolor}$0.01$&\cellcolor{customrowcolor}$0.042$\\
        &&$+$Expl&$F(1,227)=8.58$&$0.0038$&$*$&$-0.10$&$0.035$\\
        &&$+$Bias Disclosure&$F(1,182)=0.77$&$0.3815$&$ $&$0.04$&$0.050$\\
        &&$+$Bias and Corr Disclosure&$F(1,182)=1.29$&$0.2574$&$ $&$0.06$&$0.051$\\
        &&$+$Bias Disclosure with Explanation&$F(1,188)=7.75$&$0.0059$&$*$&$0.10$&$0.035$\\
        &&$+$Bias and Corr Disclosure with Explanation&$F(1,188)=20.26$&$0.0000$&$*$&$0.16$&$0.035$\\
        &&Joint Intervention (BD)&$F(1,277)=0.78$&$0.3785$&$ $&$-0.03$&$0.038$\\
        \multirow{-11}{*}{\parbox{1.5cm}{Parity}}&\multirow{-7}{*}{proxy}&Joint Intervention (BD+CD)&$F(1,277)=1.01$&$0.3149$&$ $&$0.03$&$0.038$\\
        \midrule
        \multicolumn{8}{r}{{(\textit{table continues})}}\\
    \end{tabular}}

\newpage

\centering{\small \centering\begin{tabular}{lllrrlrc}
        \toprule
            Metric&Feature&Effect& \multicolumn{1}{c}{$F$}&\multicolumn{1}{c}{$p$}&&Coef&Std Error\\
        \midrule
&\cellcolor{customrowcolor}&\cellcolor{customrowcolor}$+$Expl&\cellcolor{customrowcolor}$F(1,115)=20.89$&\cellcolor{customrowcolor}$0.0000$&\cellcolor{customrowcolor}$*$&\cellcolor{customrowcolor}$0.04$&\cellcolor{customrowcolor}$0.009$\\
&\cellcolor{customrowcolor}&\cellcolor{customrowcolor}$+$Bias Disclosure&\cellcolor{customrowcolor}$F(1,50)=1.28$&\cellcolor{customrowcolor}$0.2630$&\cellcolor{customrowcolor}$ $&\cellcolor{customrowcolor}$-0.01$&\cellcolor{customrowcolor}$0.009$\\
&\cellcolor{customrowcolor}&\cellcolor{customrowcolor}$+$Bias Disclosure with Explanation&\cellcolor{customrowcolor}$F(1,67)=0.79$&\cellcolor{customrowcolor}$0.3772$&\cellcolor{customrowcolor}$ $&\cellcolor{customrowcolor}$-0.01$&\cellcolor{customrowcolor}$0.006$\\
&\cellcolor{customrowcolor}\multirow{-4}{*}{protected}&\cellcolor{customrowcolor}Joint Intervention (BD)&\cellcolor{customrowcolor}$F(1,208)=3.5$&\cellcolor{customrowcolor}$0.0628$&\cellcolor{customrowcolor}$ $&\cellcolor{customrowcolor}$0.01$&\cellcolor{customrowcolor}$0.007$\\
        &&$+$Expl&$F(1,227)=7.3$&$0.0074$&$*$&$0.02$&$0.006$\\
        &&$+$Bias Disclosure&$F(1,123)=0.02$&$0.8846$&$ $&$0.00$&$0.009$\\
        &&$+$Bias and Corr Disclosure&$F(1,126)=6.66$&$0.0110$&$*$&$-0.02$&$0.010$\\
        &&$+$Bias Disclosure with Explanation&$F(1,182)=9.18$&$0.0028$&$*$&$-0.02$&$0.006$\\
        &&$+$Bias and Corr Disclosure with Explanation&$F(1,182)=10.26$&$0.0016$&$*$&$-0.02$&$0.006$\\
        &&Joint Intervention (BD)&$F(1,277)=2.44$&$0.1197$&$ $&$0.01$&$0.007$\\
        \multirow{-11}{*}{\parbox{1.5cm}{Accuracy}}&\multirow{-7}{*}{proxy}&Joint Intervention (BD+CD)&$F(1,277)=0.73$&$0.3933$&$ $&$0.01$&$0.007$\\
        \midrule
&\cellcolor{customrowcolor}&\cellcolor{customrowcolor}$+$Expl&\cellcolor{customrowcolor}$F(1,115)=5.88$&\cellcolor{customrowcolor}$0.0169$&\cellcolor{customrowcolor}$*$&\cellcolor{customrowcolor}$0.07$&\cellcolor{customrowcolor}$0.028$\\
&\cellcolor{customrowcolor}&\cellcolor{customrowcolor}$+$Bias Disclosure&\cellcolor{customrowcolor}$F(1,50)=0.38$&\cellcolor{customrowcolor}$0.5426$&\cellcolor{customrowcolor}$ $&\cellcolor{customrowcolor}$-0.02$&\cellcolor{customrowcolor}$0.026$\\
&\cellcolor{customrowcolor}&\cellcolor{customrowcolor}$+$Bias Disclosure with Explanation&\cellcolor{customrowcolor}$F(1,67)=7.78$&\cellcolor{customrowcolor}$0.0069$&\cellcolor{customrowcolor}$*$&\cellcolor{customrowcolor}$-0.05$&\cellcolor{customrowcolor}$0.017$\\
&\cellcolor{customrowcolor}\multirow{-4}{*}{protected}&\cellcolor{customrowcolor}Joint Intervention (BD)&\cellcolor{customrowcolor}$F(1,208)=0.05$&\cellcolor{customrowcolor}$0.8269$&\cellcolor{customrowcolor}$ $&\cellcolor{customrowcolor}$0.00$&\cellcolor{customrowcolor}$0.022$\\
        &&$+$Expl&$F(1,227)=3.78$&$0.0530$&$ $&$0.03$&$0.017$\\
        &&$+$Bias Disclosure&$F(1,113)=0.48$&$0.4921$&$ $&$-0.01$&$0.018$\\
        &&$+$Bias and Corr Disclosure&$F(1,115)=2.95$&$0.0887$&$ $&$-0.03$&$0.019$\\
        &&$+$Bias Disclosure with Explanation&$F(1,169)=1.6$&$0.2073$&$ $&$-0.02$&$0.014$\\
        &&$+$Bias and Corr Disclosure with Explanation&$F(1,169)=1.65$&$0.2006$&$ $&$-0.02$&$0.014$\\
        &&Joint Intervention (BD)&$F(1,277)=1.21$&$0.2722$&$ $&$0.02$&$0.020$\\
        \multirow{-11}{*}{\parbox{1.5cm}{FNR}}&\multirow{-7}{*}{proxy}&Joint Intervention (BD+CD)&$F(1,277)=1.39$&$0.2396$&$ $&$0.03$&$0.020$\\
        \midrule
&\cellcolor{customrowcolor}&\cellcolor{customrowcolor}$+$Expl&\cellcolor{customrowcolor}$F(1,115)=9.94$&\cellcolor{customrowcolor}$0.0021$&\cellcolor{customrowcolor}$*$&\cellcolor{customrowcolor}$-0.10$&\cellcolor{customrowcolor}$0.031$\\
&\cellcolor{customrowcolor}&\cellcolor{customrowcolor}$+$Bias Disclosure&\cellcolor{customrowcolor}$F(1,50)=2.09$&\cellcolor{customrowcolor}$0.1546$&\cellcolor{customrowcolor}$ $&\cellcolor{customrowcolor}$0.04$&\cellcolor{customrowcolor}$0.028$\\
&\cellcolor{customrowcolor}&\cellcolor{customrowcolor}$+$Bias Disclosure with Explanation&\cellcolor{customrowcolor}$F(1,67)=8.35$&\cellcolor{customrowcolor}$0.0052$&\cellcolor{customrowcolor}$*$&\cellcolor{customrowcolor}$0.05$&\cellcolor{customrowcolor}$0.019$\\
&\cellcolor{customrowcolor}\multirow{-4}{*}{protected}&\cellcolor{customrowcolor}Joint Intervention (BD)&\cellcolor{customrowcolor}$F(1,208)=0.17$&\cellcolor{customrowcolor}$0.6832$&\cellcolor{customrowcolor}$ $&\cellcolor{customrowcolor}$-0.01$&\cellcolor{customrowcolor}$0.025$\\
        &&$+$Expl&$F(1,227)=5.45$&$0.0205$&$ $&$-0.04$&$0.019$\\
        &&$+$Bias Disclosure&$F(1,116)=1.34$&$0.2502$&$ $&$0.03$&$0.024$\\
        &&$+$Bias and Corr Disclosure&$F(1,118)=5.95$&$0.0162$&$*$&$0.06$&$0.024$\\
        &&$+$Bias Disclosure with Explanation&$F(1,169)=10.79$&$0.0012$&$*$&$0.05$&$0.017$\\
        &&$+$Bias and Corr Disclosure with Explanation&$F(1,169)=7.44$&$0.0071$&$*$&$0.05$&$0.017$\\
        &&Joint Intervention (BD)&$F(1,277)=0.06$&$0.8092$&$ $&$-0.01$&$0.023$\\
        \multirow{-11}{*}{\parbox{1.5cm}{FPR}}&\multirow{-7}{*}{proxy}&Joint Intervention (BD+CD)&$F(1,277)=0.84$&$0.3593$&$ $&$-0.02$&$0.023$\\
        \midrule
        \multicolumn{8}{r}{{(\textit{table continues})}}\\
    \end{tabular}}
        \newpage

\begin{table*}
{\small \centering\begin{tabular}{lllrrlrc}
        \toprule
            Metric&Feature&Effect& \multicolumn{1}{c}{$F$}&\multicolumn{1}{c}{$p$}&&Coef&Std Error\\
        \midrule
&\cellcolor{customrowcolor}&\cellcolor{customrowcolor}$+$Expl&\cellcolor{customrowcolor}$F(1,116)=2.57$&\cellcolor{customrowcolor}$0.1117$&\cellcolor{customrowcolor}&\cellcolor{customrowcolor}$-0.32$&\cellcolor{customrowcolor}$0.199$\\
&\cellcolor{customrowcolor}&\cellcolor{customrowcolor}$+$Bias Disclosure&\cellcolor{customrowcolor}$F(1,50)=0.39$&\cellcolor{customrowcolor}$0.5369$&\cellcolor{customrowcolor}&\cellcolor{customrowcolor}$-0.06$&\cellcolor{customrowcolor}$0.095$\\
&\cellcolor{customrowcolor}\multirow{-3}{*}{protected}&\cellcolor{customrowcolor}$+$Bias Disclosure with Explanation&\cellcolor{customrowcolor}$F(1,67)=1.06$&\cellcolor{customrowcolor}$0.3070$&\cellcolor{customrowcolor}&\cellcolor{customrowcolor}$-0.09$&\cellcolor{customrowcolor}$0.086$\\
        &&$+$Expl&$F(1,228)=0.0$&$0.9726$&&$0.00$&$0.143$\\
        &&$+$Bias Disclosure&$F(1,101)=5.0$&$0.0276$&&$-0.21$&$0.095$\\
        &&$+$Bias and Corr Disclosure&$F(1,102)=0.77$&$0.3838$&&$-0.09$&$0.098$\\
        &&$+$Bias Disclosure with Explanation&$F(1,151)=0.07$&$0.7884$&&$0.02$&$0.085$\\
        \multirow{-8}{*}{\parbox{1.5cm}{better}}&\multirow{-5}{*}{proxy}&$+$Bias and Corr Disclosure with Explanation&$F(1,151)=10.12$&$0.0018$&$*$&$-0.27$&$0.085$\\
        \midrule
&\cellcolor{customrowcolor}&\cellcolor{customrowcolor}$+$Expl&\cellcolor{customrowcolor}$F(1,116)=1.71$&\cellcolor{customrowcolor}$0.1933$&\cellcolor{customrowcolor}&\cellcolor{customrowcolor}$-0.23$&\cellcolor{customrowcolor}$0.176$\\
&\cellcolor{customrowcolor}&\cellcolor{customrowcolor}$+$Bias Disclosure&\cellcolor{customrowcolor}$F(1,50)=0.04$&\cellcolor{customrowcolor}$0.8496$&\cellcolor{customrowcolor}&\cellcolor{customrowcolor}$-0.02$&\cellcolor{customrowcolor}$0.103$\\
&\cellcolor{customrowcolor}\multirow{-3}{*}{protected}&\cellcolor{customrowcolor}$+$Bias Disclosure with Explanation&\cellcolor{customrowcolor}$F(1,67)=1.2$&\cellcolor{customrowcolor}$0.2765$&\cellcolor{customrowcolor}&\cellcolor{customrowcolor}$-0.09$&\cellcolor{customrowcolor}$0.080$\\
        &&$+$Expl&$F(1,228)=1.41$&$0.2356$&&$0.16$&$0.131$\\
        &&$+$Bias Disclosure&$F(1,109)=0.5$&$0.4798$&&$-0.08$&$0.106$\\
        &&$+$Bias and Corr Disclosure&$F(1,110)=3.66$&$0.0583$&&$-0.21$&$0.109$\\
        &&$+$Bias Disclosure with Explanation&$F(1,149)=13.17$&$0.0004$&$*$&$-0.28$&$0.076$\\
        \multirow{-8}{*}{\parbox{1.5cm}{confidence}}&\multirow{-5}{*}{proxy}&$+$Bias and Corr Disclosure with Explanation&$F(1,149)=19.32$&$0.0000$&$*$&$-0.33$&$0.076$\\
        \midrule
&\cellcolor{customrowcolor}&\cellcolor{customrowcolor}$+$Expl&\cellcolor{customrowcolor}$F(1,116)=0.48$&\cellcolor{customrowcolor}$0.4883$&\cellcolor{customrowcolor}&\cellcolor{customrowcolor}$0.14$&\cellcolor{customrowcolor}$0.204$\\
&\cellcolor{customrowcolor}&\cellcolor{customrowcolor}$+$Bias Disclosure&\cellcolor{customrowcolor}$F(1,50)=0.0$&\cellcolor{customrowcolor}$1.0000$&\cellcolor{customrowcolor}&\cellcolor{customrowcolor}$0.00$&\cellcolor{customrowcolor}$0.125$\\
&\cellcolor{customrowcolor}\multirow{-3}{*}{protected}&\cellcolor{customrowcolor}$+$Bias Disclosure with Explanation&\cellcolor{customrowcolor}$F(1,67)=0.01$&\cellcolor{customrowcolor}$0.9038$&\cellcolor{customrowcolor}&\cellcolor{customrowcolor}$-0.01$&\cellcolor{customrowcolor}$0.121$\\
        &&$+$Expl&$F(1,228)=18.55$&$0.0000$&$*$&$0.69$&$0.161$\\
        &&$+$Bias Disclosure&$F(1,103)=0.05$&$0.8298$&&$0.02$&$0.110$\\
        &&$+$Bias and Corr Disclosure&$F(1,103)=6.39$&$0.0130$&$*$&$0.29$&$0.113$\\
        &&$+$Bias Disclosure with Explanation&$F(1,156)=0.2$&$0.6559$&&$0.05$&$0.110$\\
        \multirow{-8}{*}{\parbox{1.5cm}{predictable}}&\multirow{-5}{*}{proxy}&$+$Bias and Corr Disclosure with Explanation&$F(1,156)=1.22$&$0.2716$&&$-0.12$&$0.110$\\
        \midrule
&\cellcolor{customrowcolor}&\cellcolor{customrowcolor}$+$Expl&\cellcolor{customrowcolor}$F(1,116)=0.03$&\cellcolor{customrowcolor}$0.8677$&\cellcolor{customrowcolor}&\cellcolor{customrowcolor}$0.03$&\cellcolor{customrowcolor}$0.176$\\
&\cellcolor{customrowcolor}&\cellcolor{customrowcolor}$+$Bias Disclosure&\cellcolor{customrowcolor}$F(1,50)=1.69$&\cellcolor{customrowcolor}$0.1997$&\cellcolor{customrowcolor}&\cellcolor{customrowcolor}$-0.10$&\cellcolor{customrowcolor}$0.075$\\
&\cellcolor{customrowcolor}\multirow{-3}{*}{protected}&\cellcolor{customrowcolor}$+$Bias Disclosure with Explanation&\cellcolor{customrowcolor}$F(1,67)=2.96$&\cellcolor{customrowcolor}$0.0898$&\cellcolor{customrowcolor}&\cellcolor{customrowcolor}$-0.18$&\cellcolor{customrowcolor}$0.103$\\
        &&$+$Expl&$F(1,228)=4.17$&$0.0422$&&$0.24$&$0.119$\\
        &&$+$Bias Disclosure&$F(1,107)=0.08$&$0.7787$&&$0.03$&$0.099$\\
        &&$+$Bias and Corr Disclosure&$F(1,108)=0.01$&$0.9419$&&$-0.01$&$0.102$\\
        &&$+$Bias Disclosure with Explanation&$F(1,158)=3.27$&$0.0727$&&$-0.17$&$0.092$\\
        \multirow{-8}{*}{\parbox{1.5cm}{safe}}&\multirow{-5}{*}{proxy}&$+$Bias and Corr Disclosure with Explanation&$F(1,158)=7.76$&$0.0060$&$*$&$-0.25$&$0.092$\\
        \midrule
&\cellcolor{customrowcolor}&\cellcolor{customrowcolor}$+$Expl&\cellcolor{customrowcolor}$F(1,116)=0.12$&\cellcolor{customrowcolor}$0.7283$&\cellcolor{customrowcolor}&\cellcolor{customrowcolor}$0.06$&\cellcolor{customrowcolor}$0.183$\\
&\cellcolor{customrowcolor}&\cellcolor{customrowcolor}$+$Bias Disclosure&\cellcolor{customrowcolor}$F(1,50)=0.0$&\cellcolor{customrowcolor}$1.0000$&\cellcolor{customrowcolor}&\cellcolor{customrowcolor}$0.00$&\cellcolor{customrowcolor}$0.097$\\
&\cellcolor{customrowcolor}\multirow{-3}{*}{protected}&\cellcolor{customrowcolor}$+$Bias Disclosure with Explanation&\cellcolor{customrowcolor}$F(1,67)=1.6$&\cellcolor{customrowcolor}$0.2100$&\cellcolor{customrowcolor}&\cellcolor{customrowcolor}$0.13$&\cellcolor{customrowcolor}$0.105$\\
        &&$+$Expl&$F(1,228)=0.0$&$0.9621$&&$0.01$&$0.118$\\
        &&$+$Bias Disclosure&$F(1,113)=4.48$&$0.0365$&&$0.24$&$0.114$\\
        &&$+$Bias and Corr Disclosure&$F(1,115)=1.05$&$0.3066$&&$0.12$&$0.117$\\
        &&$+$Bias Disclosure with Explanation&$F(1,153)=2.69$&$0.1030$&&$0.13$&$0.080$\\
        \multirow{-8}{*}{\parbox{1.5cm}{wary}}&\multirow{-5}{*}{proxy}&$+$Bias and Corr Disclosure with Explanation&$F(1,153)=13.28$&$0.0004$&$*$&$0.29$&$0.080$\\
        \midrule
&\cellcolor{customrowcolor}&\cellcolor{customrowcolor}$+$Expl&\cellcolor{customrowcolor}$F(1,116)=6.0$&\cellcolor{customrowcolor}$0.0158$&\cellcolor{customrowcolor}$*$&\cellcolor{customrowcolor}$-0.38$&\cellcolor{customrowcolor}$0.154$\\
&\cellcolor{customrowcolor}&\cellcolor{customrowcolor}$+$Bias Disclosure&\cellcolor{customrowcolor}$F(1,50)=1.5$&\cellcolor{customrowcolor}$0.2265$&\cellcolor{customrowcolor}&\cellcolor{customrowcolor}$-0.14$&\cellcolor{customrowcolor}$0.112$\\
&\cellcolor{customrowcolor}\multirow{-3}{*}{protected}&\cellcolor{customrowcolor}$+$Bias Disclosure with Explanation&\cellcolor{customrowcolor}$F(1,67)=2.56$&\cellcolor{customrowcolor}$0.1145$&\cellcolor{customrowcolor}&\cellcolor{customrowcolor}$-0.15$&\cellcolor{customrowcolor}$0.092$\\
        &&$+$Expl&$F(1,228)=0.06$&$0.8122$&&$0.03$&$0.138$\\
        &&$+$Bias Disclosure&$F(1,105)=4.43$&$0.0376$&&$-0.22$&$0.104$\\
        &&$+$Bias and Corr Disclosure&$F(1,106)=14.58$&$0.0002$&$*$&$-0.41$&$0.107$\\
        &&$+$Bias Disclosure with Explanation&$F(1,153)=5.43$&$0.0211$&&$-0.20$&$0.088$\\
        \multirow{-8}{*}{\parbox{1.5cm}{works}}&\multirow{-5}{*}{proxy}&$+$Bias and Corr Disclosure with Explanation&$F(1,153)=14.19$&$0.0002$&$*$&$-0.33$&$0.088$\\
        \bottomrule
        \end{tabular}}
        \caption{Overall results of tests regarding the primary effects of our study on fairness perceptions, decision-making fairness, decision-making quality, and learned trust.}
        \label{tab:full}
    \end{table*}
}

\aptLtoX[graphic=no,type=html]{
\begin{table*}{\small \centering
    \begin{tabular}{lllrrlrc}
        \toprule
            Metric&Feature&Effect& \multicolumn{1}{c}{$F$}&\multicolumn{1}{c}{$p$}&&Coef&Std Error\\
        \midrule
        Fairness Rating&protected\cellcolor{customrowcolor}\ &\cellcolor{customrowcolor}$+$Expl&\cellcolor{customrowcolor}$F(1,115)=26.33$&\cellcolor{customrowcolor}$0.0000$&\cellcolor{customrowcolor}$*$&\cellcolor{customrowcolor}$-0.89$&\cellcolor{customrowcolor}$0.174$\\
        Fairness Rating&protected\cellcolor{customrowcolor}\ &\cellcolor{customrowcolor}$+$Bias Disclosure&\cellcolor{customrowcolor}$F(1,50)=5.43$&\cellcolor{customrowcolor}$0.0238$&\cellcolor{customrowcolor}$ $&\cellcolor{customrowcolor}$-0.29$&\cellcolor{customrowcolor}$0.126$\\
        Fairness Rating&protected\cellcolor{customrowcolor}\ &\cellcolor{customrowcolor}$+$Bias Disclosure with Explanation&\cellcolor{customrowcolor}$F(1,67)=0.01$&\cellcolor{customrowcolor}$0.9091$&\cellcolor{customrowcolor}$ $&\cellcolor{customrowcolor}$-0.01$&\cellcolor{customrowcolor}$0.128$\\
        Fairness Rating&\cellcolor{customrowcolor}protected&\cellcolor{customrowcolor}Joint Intervention (BD)&\cellcolor{customrowcolor}$F(1,208)=73.06$&\cellcolor{customrowcolor}$0.0000$&\cellcolor{customrowcolor}$*$&\cellcolor{customrowcolor}$-1.01$&\cellcolor{customrowcolor}$0.118$\\
        Fairness Rating&proxy&$+$Expl&$F(1,227)=1.51$&$0.2207$&$ $&$0.11$&$0.088$\\
        Fairness Rating&proxy&$+$Bias Disclosure&$F(1,122)=3.54$&$0.0623$&$ $&$-0.21$&$0.114$\\
        Fairness Rating&proxy&$+$Bias and Corr Disclosure&$F(1,125)=12.84$&$0.0005$&$*$&$-0.42$&$0.116$\\
        Fairness Rating&proxy&$+$Bias Disclosure with Explanation&$F(1,178)=19.84$&$0.0000$&$*$&$-0.43$&$0.096$\\
        Fairness Rating&proxy&$+$Bias and Corr Disclosure with Explanation&$F(1,178)=46.57$&$0.0000$&$*$&$-0.66$&$0.096$\\
        Fairness Rating&proxy&Joint Intervention (BD)&$F(1,277)=6.14$&$0.0138$&$*$&$-0.27$&$0.109$\\
        Fairness Rating&proxy&Joint Intervention (BD+CD)&$F(1,277)=13.19$&$0.0003$&$*$&$-0.46$&$0.109$\\
        \midrule
        Fairness Saliency&protected\cellcolor{customrowcolor}\ &\cellcolor{customrowcolor}$+$Expl&\cellcolor{customrowcolor}$F(1,115)=15.83$&\cellcolor{customrowcolor}$0.0001$&\cellcolor{customrowcolor}$*$&\cellcolor{customrowcolor}$0.33$&\cellcolor{customrowcolor}$0.083$\\
        Fairness Saliency&protected\cellcolor{customrowcolor}\ &\cellcolor{customrowcolor}$+$Bias Disclosure&\cellcolor{customrowcolor}$F(1,50)=2.33$&\cellcolor{customrowcolor}$0.1330$&\cellcolor{customrowcolor}$ $&\cellcolor{customrowcolor}$0.10$&\cellcolor{customrowcolor}$0.064$\\
        Fairness Saliency&protected\cellcolor{customrowcolor}\ &\cellcolor{customrowcolor}$+$Bias Disclosure with Explanation&\cellcolor{customrowcolor}$F(1,67)=1.19$&\cellcolor{customrowcolor}$0.2785$&\cellcolor{customrowcolor}$ $&\cellcolor{customrowcolor}$0.07$&\cellcolor{customrowcolor}$0.067$\\
        Fairness Saliency&\cellcolor{customrowcolor}protected&\cellcolor{customrowcolor}Joint Intervention (BD)&\cellcolor{customrowcolor}$F(1,208)=72.7$&\cellcolor{customrowcolor}$0.0000$&\cellcolor{customrowcolor}$*$&\cellcolor{customrowcolor}$0.47$&\cellcolor{customrowcolor}$0.055$\\
        Fairness Saliency&proxy&$+$Expl&$F(1,227)=1.21$&$0.2733$&$ $&$-0.03$&$0.028$\\
        Fairness Saliency&proxy&$+$Bias Disclosure&$F(1,122)=0.67$&$0.4154$&$ $&$0.04$&$0.050$\\
        Fairness Saliency&proxy&$+$Bias and Corr Disclosure&$F(1,125)=9.19$&$0.0030$&$*$&$0.16$&$0.052$\\
        Fairness Saliency&proxy&$+$Bias Disclosure with Explanation&$F(1,182)=0.62$&$0.4311$&$ $&$0.03$&$0.037$\\
        Fairness Saliency&proxy&$+$Bias and Corr Disclosure with Explanation&$F(1,182)=35.06$&$0.0000$&$*$&$0.22$&$0.037$\\
        Fairness Saliency&proxy&Joint Intervention (BD)&$F(1,277)=0.58$&$0.4458$&$ $&$-0.04$&$0.047$\\
        Fairness Saliency&proxy&Joint Intervention (BD+CD)&$F(1,277)=13.42$&$0.0003$&$*$&$0.15$&$0.048$\\
        \midrule
        Parity&protected\cellcolor{customrowcolor}&\cellcolor{customrowcolor}$+$Expl&\cellcolor{customrowcolor}$F(1,115)=8.32$&\cellcolor{customrowcolor}$0.0047$&\cellcolor{customrowcolor}$*$&\cellcolor{customrowcolor}$-0.13$&\cellcolor{customrowcolor}$0.044$\\
        Parity&protected\cellcolor{customrowcolor}&\cellcolor{customrowcolor}$+$Bias Disclosure&\cellcolor{customrowcolor}$F(1,99)=2.58$&\cellcolor{customrowcolor}$0.1117$&\cellcolor{customrowcolor}$ $&\cellcolor{customrowcolor}$0.09$&\cellcolor{customrowcolor}$0.055$\\
        Parity&protected\cellcolor{customrowcolor}&\cellcolor{customrowcolor}$+$Bias Disclosure with Explanation&\cellcolor{customrowcolor}$F(1,133)=4.08$&\cellcolor{customrowcolor}$0.0455$&\cellcolor{customrowcolor}$ $&\cellcolor{customrowcolor}$0.09$&\cellcolor{customrowcolor}$0.047$\\
        Parity&protected\cellcolor{customrowcolor}&\cellcolor{customrowcolor}Joint Intervention (BD)&\cellcolor{customrowcolor}$F(1,208)=0.06$&\cellcolor{customrowcolor}$0.8096$&\cellcolor{customrowcolor}$ $&\cellcolor{customrowcolor}$0.01$&\cellcolor{customrowcolor}$0.042$\\
        Parity&proxy&$+$Expl&$F(1,227)=8.58$&$0.0038$&$*$&$-0.10$&$0.035$\\
        Parity&proxy&$+$Bias Disclosure&$F(1,182)=0.77$&$0.3815$&$ $&$0.04$&$0.050$\\
        Parity&proxy&$+$Bias and Corr Disclosure&$F(1,182)=1.29$&$0.2574$&$ $&$0.06$&$0.051$\\
        Parity&proxy&$+$Bias Disclosure with Explanation&$F(1,188)=7.75$&$0.0059$&$*$&$0.10$&$0.035$\\
        Parity&proxy&$+$Bias and Corr Disclosure with Explanation&$F(1,188)=20.26$&$0.0000$&$*$&$0.16$&$0.035$\\
        Parity&proxy&Joint Intervention (BD)&$F(1,277)=0.78$&$0.3785$&$ $&$-0.03$&$0.038$\\
        Parity&proxy&Joint Intervention (BD+CD)&$F(1,277)=1.01$&$0.3149$&$ $&$0.03$&$0.038$\\
        \midrule
        Accuracy&protected\cellcolor{customrowcolor}&\cellcolor{customrowcolor}$+$Expl&\cellcolor{customrowcolor}$F(1,115)=20.89$&\cellcolor{customrowcolor}$0.0000$&\cellcolor{customrowcolor}$*$&\cellcolor{customrowcolor}$0.04$&\cellcolor{customrowcolor}$0.009$\\
        Accuracy&protected\cellcolor{customrowcolor}&\cellcolor{customrowcolor}$+$Bias Disclosure&\cellcolor{customrowcolor}$F(1,50)=1.28$&\cellcolor{customrowcolor}$0.2630$&\cellcolor{customrowcolor}$ $&\cellcolor{customrowcolor}$-0.01$&\cellcolor{customrowcolor}$0.009$\\
        Accuracy&protected\cellcolor{customrowcolor}&\cellcolor{customrowcolor}$+$Bias Disclosure with Explanation&\cellcolor{customrowcolor}$F(1,67)=0.79$&\cellcolor{customrowcolor}$0.3772$&\cellcolor{customrowcolor}$ $&\cellcolor{customrowcolor}$-0.01$&\cellcolor{customrowcolor}$0.006$\\
        Accuracy&protected\cellcolor{customrowcolor}&\cellcolor{customrowcolor}Joint Intervention (BD)&\cellcolor{customrowcolor}$F(1,208)=3.5$&\cellcolor{customrowcolor}$0.0628$&\cellcolor{customrowcolor}$ $&\cellcolor{customrowcolor}$0.01$&\cellcolor{customrowcolor}$0.007$\\
        Accuracy&proxy&$+$Expl&$F(1,227)=7.3$&$0.0074$&$*$&$0.02$&$0.006$\\
        Accuracy&proxy&$+$Bias Disclosure&$F(1,123)=0.02$&$0.8846$&$ $&$0.00$&$0.009$\\
        Accuracy&proxy&$+$Bias and Corr Disclosure&$F(1,126)=6.66$&$0.0110$&$*$&$-0.02$&$0.010$\\
        Accuracy&proxy&$+$Bias Disclosure with Explanation&$F(1,182)=9.18$&$0.0028$&$*$&$-0.02$&$0.006$\\
        Accuracy&proxy&$+$Bias and Corr Disclosure with Explanation&$F(1,182)=10.26$&$0.0016$&$*$&$-0.02$&$0.006$\\
        Accuracy&proxy&Joint Intervention (BD)&$F(1,277)=2.44$&$0.1197$&$ $&$0.01$&$0.007$\\
        Accuracy&proxy&Joint Intervention (BD+CD)&$F(1,277)=0.73$&$0.3933$&$ $&$0.01$&$0.007$\\
        \midrule
        FNR&protected\cellcolor{customrowcolor}&\cellcolor{customrowcolor}$+$Expl&\cellcolor{customrowcolor}$F(1,115)=5.88$&\cellcolor{customrowcolor}$0.0169$&\cellcolor{customrowcolor}$*$&\cellcolor{customrowcolor}$0.07$&\cellcolor{customrowcolor}$0.028$\\
        FNR&protected\cellcolor{customrowcolor}&\cellcolor{customrowcolor}$+$Bias Disclosure&\cellcolor{customrowcolor}$F(1,50)=0.38$&\cellcolor{customrowcolor}$0.5426$&\cellcolor{customrowcolor}$ $&\cellcolor{customrowcolor}$-0.02$&\cellcolor{customrowcolor}$0.026$\\
        FNR&protected\cellcolor{customrowcolor}&\cellcolor{customrowcolor}$+$Bias Disclosure with Explanation&\cellcolor{customrowcolor}$F(1,67)=7.78$&\cellcolor{customrowcolor}$0.0069$&\cellcolor{customrowcolor}$*$&\cellcolor{customrowcolor}$-0.05$&\cellcolor{customrowcolor}$0.017$\\
        FNR&protected\cellcolor{customrowcolor}&\cellcolor{customrowcolor}Joint Intervention (BD)&\cellcolor{customrowcolor}$F(1,208)=0.05$&\cellcolor{customrowcolor}$0.8269$&\cellcolor{customrowcolor}$ $&\cellcolor{customrowcolor}$0.00$&\cellcolor{customrowcolor}$0.022$\\
        FNR&proxy&$+$Expl&$F(1,227)=3.78$&$0.0530$&$ $&$0.03$&$0.017$\\
        FNR&proxy&$+$Bias Disclosure&$F(1,113)=0.48$&$0.4921$&$ $&$-0.01$&$0.018$\\
        FNR&proxy&$+$Bias and Corr Disclosure&$F(1,115)=2.95$&$0.0887$&$ $&$-0.03$&$0.019$\\
        FNR&proxy&$+$Bias Disclosure with Explanation&$F(1,169)=1.6$&$0.2073$&$ $&$-0.02$&$0.014$\\
        FNR&proxy&$+$Bias and Corr Disclosure with Explanation&$F(1,169)=1.65$&$0.2006$&$ $&$-0.02$&$0.014$\\
        FNR&proxy&Joint Intervention (BD)&$F(1,277)=1.21$&$0.2722$&$ $&$0.02$&$0.020$\\
        FNR&proxy&Joint Intervention (BD+CD)&$F(1,277)=1.39$&$0.2396$&$ $&$0.03$&$0.020$\\
        \midrule
        FPR&protected\cellcolor{customrowcolor}&\cellcolor{customrowcolor}$+$Expl&\cellcolor{customrowcolor}$F(1,115)=9.94$&\cellcolor{customrowcolor}$0.0021$&\cellcolor{customrowcolor}$*$&\cellcolor{customrowcolor}$-0.10$&\cellcolor{customrowcolor}$0.031$\\
        FPR&protected\cellcolor{customrowcolor}&\cellcolor{customrowcolor}$+$Bias Disclosure&\cellcolor{customrowcolor}$F(1,50)=2.09$&\cellcolor{customrowcolor}$0.1546$&\cellcolor{customrowcolor}$ $&\cellcolor{customrowcolor}$0.04$&\cellcolor{customrowcolor}$0.028$\\
        FPR&protected\cellcolor{customrowcolor}&\cellcolor{customrowcolor}$+$Bias Disclosure with Explanation&\cellcolor{customrowcolor}$F(1,67)=8.35$&\cellcolor{customrowcolor}$0.0052$&\cellcolor{customrowcolor}$*$&\cellcolor{customrowcolor}$0.05$&\cellcolor{customrowcolor}$0.019$\\
        FPR&protected\cellcolor{customrowcolor}&\cellcolor{customrowcolor}Joint Intervention (BD)&\cellcolor{customrowcolor}$F(1,208)=0.17$&\cellcolor{customrowcolor}$0.6832$&\cellcolor{customrowcolor}$ $&\cellcolor{customrowcolor}$-0.01$&\cellcolor{customrowcolor}$0.025$\\
        FPR&proxy&$+$Expl&$F(1,227)=5.45$&$0.0205$&$ $&$-0.04$&$0.019$\\
        FPR&proxy&$+$Bias Disclosure&$F(1,116)=1.34$&$0.2502$&$ $&$0.03$&$0.024$\\
        FPR&proxy&$+$Bias and Corr Disclosure&$F(1,118)=5.95$&$0.0162$&$*$&$0.06$&$0.024$\\
        FPR&proxy&$+$Bias Disclosure with Explanation&$F(1,169)=10.79$&$0.0012$&$*$&$0.05$&$0.017$\\
        FPR&proxy&$+$Bias and Corr Disclosure with Explanation&$F(1,169)=7.44$&$0.0071$&$*$&$0.05$&$0.017$\\
        FPR&proxy&Joint Intervention (BD)&$F(1,277)=0.06$&$0.8092$&$ $&$-0.01$&$0.023$\\
        FPR&proxy&Joint Intervention (BD+CD)&$F(1,277)=0.84$&$0.3593$&$ $&$-0.02$&$0.023$\\
        \midrule
        better&protected\cellcolor{customrowcolor}&\cellcolor{customrowcolor}$+$Expl&\cellcolor{customrowcolor}$F(1,116)=2.57$&\cellcolor{customrowcolor}$0.1117$&\cellcolor{customrowcolor}&\cellcolor{customrowcolor}$-0.32$&\cellcolor{customrowcolor}$0.199$\\
        better&protected\cellcolor{customrowcolor}&\cellcolor{customrowcolor}$+$Bias Disclosure&\cellcolor{customrowcolor}$F(1,50)=0.39$&\cellcolor{customrowcolor}$0.5369$&\cellcolor{customrowcolor}&\cellcolor{customrowcolor}$-0.06$&\cellcolor{customrowcolor}$0.095$\\
        better&protected\cellcolor{customrowcolor}&\cellcolor{customrowcolor}$+$Bias Disclosure with Explanation&\cellcolor{customrowcolor}$F(1,67)=1.06$&\cellcolor{customrowcolor}$0.3070$&\cellcolor{customrowcolor}&\cellcolor{customrowcolor}$-0.09$&\cellcolor{customrowcolor}$0.086$\\
        better&proxy&$+$Expl&$F(1,228)=0.0$&$0.9726$&&$0.00$&$0.143$\\
        better&proxy&$+$Bias Disclosure&$F(1,101)=5.0$&$0.0276$&&$-0.21$&$0.095$\\
        better&proxy&$+$Bias and Corr Disclosure&$F(1,102)=0.77$&$0.3838$&&$-0.09$&$0.098$\\
        better&proxy&$+$Bias Disclosure with Explanation&$F(1,151)=0.07$&$0.7884$&&$0.02$&$0.085$\\
        better&proxy&$+$Bias and Corr Disclosure with Explanation&$F(1,151)=10.12$&$0.0018$&$*$&$-0.27$&$0.085$\\
        \midrule
        confidence&protected\cellcolor{customrowcolor}&\cellcolor{customrowcolor}$+$Expl&\cellcolor{customrowcolor}$F(1,116)=1.71$&\cellcolor{customrowcolor}$0.1933$&\cellcolor{customrowcolor}&\cellcolor{customrowcolor}$-0.23$&\cellcolor{customrowcolor}$0.176$\\
        confidence&protected\cellcolor{customrowcolor}&\cellcolor{customrowcolor}$+$Bias Disclosure&\cellcolor{customrowcolor}$F(1,50)=0.04$&\cellcolor{customrowcolor}$0.8496$&\cellcolor{customrowcolor}&\cellcolor{customrowcolor}$-0.02$&\cellcolor{customrowcolor}$0.103$\\
        confidence&protected\cellcolor{customrowcolor}&\cellcolor{customrowcolor}$+$Bias Disclosure with Explanation&\cellcolor{customrowcolor}$F(1,67)=1.2$&\cellcolor{customrowcolor}$0.2765$&\cellcolor{customrowcolor}&\cellcolor{customrowcolor}$-0.09$&\cellcolor{customrowcolor}$0.080$\\
        confidence&proxy&$+$Expl&$F(1,228)=1.41$&$0.2356$&&$0.16$&$0.131$\\
        confidence&proxy&$+$Bias Disclosure&$F(1,109)=0.5$&$0.4798$&&$-0.08$&$0.106$\\
        confidence&proxy&$+$Bias and Corr Disclosure&$F(1,110)=3.66$&$0.0583$&&$-0.21$&$0.109$\\
        confidence&proxy&$+$Bias Disclosure with Explanation&$F(1,149)=13.17$&$0.0004$&$*$&$-0.28$&$0.076$\\
        confidence&proxy&$+$Bias and Corr Disclosure with Explanation&$F(1,149)=19.32$&$0.0000$&$*$&$-0.33$&$0.076$\\
        \midrule
        predictable&protected\cellcolor{customrowcolor}&\cellcolor{customrowcolor}$+$Expl&\cellcolor{customrowcolor}$F(1,116)=0.48$&\cellcolor{customrowcolor}$0.4883$&\cellcolor{customrowcolor}&\cellcolor{customrowcolor}$0.14$&\cellcolor{customrowcolor}$0.204$\\
        predictable&protected\cellcolor{customrowcolor}&\cellcolor{customrowcolor}$+$Bias Disclosure&\cellcolor{customrowcolor}$F(1,50)=0.0$&\cellcolor{customrowcolor}$1.0000$&\cellcolor{customrowcolor}&\cellcolor{customrowcolor}$0.00$&\cellcolor{customrowcolor}$0.125$\\
        predictable&protected\cellcolor{customrowcolor}&\cellcolor{customrowcolor}$+$Bias Disclosure with Explanation&\cellcolor{customrowcolor}$F(1,67)=0.01$&\cellcolor{customrowcolor}$0.9038$&\cellcolor{customrowcolor}&\cellcolor{customrowcolor}$-0.01$&\cellcolor{customrowcolor}$0.121$\\
        predictable&proxy&$+$Expl&$F(1,228)=18.55$&$0.0000$&$*$&$0.69$&$0.161$\\
        predictable&proxy&$+$Bias Disclosure&$F(1,103)=0.05$&$0.8298$&&$0.02$&$0.110$\\
        predictable&proxy&$+$Bias and Corr Disclosure&$F(1,103)=6.39$&$0.0130$&$*$&$0.29$&$0.113$\\
        predictable&proxy&$+$Bias Disclosure with Explanation&$F(1,156)=0.2$&$0.6559$&&$0.05$&$0.110$\\
        predictable&proxy&$+$Bias and Corr Disclosure with Explanation&$F(1,156)=1.22$&$0.2716$&&$-0.12$&$0.110$\\
        \midrule
        safe&protected\cellcolor{customrowcolor}&\cellcolor{customrowcolor}$+$Expl&\cellcolor{customrowcolor}$F(1,116)=0.03$&\cellcolor{customrowcolor}$0.8677$&\cellcolor{customrowcolor}&\cellcolor{customrowcolor}$0.03$&\cellcolor{customrowcolor}$0.176$\\
        safe&protected\cellcolor{customrowcolor}&\cellcolor{customrowcolor}$+$Bias Disclosure&\cellcolor{customrowcolor}$F(1,50)=1.69$&\cellcolor{customrowcolor}$0.1997$&\cellcolor{customrowcolor}&\cellcolor{customrowcolor}$-0.10$&\cellcolor{customrowcolor}$0.075$\\
        safe&protected\cellcolor{customrowcolor}&\cellcolor{customrowcolor}$+$Bias Disclosure with Explanation&\cellcolor{customrowcolor}$F(1,67)=2.96$&\cellcolor{customrowcolor}$0.0898$&\cellcolor{customrowcolor}&\cellcolor{customrowcolor}$-0.18$&\cellcolor{customrowcolor}$0.103$\\
        safe&proxy&$+$Expl&$F(1,228)=4.17$&$0.0422$&&$0.24$&$0.119$\\
        safe&proxy&$+$Bias Disclosure&$F(1,107)=0.08$&$0.7787$&&$0.03$&$0.099$\\
        safe&proxy&$+$Bias and Corr Disclosure&$F(1,108)=0.01$&$0.9419$&&$-0.01$&$0.102$\\
        safe&proxy&$+$Bias Disclosure with Explanation&$F(1,158)=3.27$&$0.0727$&&$-0.17$&$0.092$\\
        safe&proxy&$+$Bias and Corr Disclosure with Explanation&$F(1,158)=7.76$&$0.0060$&$*$&$-0.25$&$0.092$\\
        \midrule
        wary&protected\cellcolor{customrowcolor}&\cellcolor{customrowcolor}$+$Expl&\cellcolor{customrowcolor}$F(1,116)=0.12$&\cellcolor{customrowcolor}$0.7283$&\cellcolor{customrowcolor}&\cellcolor{customrowcolor}$0.06$&\cellcolor{customrowcolor}$0.183$\\
        wary&protected\cellcolor{customrowcolor}&\cellcolor{customrowcolor}$+$Bias Disclosure&\cellcolor{customrowcolor}$F(1,50)=0.0$&\cellcolor{customrowcolor}$1.0000$&\cellcolor{customrowcolor}&\cellcolor{customrowcolor}$0.00$&\cellcolor{customrowcolor}$0.097$\\
        wary&protected\cellcolor{customrowcolor}&\cellcolor{customrowcolor}$+$Bias Disclosure with Explanation&\cellcolor{customrowcolor}$F(1,67)=1.6$&\cellcolor{customrowcolor}$0.2100$&\cellcolor{customrowcolor}&\cellcolor{customrowcolor}$0.13$&\cellcolor{customrowcolor}$0.105$\\
        wary&proxy&$+$Expl&$F(1,228)=0.0$&$0.9621$&&$0.01$&$0.118$\\
        wary&proxy&$+$Bias Disclosure&$F(1,113)=4.48$&$0.0365$&&$0.24$&$0.114$\\
        wary&proxy&$+$Bias and Corr Disclosure&$F(1,115)=1.05$&$0.3066$&&$0.12$&$0.117$\\
        wary&proxy&$+$Bias Disclosure with Explanation&$F(1,153)=2.69$&$0.1030$&&$0.13$&$0.080$\\
        wary&proxy&$+$Bias and Corr Disclosure with Explanation&$F(1,153)=13.28$&$0.0004$&$*$&$0.29$&$0.080$\\
        \midrule
        works&protected\cellcolor{customrowcolor}&\cellcolor{customrowcolor}$+$Expl&\cellcolor{customrowcolor}$F(1,116)=6.0$&\cellcolor{customrowcolor}$0.0158$&\cellcolor{customrowcolor}$*$&\cellcolor{customrowcolor}$-0.38$&\cellcolor{customrowcolor}$0.154$\\
        works&protected\cellcolor{customrowcolor}&\cellcolor{customrowcolor}$+$Bias Disclosure&\cellcolor{customrowcolor}$F(1,50)=1.5$&\cellcolor{customrowcolor}$0.2265$&\cellcolor{customrowcolor}&\cellcolor{customrowcolor}$-0.14$&\cellcolor{customrowcolor}$0.112$\\
        works&protected\cellcolor{customrowcolor}&\cellcolor{customrowcolor}$+$Bias Disclosure with Explanation&\cellcolor{customrowcolor}$F(1,67)=2.56$&\cellcolor{customrowcolor}$0.1145$&\cellcolor{customrowcolor}&\cellcolor{customrowcolor}$-0.15$&\cellcolor{customrowcolor}$0.092$\\
        works&proxy&$+$Expl&$F(1,228)=0.06$&$0.8122$&&$0.03$&$0.138$\\
        works&proxy&$+$Bias Disclosure&$F(1,105)=4.43$&$0.0376$&&$-0.22$&$0.104$\\
        works&proxy&$+$Bias and Corr Disclosure&$F(1,106)=14.58$&$0.0002$&$*$&$-0.41$&$0.107$\\
        works&proxy&$+$Bias Disclosure with Explanation&$F(1,153)=5.43$&$0.0211$&&$-0.20$&$0.088$\\
        works&proxy&$+$Bias and Corr Disclosure with Explanation&$F(1,153)=14.19$&$0.0002$&$*$&$-0.33$&$0.088$\\
        \bottomrule
        \end{tabular}}
        \caption{Overall results of tests regarding the primary effects of our study on fairness perceptions, decision-making fairness, decision-making quality, and learned trust.}
        \label{tab:full}
    \end{table*}
}{}

\clearpage

\section{Human Study Interface}

\begin{figure}[ht]
    \centering
    \includegraphics[width=\textwidth, trim={0 44 0 0},clip]{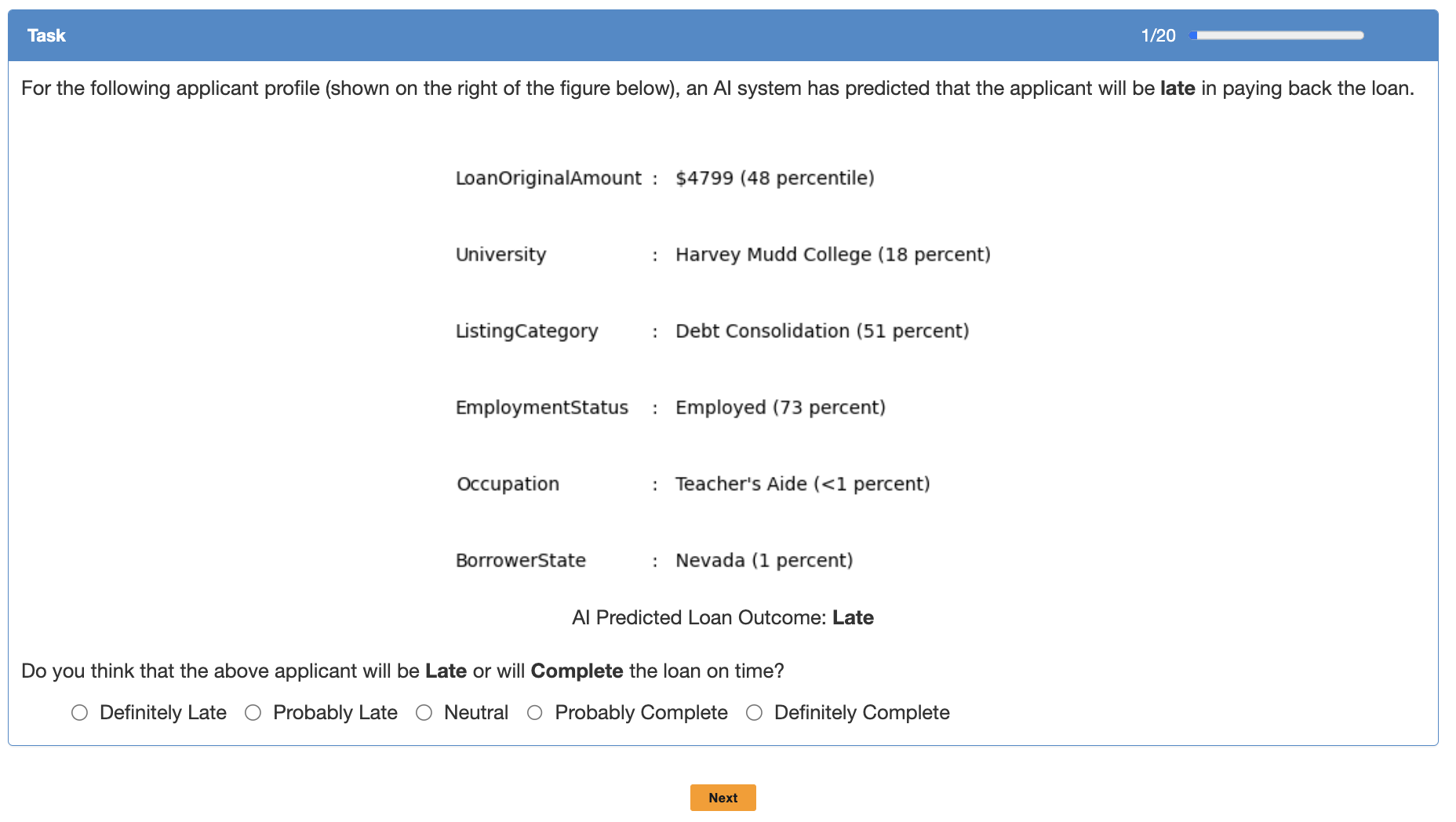}
    \caption{Example task question without explanations.}
    \label{fig:task}
\end{figure}

\begin{figure}[ht]
    \centering
    \includegraphics[width=\textwidth, trim={0 44 0 0},clip]{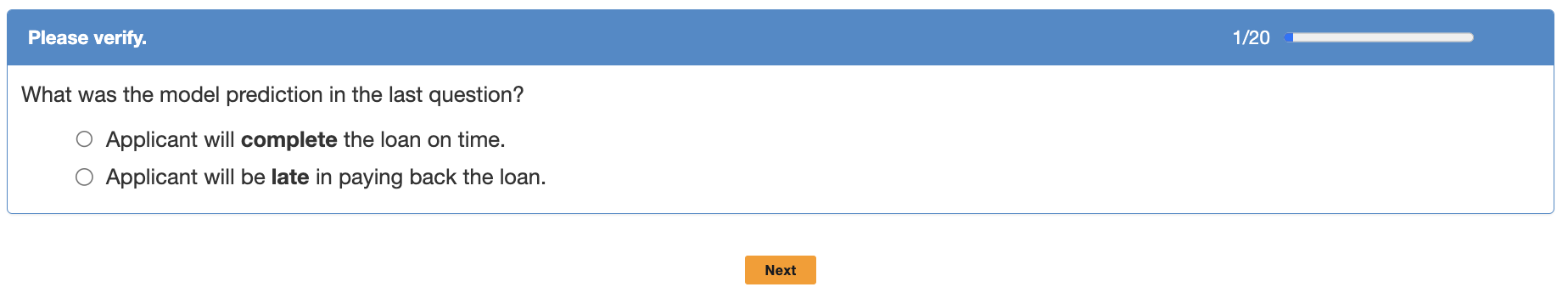}
    \caption{Example attention check question.}
    \label{fig:attention_check}
\end{figure}

\begin{figure}[ht]
    \centering
    \includegraphics[width=\textwidth, trim={0 44 0 0},clip]{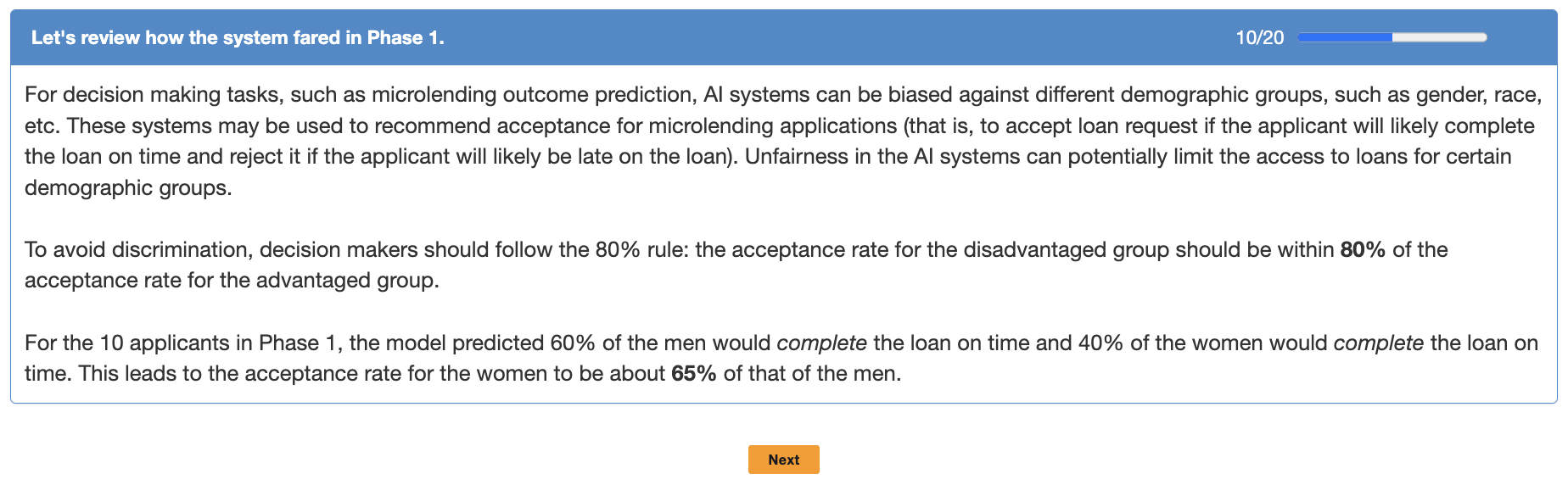}
    \caption{Bias disclosure showing the demographic parity of the model in phase 1.}
    \label{fig:bias_disclosure}
\end{figure}

\begin{figure}[ht]
    \centering
    \includegraphics[width=\textwidth, trim={0 44 0 0},clip]{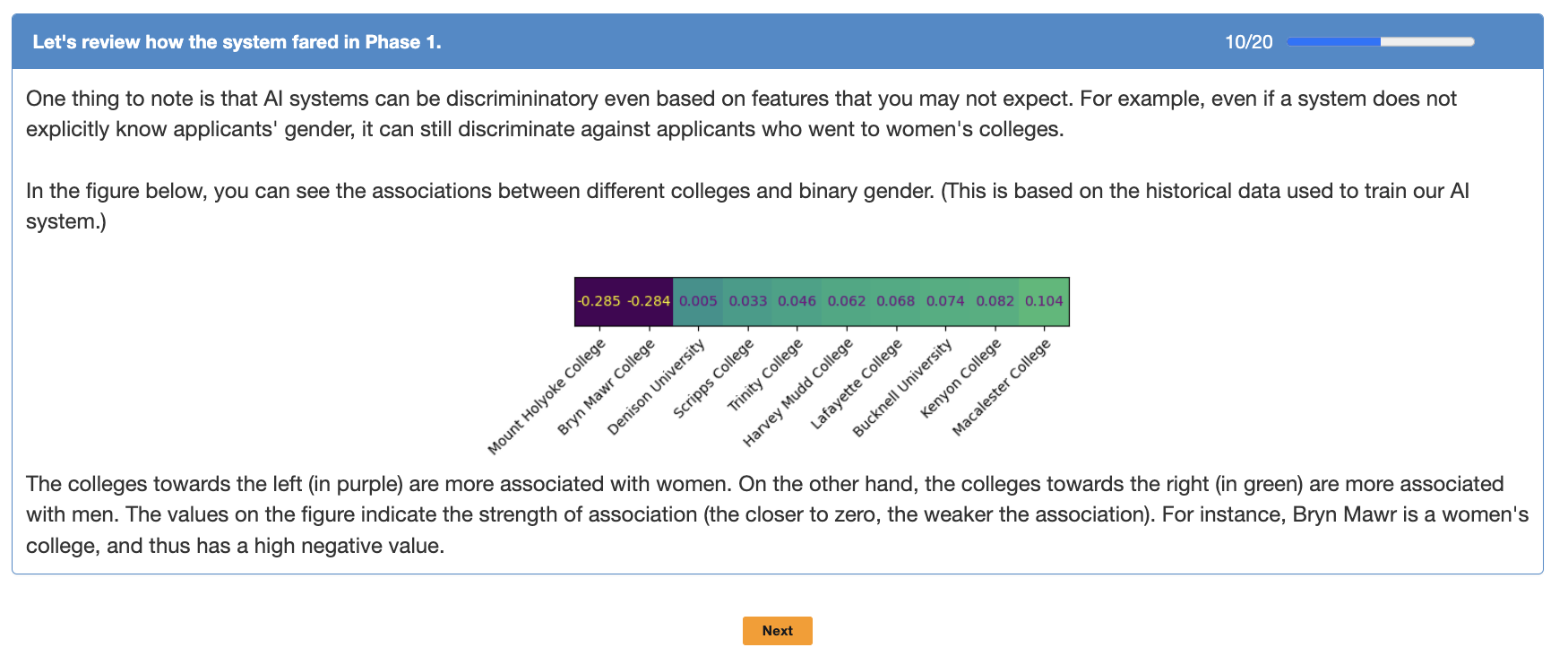}
    \caption{Correlation disclosure showing the relationship between university and gender in our synthetic data.}
    \label{fig:corr_disclosure}
\end{figure}

\begin{figure}[ht]
    \centering
    \includegraphics[width=\textwidth, trim={0 44 0 0},clip]{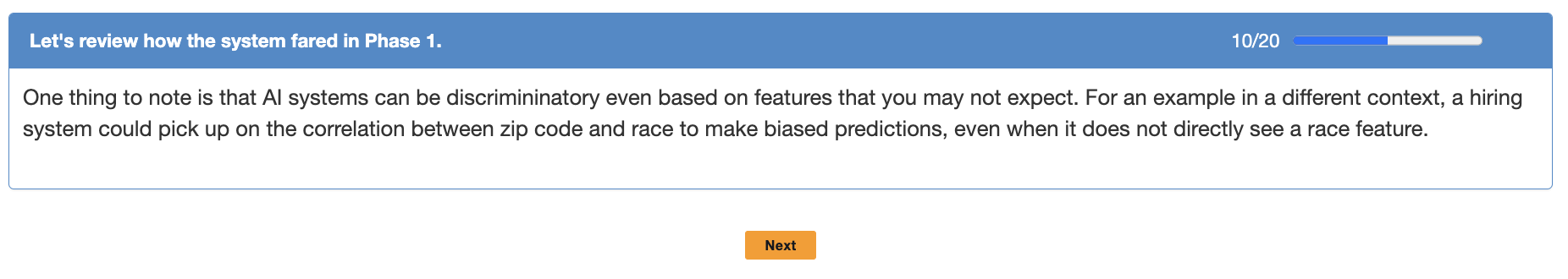}
    \caption{In proxy conditions where participants are not given correlation disclosure. They are instead given this screen explaining that proxies can general, without mentioning the relationship in our data.}
    \label{fig:dummy_corr_disclosure}
\end{figure}

\begin{figure}[ht]
    \centering
    \includegraphics[width=\textwidth, trim={0 44 0 0},clip]{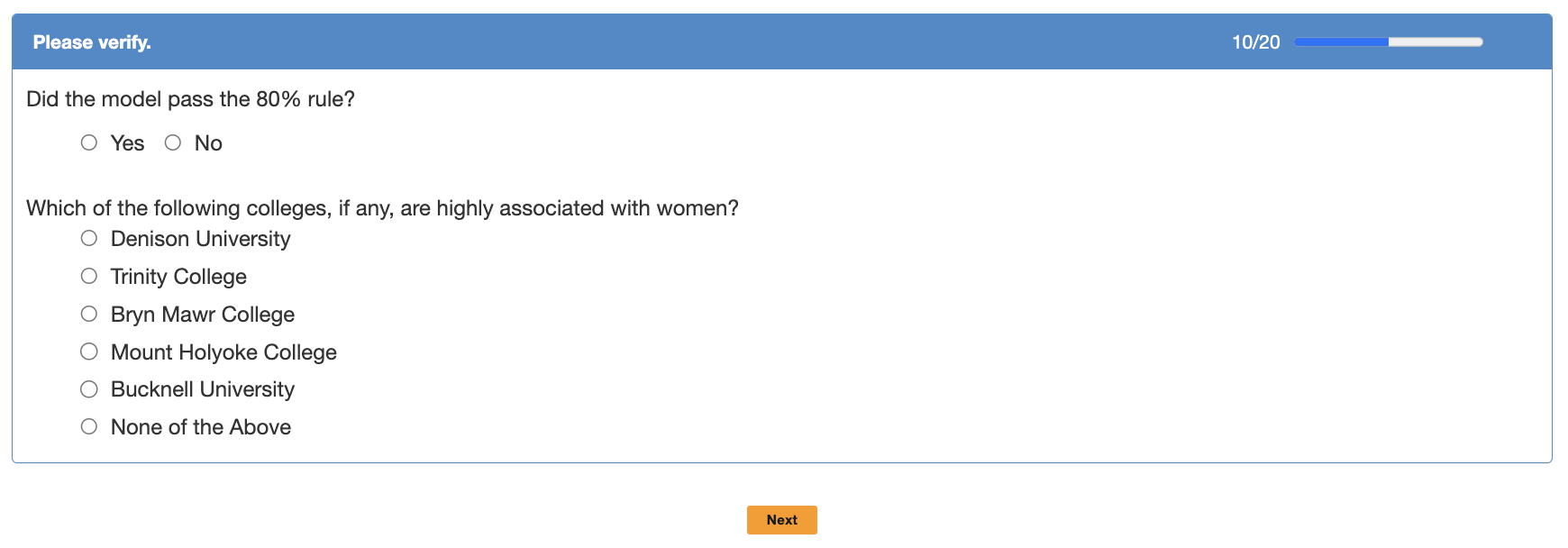}
    \caption{Comprehension check screen testing both understanding of bias disclosure (Figure~\ref{fig:bias_disclosure}) and correlation disclosure (Figure~\ref{fig:corr_disclosure}). The correlation disclosure question is only shown in proxy conditions where the participants are given correlation disclosure.}
    \label{fig:corr_check}
\end{figure}

\begin{figure}[ht]
    \centering
    \includegraphics[width=\textwidth, trim={0 44 0 0},clip]{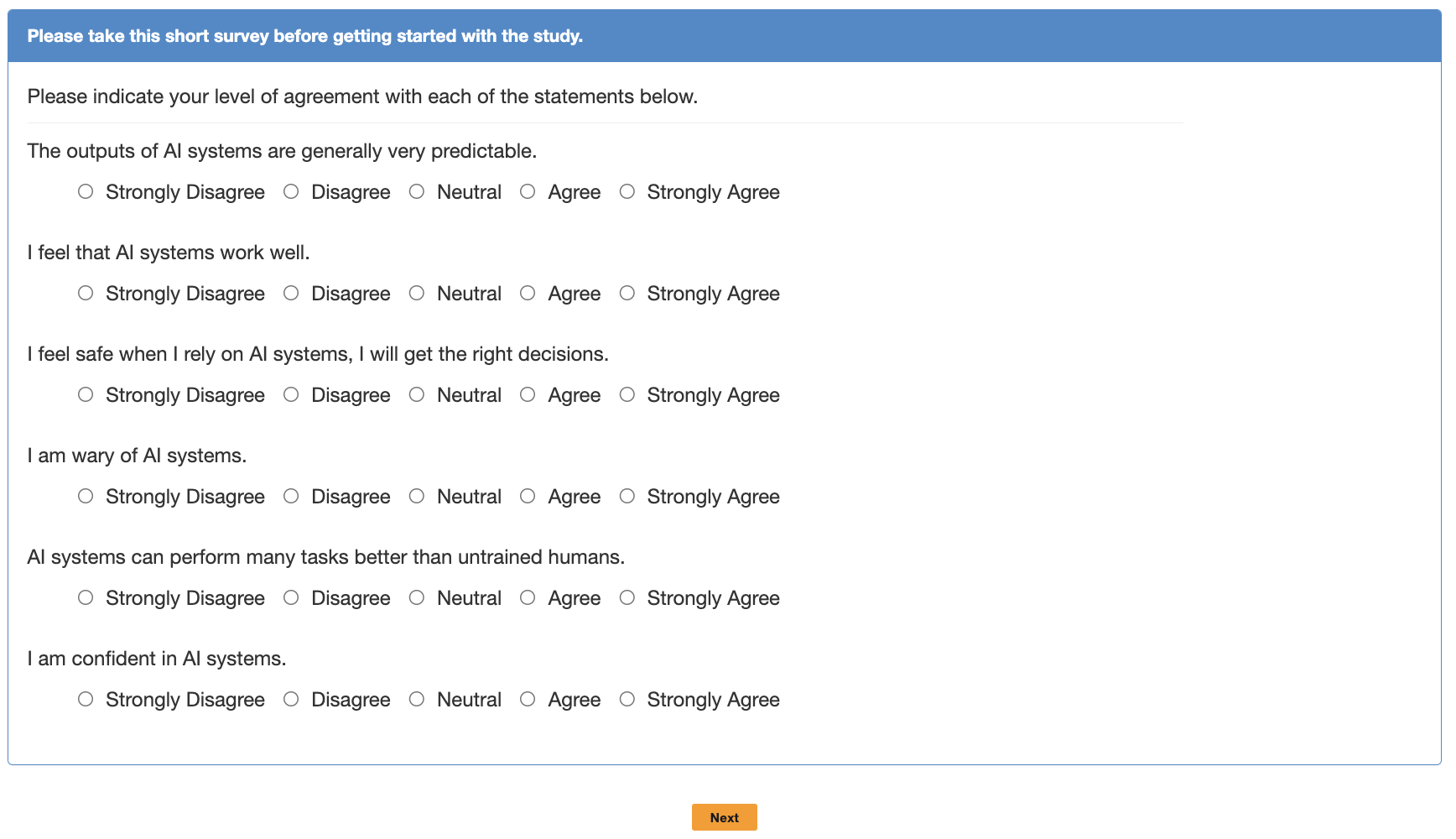}
    \caption{Initial trust survey given before the task is introduced.}
    \label{fig:pre_study_survey}
\end{figure}

\begin{figure}[ht]
    \centering
    \includegraphics[width=\textwidth, trim={0 30 0 0},clip]{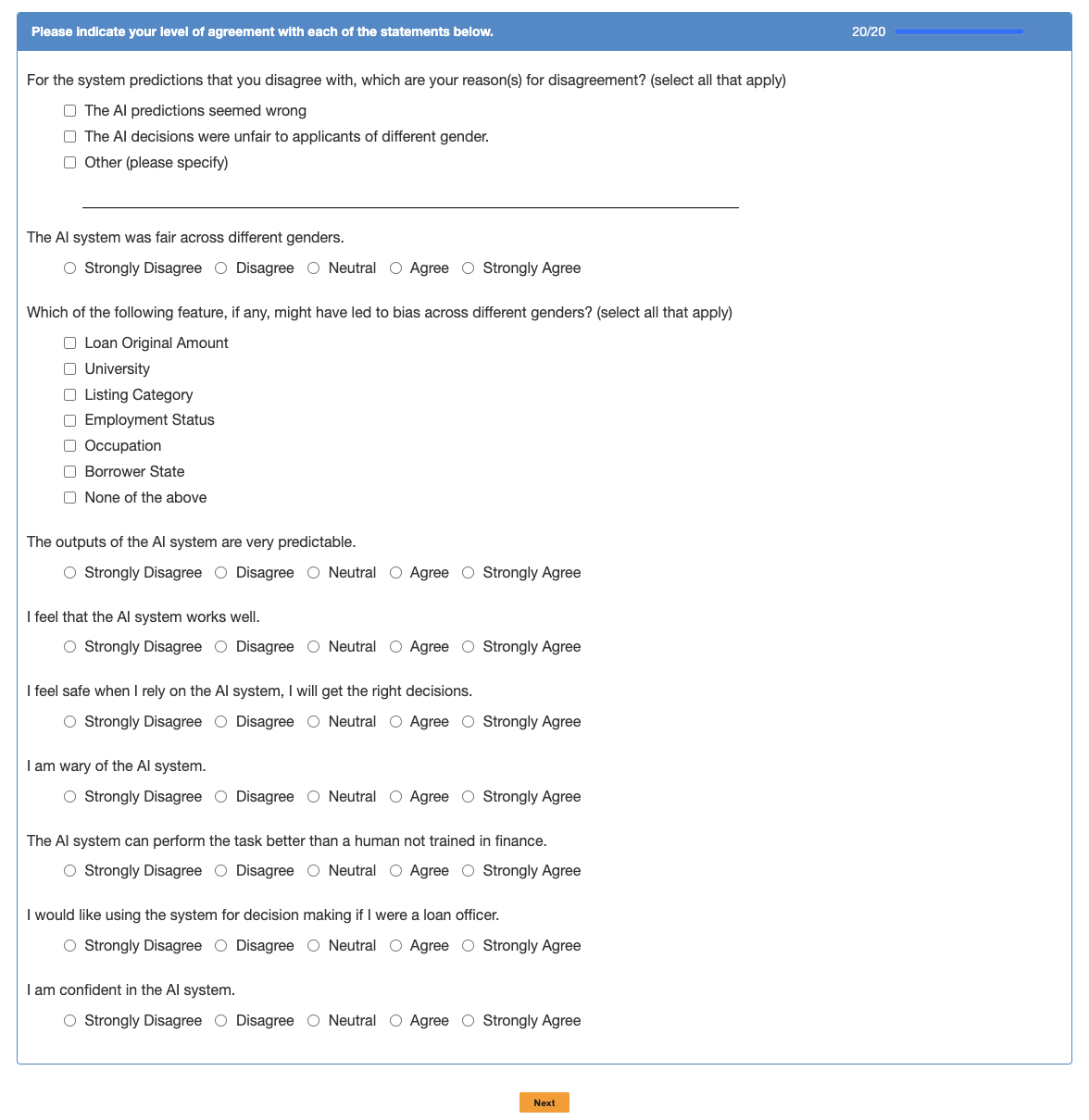}
    \caption{Example post-task survey. This is the version that is shown after phase 2 of proxy conditions. In protected conditions and after phase 1 of proxy conditions, the question about which feature might have lead to gender bias is omitted.}
    \label{fig:survey_p2}
\end{figure}

\begin{figure}[ht]
    \centering
    \includegraphics[width=\textwidth, trim={0 44 0 0},clip]{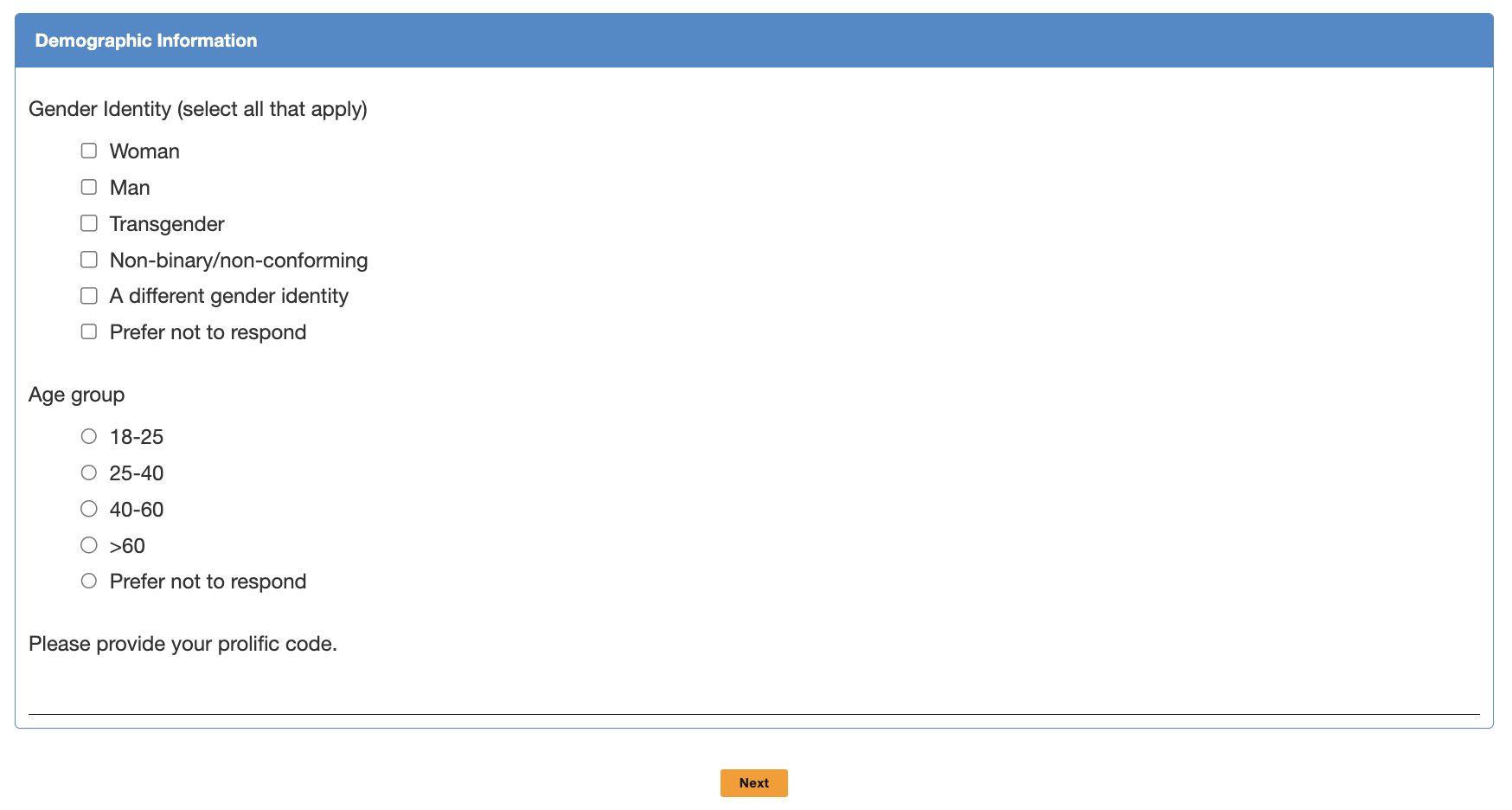}
    \caption{Participant demographic questions.}
    \label{fig:demographic}
\end{figure}

\end{document}